\documentclass[10pt,twocolumn,letterpaper]{article}

\usepackage[pagenumbers]{cvpr} 

\usepackage[T1]{fontenc}
\usepackage{booktabs, multirow} 
\usepackage{soul}
\usepackage{xcolor,colortbl} 
\usepackage{changepage,threeparttable} 
\usepackage{svg}
\usepackage[normalem]{ulem}
\usepackage{graphicx}
\usepackage[accsupp]{axessibility}  

\definecolor{cvprblue}{rgb}{0.21,0.49,0.74}
\usepackage[pagebackref,breaklinks,colorlinks,allcolors=cvprblue]{hyperref}

\def\paperID{7521} 
\def\confName{CVPR}
\def\confYear{2025}

\title{Any-Resolution AI-Generated Image Detection by Spectral Learning}
\author{
Dimitrios Karageorgiou$^{1,2}$ \quad Symeon Papadopoulos$^1$ \quad Ioannis Kompatsiaris$^1$ \quad Efstratios Gavves$^{2,3}$  \vspace{2pt} \\
$^1$Information Technologies Institute, CERTH, Greece\\
$^2$University of Amsterdam, The Netherlands \\
$^3$Archimedes/Athena RC, Greece \\
{\tt\small \{dkarageo,papadop,ikom\}@iti.gr}  \quad
{\tt\small \{d.karageorgiou,e.gavves\}@uva.nl} \vspace{-4pt}
}

\begin{document}
\maketitle
\begin{abstract}
Recent works have established that AI models introduce spectral artifacts into generated images and propose approaches for learning to capture them using labeled data. However, the significant differences in such artifacts among different generative models hinder these approaches from generalizing to generators not seen during training. In this work, we build upon the key idea that the spectral distribution of real images constitutes both an invariant and highly discriminative pattern for AI-generated image detection. To model this under a self-supervised setup, we employ masked spectral learning using the pretext task of frequency reconstruction. Since generated images constitute out-of-distribution samples for this model, we propose spectral reconstruction similarity to capture this divergence. Moreover, we introduce spectral context attention, which enables our approach to efficiently capture subtle spectral inconsistencies in images of any resolution. Our spectral AI-generated image detection approach (SPAI) achieves a 5.5\% absolute improvement in AUC over the previous state-of-the-art across 13 recent generative approaches, while exhibiting robustness against common online perturbations. Code is available on \href{https://mever-team.github.io/spai}{https://mever-team.github.io/spai}.
\end{abstract}    
\section{Introduction}
\label{sec:introduction}

\setlength{\textfloatsep}{12pt}{
\begin{figure}
  \centering
  \includesvg[width=0.99999\columnwidth]{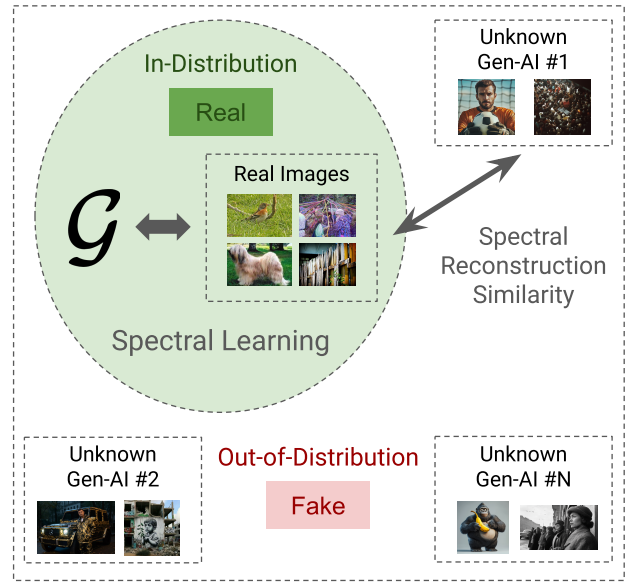}
  \caption{ SPAI employs spectral learning to learn the spectral distribution of real images under a self-supervised setup. Then, using the spectral reconstruction similarity it detects AI-generated images as out-of-distribution samples of this learned model.}
  \label{fig:summary}
\end{figure}
}

Generative AI technology advances at an outstanding pace \cite{elasri2022image}. Early AI approaches for image synthesis, primarily based on Generative Adversarial Networks (GANs) \cite{de2023review}, have recently been superseded by Diffusion Models (DMs) \cite{croitoru2023diffusion}, capable of producing high-fidelity imagery, while supporting advanced conditioning schemes \cite{cao2024controllable}. Recent methods for efficient training \cite{hu2021lora,kasymov2024autolora} have further lowered the barrier for creating new generative models. In this landscape of numerous accessible high-performing commercial and open-source models, making strong assumptions about safeguards implemented at the model level \cite{li2024art} seems intractable, leading to an increasing numbers of harmful synthetic imagery on the Internet~\cite{dufour2024ammeba}, which calls for robust AI-generated Image Detection (AID) approaches. 

To this end, the image forensics community has established that generative models introduce subtle artifacts to the generated images~\cite{wang2020cnn,tan2024rethinking,sarkar2024shadows} and proposed several AID methods for capturing them \cite{lin2024detecting}. However, even generators with minimal differences introduce significantly different artifacts \cite{bammey2023synthbuster,tariang2024synthetic}. Consequently, existing detectors poorly generalize to images originating from generators not seen during training~\cite{tariang2024synthetic,cozzolino2025zero}, while, due to the large number of available generative models, it is very difficult, if not intractable, to maintain an exhaustive up-to-date training dataset. Thus, current AID approaches perform poorly when tested outside of lab-controlled environments\cite{karageorgiou2024evolution,dufour2024ammeba}. 

Recently, the idea of modeling the distribution of real images emerged as an alternative to learning to capture the artifacts introduced by specific detectors \cite{cozzolino2025zero}. Under this new paradigm, synthetic images can be considered as out-of-distribution samples with respect to a model of real images. Early attempts to implement this idea mostly focused on modeling the spatial relationships among pixels of real images \cite{cozzolino2025zero,he2024rigid}. However, several works found that real and AI-generated images are better distinguishable in the spectral domain~\cite{bammey2023synthbuster,li2024masksim}. A fundamental principle of computer vision suggests looking for the most suitable invariant patterns for the task to be solved~\cite{forsyth2002computer}. In the case of the AID task we argue that the spectral distribution of real images is such a suitable invariant pattern as it is not directly affected by the introduction of specific generative models, while it provides significant discriminative power. In this paper, we present SPectral AI-generated Image detection (SPAI) that introduces both a modeling approach for the spectral distribution of real images as well as an architecture for capturing instances that deviate from this learned distribution.

To learn a spectral model of the real images we show that the pretext task of frequency reconstruction is an effective approach for modeling their frequency distribution, under a self-supervised training setup that uses only real images. Then, to detect AI-generated images as out-of-distribution samples of this model, we exploit the observation that a frequency reconstruction model trained on real images will reconstruct their frequencies more accurately, compared to the generated ones. We introduce the concept of spectral reconstruction similarity for estimating the divergence of the reconstructed frequencies from the ones actually present in the image under question.

Moreover, capturing subtle clues in images is crucial for effectively distinguishing among real and AI-generated ones~\cite{corvi2023detection}. However, most computer vision models, having been designed for semantic-based tasks, cannot efficiently scale to the native resolution of modern photos~\cite{yao2024hiri}, i.e. many megapixels. Therefore, pre-processing operations are required~\cite{konstantinidou2024texturecrop}, effectively dropping a significant amount of discriminative information. To utilize all the spectral information present in the image we introduce the spectral context attention that enables us to efficiently process any-resolution images without prior pre-processing. We evaluate our approach across a set of 13 recently proposed generative approaches and five sources of real images, and achieve a 5.5\% absolute improvement over state-of-the-art, while achieving superior robustness to common online perturbations. \cref{fig:summary} presents an overview of the paper.

In summary, our contributions include the following:
\begin{itemize}
    \item We introduce the spectral distribution of real images as a suitable invariant pattern for distinguishing between real and AI-generated images and we propose the use of masked spectral learning for modeling it.    
    \item We show that the pretext task of frequency reconstruction is an effective approach for modeling the spectral distribution of real images under a self-supervised setup.
    \item We introduce Spectral Reconstruction Similarity (SRS) for detecting AI-generated images as out-of-distribution samples of our learned spectral model of real images.
    \item We introduce Spectral Context Attention (SCA) to capture subtle spectral details in any-resolution images.
    \item We show a 5.5\% performance increase compared to the state-of-the-art across a set of 13 generative models, while our approach exhibits superior robustness against several perturbations encountered online.
\end{itemize}

\section{Related Work}
\label{sec:related_work}

\textbf{Image Generation:} Image generation approaches have rapidly evolved from the generation of low-quality imagery~\cite{gregor2015draw} to that of photorealistic visual content of any topic~\cite{podell2023sdxl,esser2024scaling}. Early approaches, primarily based on GANs~\cite{goodfellow2014generative,goodfellow2020generative}, established the capability of conditionally generating visual content from random noise, using additional textual~\cite{xu2018attngan,zhu2019dm,tao2020df} or visual~\cite{zhu2017unpaired,choi2018stargan,park2019semantic} inputs to control the process. More recently, the introduction of DMs~\cite{ho2020denoising,sohl2015deep} for image synthesis has set new standards to the quality of the generated content, using architectures that learn to reverse the diffusion process, either in the spatial~\cite{ho2020denoising} or in the latent~\cite{rombach2022high} domain. As such, DMs have mostly superseded GANs in generative modeling, with only a few exceptions that adopt ideas introduced for DMs to GANs~\cite{kang2023scaling}. Moreover, recent works greatly increased the resolution of the generated images~\cite{podell2023sdxl,esser2024scaling}, while techniques such as the Low-Rank Adaptation~\cite{gu2024mix,hu2021lora} significantly enhanced the efficiency of adapting generative models to any domain. As a consequence, a large and continuously increasing arsenal of open-source and commercial image generation approaches is currently available~\cite{zhang2023text,betker2023improving,esser2024scaling}.

\begin{figure*}
  \centering
  \includesvg[width=0.999\textwidth]{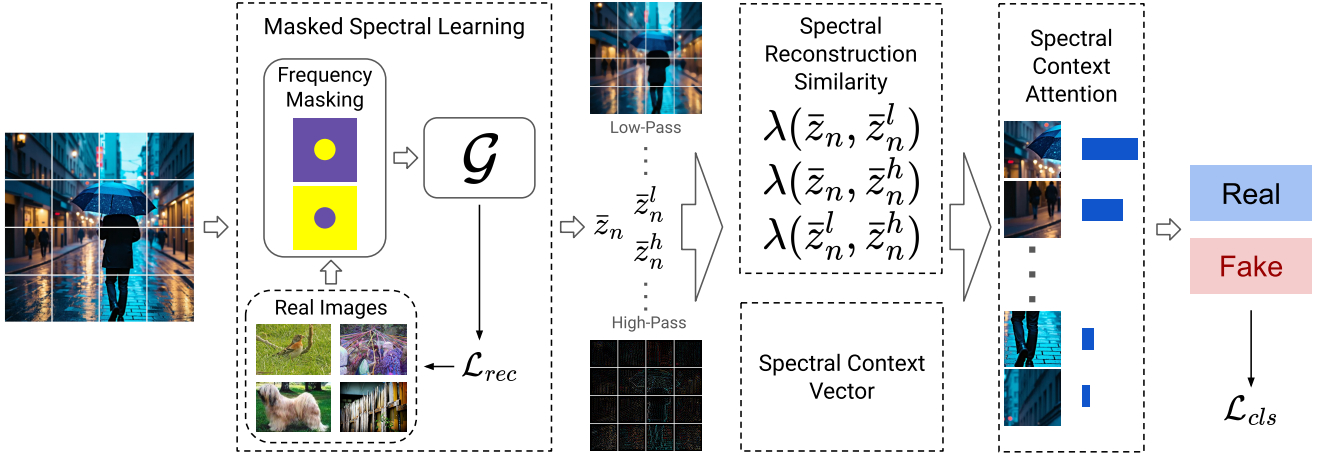}
  \vspace{-13pt}
  \caption{Overview of the SPAI approach. We learn a model of the spectral distribution of real images under a self-supervised setup using masked spectral learning. Then, we use the spectral reconstruction similarity to measure the divergence from this learned distribution and detect AI-generated images as out-of-distribution samples of this model. Spectral context vector captures the spectral context under which the spectral reconstruction similarity values are computed, while spectral context attention enables the processing of any-resolution images for capturing subtle spectral inconsistencies.}
  \vspace{-10pt}
  \label{fig:approach_overview}
\end{figure*}

\textbf{AI-Generated Image Detection:} In response, several approaches have been proposed for distinguishing real from AI-generated (synthetic) images, either by detecting semantic inconsistencies or by capturing low-level artifacts introduced by the generative models. In the former group, recent
approaches detect inconsistencies in facial geometry~\cite{bohavcek2023geometric} as well as in shadows~\cite{sarkar2024shadows}, perspective~\cite{farid2022perspective} and lighting~\cite{farid2022lighting} of the image. However, as improved generative models get released, such visible inconsistencies cease to exist. To this end, recent works have established that generative models introduce low-level artifacts and proposed approaches for capturing them either in the spatial~\cite{wang2020cnn,corvi2023detection} or in the spectral~\cite{durall2020watch,bammey2023synthbuster,li2024masksim} domain. However, such artifacts significantly differ even between models with minimal differences~\cite{bammey2023synthbuster}. In pursuit of artifacts that better generalize across different generators, others have proposed architectures for capturing inconsistencies in texture-rich image regions~\cite{liu2020global,ju2022fusing,zhong2024patchcraft}, gradient differences~\cite{tan2023learning} or artifacts introduced by the upscaling layers of the generative models~\cite{tan2024rethinking} as well as spectral augmentation approaches~\cite{doloriel2024frequency}. Moreover, several recent works have employed features from pre-trained CLIP~\cite{radford2021learning} encoders for capturing both low-level and semantic artifacts~\cite{koutlis2024leveraging,cozzolino2024raising,ojha2023towards,sha2023fake,yan2024sanity}. Common ground in all the aforementioned works is the attempt to model artifacts introduced by the generative approaches, with the intention of using them for the AID task. As such, they fail to generalize on unseen generators that introduce artifacts different than the ones present in the training set. 

\textbf{Modeling Real Images:} To better generalize on unseen generative models the direction of modeling the distribution of real images has recently emerged as an alternative solution to the AID task. The key idea is that AI-generated images constitute out-of-distribution samples with respect to a model of real (authentic) images. Early approaches towards this direction noticed that generative models cannot reconstruct real images as accurately as AI-generated ones~\cite{wang2023dire,ricker2024aeroblade,cazenavette2024fakeinversion}. Recently, He et al.~\cite{he2024rigid} found that noise perturbations on generated images introduce more variance to the features extracted by a pre-trained encoder, compared to real images, while Cozzolino et al.~\cite{cozzolino2025zero} observed a smaller error on the reconstruction of real images using a super-resolution model. Such works do not take into account the fact that most artifacts introduced by the generative models are better visible in the frequency domain~\cite{corvi2023detection,bammey2023synthbuster,li2024masksim}. 

In this paper, we argue that modeling the spectral distribution of real images is crucial for detecting AI-generated ones, as it constitutes a pattern that remains invariant to the introduction of new generative approaches, while providing significant discriminative capability. To the best of our knowledge, we are the first to employ frequency reconstruction~\cite{xie2022masked} as a pretext task for learning such a model. To detect synthetic images as out-of-distribution samples of the aforementioned model, we introduce the spectral reconstruction similarity to measure the divergence between the reconstructed and the actual frequencies of an image in the latent space. Furthermore, by introducing spectral context attention we retain all the spectral information of the image, by processing it at its original resolution.   

\section{Spectral AI-Generated Image Detection}
\label{sec:approach}

Distinguishing between AI-generated and real content, while generalizing to unknown generative models, requires capturing invariant features that remain distinctive across different generators and image transformations. 
While for AI-generated images, significant differences are visible in the frequency domain in the form of spectral artifacts introduced by the generators, their distribution significantly differs among different generative models. Instead, the spectral distribution of real images is not directly related to the introduction of specific generative approaches, but to long-term technological advances~\cite{kulinski2022towards,ni2024misalignment}, constituting an invariance for our task. Thus, our key idea is to build a spectral model of the real content and detect generated content as out-of-distribution samples of this model. To build such a detection pipeline we introduce i) a modeling approach for the spectral distribution of real images, ii) an architecture for capturing instances that deviate from this learned distribution. An overview of our architecture is shown in \cref{fig:approach_overview}.

\subsection{Masked Spectral Learning}\label{subsec:msl}

To build a spectral model of real images we propose using the pretext task of frequency reconstruction under a self-supervised learning setup, using only real images. In particular, we randomly mask the low- or high-frequency component of the input images and train the model $\mathcal{G}$ under the objective of reconstructing the missing frequencies.

\textbf{Frequency Masking}: To obtain the low- and high-frequency components of input image $x \in \mathbb{R}^{H \times W}$, we use the 2D Discrete Fourier Transform $\chi = \mathcal{F}(x) \in \mathbb{R}^{H \times W}$, where $\chi(u,v)$ defines the complex frequency value at coordinates $(u, v)$. Next, inspired by \cite{xie2022masked}, we define a mask $M$ for masking parts of the spectrum according to a fixed radius $r$ from its center, that we denote as $(c_H, c_W)$, with $d(\cdot,\cdot)$ denoting the Euclidean distance:
\begin{equation}
    M(u, v) = 
        \begin{cases}
            0 \quad\text{, }d((u,v),(c_H, c_W)) < r \\
            1 \quad\text{, otherwise}
        \end{cases}
\end{equation}
We then define the low- and high-frequency components as:
\begin{equation} 
x^{h} = \mathcal{F}^{-1}(\chi \odot M)
\end{equation}
\begin{equation}
x^{l} = \mathcal{F}^{-1}(\chi \odot (\mathbf{1} - M))
\end{equation}
\noindent where $\mathbf{1}$ defines an all-ones matrix, $\odot$ point-wise multiplication and $\mathcal{F}^{-1}$ the Inverse 2D Discrete Fourier Transform.

To train $\mathcal{G}$ we use an auxiliary decoding head $\mathcal{H}$ to predict from its output the reconstructed image $\hat{x} = \mathcal{H}(\mathcal{G}(\mathcal{B}(x^l, x^h)))$, where $\mathcal{B}(\cdot,\cdot)$ randomly samples one of its inputs according to a binomial distribution. We train the combination of $\mathcal{G}$ and $\mathcal{H}$ using the objective $\mathcal{L}_{rec}=\mathcal{D}(\mathcal{F}(x), \mathcal{F}(\hat{x}))$, where $\mathcal{D}$ defines the frequency distance introduced in \cite{xie2022masked}. $\mathcal{H}$ is discarded after training.

\subsection{Spectral Reconstruction Similarity} \label{subsec:srs}

Up to this point, $\mathcal{G}$ has learned to reconstruct the spectral distribution of real images given their low- or high-frequency components. 
Here, we introduce an approach to detect the out-of-distribution samples of $\mathcal{G}$.
In the rest of the text, we assume that $\mathcal{G}$ is a Vision Transformer (ViT)~\cite{dosovitskiy2020image}.

We apply ViT's tokenization procedure to process the input image: $x$ is split into $L=H \cdot W / p^2$ tokens of size $p \times p$. Each token is embedded to a space of size $d$ through a linear projection to formulate ViT's input sequence $z_{0} \in \mathbb{R}^{L \times d}$. Moreover, $\mathcal{G}$ comprises $N$ transformer~\cite{vaswani2017attention} blocks. We denote the output of the $n$-th transformer block as:
\begin{equation}\label{eq:vit_encoding}
    z_{n}=\mathcal{G}_n(z_0) \in \mathbb{R}^{L \times d}, n=\{1,...,N\}    
\end{equation}

We encode $x$, $x^h$ and $x^l$ into $z_n$, $z_{n}^{h}$ and $z_{n}^{l}$ respectively, using \cref{eq:vit_encoding}. Then, we use $N$ projection operators, denoted $\mathbf{P}_{n}: R^d \rightarrow R^D$, one per block of $\mathcal{G}$, to project the three aforementioned groups of representations from the feature space of $\mathcal{G}$ to one that facilitates the similarity operations that we define next. Each projection operator comprises a series of linear projection, layer normalization~\cite{lei2016layer} and GELU activation layers~\cite{hendrycks2016gaussian}. Further details regarding their structure are provided in the supplementary material. We use them to project $z_n$, $z_{n}^{h}$ and $z_{n}^{l}$ into $\bar{z}_n$, $\bar{z}_{n}^{h}$ and $\bar{z}_{n}^{l}$, all belonging to the space of $\mathbb{R}^{L \times D}$. In the rest of our approach we only use these latter representations.

As $\mathcal{G}$ constitutes a spectral model of real images, it is expected to better reconstruct their missing frequencies compared to the ones of AI-generated images. Therefore, we expect bigger distances between $\bar{z}_n$, $\bar{z}_{n}^{h}$ and $\bar{z}_{n}^{l}$ for AI-generated images. As such, we expect the similarity among the features of the original image, the low- and the high-frequency components to indicate the alignment of the spectral distribution of an image with the learned distribution of $\mathcal{G}$. To measure this divergence we introduce the concept of Spectral Reconstruction Similarity (SRS). For two representations $z^{A}$ and $z^B$ in space $\mathbb{R}^{L \times D}$ we define SRS as:
\begin{equation}\label{eq:frs}
    \lambda(z^{A}, z^{B}) = \frac{z^{A} \cdot z^{B}}{\parallel z^{A} \parallel \parallel z^{B} \parallel} \in [-1, 1]^{L \times 1}.
\end{equation}
\noindent Then, using \cref{eq:frs} we define i) the SRS between the representations of the original image and the low-pass filtered one $\omega_{n}^{ol}=\lambda({\bar{z}_n}, \bar{z}_{n}^{l})$, ii) between the original image and the high-pass filtered one $\omega_{n}^{oh}=\lambda({\bar{z}_n}, \bar{z}_{n}^{h})$ and iii) between the low and high-pass filtered components $\omega_{n}^{lh}=\lambda({\bar{z}_n}^{l}, \bar{z}_{n}^{h})$. 

Each of the SRS vectors $\omega_{n}^{ol}$, $\omega_{n}^{oh}$, $\omega_{n}^{lh}$ includes $L$ values, i.e. one for each of the $L$ tokens of the ViT that capture information of a $p \times p$ region of the image. We then calculate their mean and standard deviation, resulting in two scalars for each SRS vector. We concatenate these six values computed for the $N$ blocks of $\mathcal{G}$ into $z^{\lambda} \in {[-1, 1] ^ {6 N}}$.

\subsection{Spectral Context Vector}\label{subsec:scv}

Intuitively, we expect different values of $z^{\lambda}$ to be useful for the discrimination of images with different levels of spectral information. For example, on images without significant high-frequency content, SRS values that relate to the reconstruction of high frequencies should provide little value. To capture this information, the network needs access to the context where the values of $z^{\lambda}$ have been computed. To this end, we introduce the Spectral Context Vector (SCV) $z^{C} \in \mathbb{R}^D$ that summarizes the representations of the original image from the $N$ blocks of $\mathcal{G}$. In particular, starting from the projected representations $\bar{z}_n$ defined in \cref{subsec:srs}, we compute their mean and standard deviation over the $L$ different ViT's tokens. Then, we concatenate these two derivative representations of dimensionality $D$ from all the $N$ transformer blocks into the vector $z' \in \mathbb{R}^{N \times 2D}$.

Next, we define a learnable spectral map $C \in \mathbb{R}^{N \times D}$ and two projection functions $\mathcal{P}_1(\cdot): R^{2 D} \rightarrow R^{D}$ and $\mathcal{P}_2(\cdot): R^{D} \rightarrow R^{D}$. Further details regarding the structure of $\mathcal{P}_1$ and $\mathcal{P}_2$ are provided in supplementary material. Using these three components the network learns to construct the SCV by attending to the most useful of the $2D$ feature values for each of the $N$ transformer blocks of $\mathcal{G}$ as following:
\begin{equation}
    C' = \mathcal{P}_2 (\text{softmax}(C) \odot \mathcal{P}_1(z')) \in \mathbb{R}^{N \times D}
\end{equation}
\begin{equation}
    z^{C} = \sum_{n=1}^{N} C_{n}' \in \mathbb{R}^D
\end{equation}

\noindent where $C_{n}'$ denotes the $n$-th row of $C'$. Finally, we concatenate $z^{C}$ and $z^{\lambda}$ into the spectral vector:
\begin{equation}
    z^{S} = [z^{C}; z^{\lambda}] \in \mathbb{R} ^ {D + 6N}.    
\end{equation}

\subsection{Spectral Context Attention}\label{subsec:spa}

In practice, the height and width of the input image $x$ may receive arbitrarily large values, i.e. images of many megapixels. While ViTs in theory can scale to inputs of arbitrary size, their quadratic computational complexity w.r.t the length of their token sequence practically prohibits their use on such large inputs. Similar limitations are faced by convolutional neural networks. 
Thus, a typical approach in most computer vision tasks that rely on the semantic information of the image is to resize it to a fixed size. However, in verification tasks, such as AID, the ability to capture anomalies in subtle details of the image is crucial \cite{corvi2023detection}. So, a resizing operation, that effectively discards the high-frequency information of the image, is problematic. To tackle this issue, we introduce the concept of Spectral Context Attention (SCA), for combining the most discriminative SRS values computed for different patches of the image, according to their respective SCV and a learnable spectral reconstruction importance vector $q \in \mathbb{R}^{D_h}$.    

Initially, we split the original image into $K$ patches of size $h \times w$, forming the patched representation of the image $x^{P} \in \mathbb{R}^{K \times h \times w}$, while $x_{k}$ denotes the $k$-th patch of this sequence. For each patch, we construct its spectral vector $z_{k}^{S}$ following the approach introduced throughout the previous paragraphs. 
Next, following the transformer's attention notation~\cite{vaswani2017attention} we define the weight matrices $W_K, W_V \in \mathbb{R}^{(D+6N) \times D_h}$ and $W_O \in \mathbb{R}^{D_h \times (D+6N)}$. Conveniently, we define $\bar{z}^{S} \in \mathbb{R}^{K \times (D + 6N)}$ as the concatenation of spectral vectors for all $K$ image patches. Then, we fuse them into the image-level spectral vector $\mathbf{z}^{S}$ through the following:
\begin{equation}
    \mathcal{A} = \text{softmax}(\frac{q * {(\bar{z}^{S} * W_{K})}^\intercal}{\sqrt{D_h}}) \in (0, 1)^{1 \times K} 
\end{equation}
\begin{equation}
    \mathbf{z}^{S} = (A * (\bar{z}^{S} * W_{V})) * W_{O} \in \mathbb{R}^{D+6N}
\end{equation}

\noindent The computational complexity of SCA is $O(K)$. 

\noindent\textbf{Classification Head}: For the final prediction $\hat{y} \in (0, 1)$, we use a three-layer MLP on $\mathbf{z}^{S}$, with RELU activations on the first two layers and a sigmoid activation at the final one.  

\subsection{Training Process and Implementation}

We train the network in an end-to-end manner with the objective of minimizing the binary cross-entropy between the predicted $\hat{y}$ and the ground-truth $y$ values for each image, i.e. $\mathcal{L}_{cls} = BCE(\hat{y}, y)$.

\textbf{Augmented Views Training:} To train our network, including the SCA module, under a single training stage, we would require images that comprise several $(h \times w)$ patches. However, this would limit us on the type of training data that we could use, while any resizing or cropping approach would introduce unwanted biases. To this end, we train our network with a fixed number of patches $K_{training}$, that we construct as random views of the input image $x$, using a random augmentation policy. Thus, any  image can be used for training, while this also enables an efficient implementation of the training pipeline. Instead, during inference, we use the actual number of patches $K$ included in each image. 

\textbf{Spectral Model:} As our spectral model of real images $\mathcal{G}$ we use a ViT-B/16  transformer~\cite{dosovitskiy2020image} pre-trained on ImageNet~\cite{deng2009imagenet} by Xie et al.~\cite{xie2022masked}, using a masking radius $r=16$. We keep its weights frozen throughout the training of the components introduced in \cref{subsec:srs,subsec:scv,subsec:spa}. 

\section{Evaluation}
\label{sec:evaluation}

\subsection{Evaluation Setup}

\textbf{Implementation Details:} We train our architecture for 35 epochs, with a learning rate ($lr$) of $5e-4$, cosine decay and five epochs of linear warmup, using the AdamW~\cite{loshchilov2017decoupled} optimizer and batch size of 72. We set $D=1024$, $K_{training}=4$ and, to align with the pretext task, $r=16$. The number of transformer blocks in the ViT we use for $\mathcal{G}$ is $N=12$, its latent dimensionality is $d=768$, while the size of each of the $K$ patches is $h=w=224$ pixels and empirically we set $D_h=1536$. We train our model by applying random resizing, cropping, rotation, Gaussian blur, Gaussian noise and JPEG compression augmentations. The SRS, SCV and SCA are trained on a single Nvidia L40S 48GB GPU.

\begingroup
\setlength{\tabcolsep}{2.75pt}
\renewcommand{\arraystretch}{0.93}
\begin{table*}
    \centering
    \scalebox{0.94}{
    \begin{tabular}{lcccccccccccccc}\toprule
\textbf{Image Size} &\multicolumn{3}{c}{< 0.5 MPixels} &\multicolumn{6}{c}{0.5 - 1.0 MPixels} &\multicolumn{4}{c}{> 1.0 MPixels} & \multirow{2}{*}{\textbf{AVG}} \\
\cmidrule(lr){2-4} \cmidrule(lr){5-10} \cmidrule(lr){11-14}
\textbf{Approach} &\textbf{Glide} &\textbf{SD1.3} &\textbf{SD1.4} &\textbf{Flux} &\textbf{DALLE2} &\textbf{SD2} &\textbf{SDXL} &\textbf{SD3} &\textbf{GigaGAN} &\textbf{MJv5} &\textbf{MJv6.1} &\textbf{DALLE3} &\textbf{Firefly} & \\
\cmidrule(rr){1-1} \cmidrule(rr){2-4} \cmidrule(rr){5-10} \cmidrule(rr){11-14} \cmidrule(rl){15-15}
NPR~\cite{tan2024rethinking} &\cellcolor[HTML]{f1f3f2}72.2 &\cellcolor[HTML]{91d0b1}89.6 &\cellcolor[HTML]{f0dfdd}60.5 &\cellcolor[HTML]{e99791}19.8 &\cellcolor[HTML]{e67c73}3.9 &\cellcolor[HTML]{e78b83}12.5 &\cellcolor[HTML]{e8958d}18.1 &\cellcolor[HTML]{f0dfde}60.6 &\cellcolor[HTML]{b4ddc9}83.2 &\cellcolor[HTML]{e89088}15.3 &\cellcolor[HTML]{e99791}19.8 &\cellcolor[HTML]{67c195}97.1 &\cellcolor[HTML]{ecb7b3}38.0 &45.4 \\
Dire~\cite{wang2023dire} &\cellcolor[HTML]{ebafaa}33.3 &\cellcolor[HTML]{f0dedc}59.9 &\cellcolor[HTML]{f1e0df}61.3 &\cellcolor[HTML]{eec5c1}45.7 &\cellcolor[HTML]{efd0ce}52.2 &\cellcolor[HTML]{f2edec}68.5 &\cellcolor[HTML]{eec7c4}46.9 &\cellcolor[HTML]{eecbc8}49.2 &\cellcolor[HTML]{ecb4b0}36.3 &\cellcolor[HTML]{edbeba}41.9 &\cellcolor[HTML]{eecdca}50.3 &\cellcolor[HTML]{f1e7e6}65.2 &\cellcolor[HTML]{eeccc9}49.9 &50.8 \\
CNNDet.~\cite{wang2020cnn} &\cellcolor[HTML]{f0dddb}59.2 &\cellcolor[HTML]{f0dcdb}59.0 &\cellcolor[HTML]{f0e0df}61.2 &\cellcolor[HTML]{ecbbb6}39.8 &\cellcolor[HTML]{f2f2f2}71.5 &\cellcolor[HTML]{f0dad8}57.5 &\cellcolor[HTML]{f2ebea}67.4 &\cellcolor[HTML]{ebaaa4}30.2 &\cellcolor[HTML]{eaf0ed}73.4 &\cellcolor[HTML]{eecac7}48.8 &\cellcolor[HTML]{f0d8d6}56.7 &\cellcolor[HTML]{e99e97}23.5 &\cellcolor[HTML]{eaf0ed}73.4 &55.5 \\
FreqDet.~\cite{frank2020leveraging} &\cellcolor[HTML]{edc1bd}43.6 &\cellcolor[HTML]{82cba7}92.3 &\cellcolor[HTML]{80caa6}92.7 &\cellcolor[HTML]{ecb5b0}36.5 &\cellcolor[HTML]{eec8c5}47.4 &\cellcolor[HTML]{edbfbb}42.5 &\cellcolor[HTML]{f2e9e9}66.5 &\cellcolor[HTML]{f2efef}69.8 &\cellcolor[HTML]{f1e4e2}63.2 &\cellcolor[HTML]{ecb5b1}36.9 &\cellcolor[HTML]{eaa59f}27.5 &\cellcolor[HTML]{edbfbb}42.2 &\cellcolor[HTML]{c1e1d2}80.9 &57.1 \\
Fusing~\cite{ju2022fusing} &\cellcolor[HTML]{f1e3e2}63.0 &\cellcolor[HTML]{f1e3e2}62.8 &\cellcolor[HTML]{f1e2e1}62.2 &\cellcolor[HTML]{f0dad8}57.5 &\cellcolor[HTML]{d8eae1}76.7 &\cellcolor[HTML]{f2eae9}66.9 &\cellcolor[HTML]{f1e2e0}62.1 &\cellcolor[HTML]{ecb9b4}38.8 &\cellcolor[HTML]{c4e2d3}80.4 &\cellcolor[HTML]{f1e5e4}64.0 &\cellcolor[HTML]{e7efeb}74.0 &\cellcolor[HTML]{eaa19b}25.2 &\cellcolor[HTML]{daeae2}76.3 &62.3 \\
LGrad~\cite{tan2023learning} &\cellcolor[HTML]{d9eae2}76.5 &\cellcolor[HTML]{b9decc}82.4 &\cellcolor[HTML]{b3dcc8}83.4 &\cellcolor[HTML]{e2ede8}74.9 &\cellcolor[HTML]{a7d8c0}85.7 &\cellcolor[HTML]{f0dfde}60.7 &\cellcolor[HTML]{f2f0f0}70.2 &\cellcolor[HTML]{e78b83}12.7 &\cellcolor[HTML]{8fcfb0}89.9 &\cellcolor[HTML]{f2eeee}69.2 &\cellcolor[HTML]{c8e4d6}79.6 &\cellcolor[HTML]{eaa9a4}30.0 &\cellcolor[HTML]{edbebb}42.0 &65.9 \\
UnivFD~\cite{ojha2023towards} &\cellcolor[HTML]{f1e4e3}63.3 &\cellcolor[HTML]{c1e1d2}80.8 &\cellcolor[HTML]{bfe1d0}81.2 &\cellcolor[HTML]{ecb4b0}36.3 &\cellcolor[HTML]{87cdaa}91.4 &\cellcolor[HTML]{aedbc5}84.3 &\cellcolor[HTML]{cfe6db}78.3 &\cellcolor[HTML]{eaa7a1}28.6 &\cellcolor[HTML]{a4d7be}86.2 &\cellcolor[HTML]{f0d9d7}57.1 &\cellcolor[HTML]{f0dfdd}60.5 &\cellcolor[HTML]{ebaba6}31.0 &\cellcolor[HTML]{70c49b}95.5 &67.3 \\
GramNet~\cite{liu2020global} &\cellcolor[HTML]{d0e7db}78.2 &\cellcolor[HTML]{b0dbc6}83.9 &\cellcolor[HTML]{aedbc5}84.3 &\cellcolor[HTML]{cee6da}78.6 &\cellcolor[HTML]{a9d9c2}85.2 &\cellcolor[HTML]{f2eae9}66.7 &\cellcolor[HTML]{d2e8dd}77.8 &\cellcolor[HTML]{e8968f}19.2 &\cellcolor[HTML]{aad9c2}85.0 &\cellcolor[HTML]{f1e5e4}63.8 &\cellcolor[HTML]{abd9c2}84.9 &\cellcolor[HTML]{edc0bc}42.9 &\cellcolor[HTML]{ecb7b3}38.0 &68.4 \\
DeFake~\cite{sha2023fake} &\cellcolor[HTML]{a4d7be}86.1 &\cellcolor[HTML]{f1e5e4}64.2 &\cellcolor[HTML]{f1e4e3}63.6 &\cellcolor[HTML]{8cceae}90.5 &\cellcolor[HTML]{edbdb9}41.4 &\cellcolor[HTML]{f1e9e8}66.2 &\cellcolor[HTML]{efd0ce}52.3 &\cellcolor[HTML]{9cd4b8}87.7 &\cellcolor[HTML]{f3f3f3}71.7 &\cellcolor[HTML]{f2eaea}67.0 &\cellcolor[HTML]{9dd4b9}87.5 &\cellcolor[HTML]{7dc9a3}93.3 &\cellcolor[HTML]{ecbab5}39.4 &70.1 \\
PatchCr.~\cite{zhong2024patchcraft} &\cellcolor[HTML]{cfe6db}78.4 &\cellcolor[HTML]{6fc49a}95.7 &\cellcolor[HTML]{6dc399}96.2 &\cellcolor[HTML]{a0d5bb}86.9 &\cellcolor[HTML]{bcdfce}81.8 &\cellcolor[HTML]{6fc49a}95.7 &\cellcolor[HTML]{6ac297}96.7 &\cellcolor[HTML]{ebb0ab}33.8 &\cellcolor[HTML]{63bf92}98.0 &\cellcolor[HTML]{cce5d9}79.0 &\cellcolor[HTML]{6dc399}96.1 &\cellcolor[HTML]{eaa6a0}28.1 &\cellcolor[HTML]{cbe5d8}79.1 &80.4 \\
DMID~\cite{corvi2023detection} &\cellcolor[HTML]{ecf1ef}73.1 &\cellcolor[HTML]{57bb8a}100.0 &\cellcolor[HTML]{58bc8b}100.0 &\cellcolor[HTML]{67c195}97.2 &\cellcolor[HTML]{efd4d2}54.3 &\cellcolor[HTML]{59bc8c}99.7 &\cellcolor[HTML]{59bc8c}99.6 &\cellcolor[HTML]{f2eceb}67.9 &\cellcolor[HTML]{f2eceb}67.9 &\cellcolor[HTML]{58bc8b}99.9 &\cellcolor[HTML]{76c79f}94.4 &\cellcolor[HTML]{edbdb9}41.3 &\cellcolor[HTML]{8ecfaf}90.2 &83.5 \\
RINE~\cite{koutlis2024leveraging} &\cellcolor[HTML]{70c49b}95.6 &\cellcolor[HTML]{58bc8b}99.9 &\cellcolor[HTML]{58bc8b}99.9 &\cellcolor[HTML]{7ec9a5}93.0 &\cellcolor[HTML]{7ec9a5}93.0 &\cellcolor[HTML]{6ac297}96.6 &\cellcolor[HTML]{5bbd8d}99.3 &\cellcolor[HTML]{ecb9b5}39.1 &\cellcolor[HTML]{7fcaa5}92.9 &\cellcolor[HTML]{6bc398}96.4 &\cellcolor[HTML]{bfe1d0}81.2 &\cellcolor[HTML]{edbeba}41.8 &\cellcolor[HTML]{b6ddca}82.9 &\ul{85.5} \\
\midrule
SPAI (Ours) &\cellcolor[HTML]{8ecfaf}90.2 &\cellcolor[HTML]{5abc8c}99.6 &\cellcolor[HTML]{5abc8c}99.6 &\cellcolor[HTML]{b5ddca}83.0 &\cellcolor[HTML]{88cdab}91.1 &\cellcolor[HTML]{6bc298}96.5 &\cellcolor[HTML]{66c194}97.4 &\cellcolor[HTML]{ddebe4}75.9 &\cellcolor[HTML]{a8d8c1}85.4 &\cellcolor[HTML]{76c69f}94.5 &\cellcolor[HTML]{b0dbc6}84.0 &\cellcolor[HTML]{8ecfaf}90.2 &\cellcolor[HTML]{6dc399}96.0 &\textbf{91.0} \\
\bottomrule
\end{tabular}
    }
    \vspace{-4pt}
    \caption{Comparison against state-of-the-art. Average AUC over 5 sources of real images is reported. Lower values are highlighted in red, while higher values are highlighted in green. Best overall average value is highlighted in bold, while second best is underlined. Our approach generalizes across all the considered generative approaches, even on ones producing imagery of extreme fidelity, such as SD3, where the single method~\cite{sha2023fake} that scores better was required to explicitly train on relevant data.}
    \label{table:sota_comparison}
    \vspace{-12pt}
\end{table*}
\endgroup

\textbf{Datasets:} Following \cite{koutlis2024leveraging,corvi2023detection}, we train our approach on the 180k low resolution images of size $256 \times 256$, from a single latent diffusion model~\cite{rombach2022high}, introduced along with 180k real images by Corvi et al.~\cite{corvi2023detection}. Then, to evaluate our approach, we compose a representative set of images originating from recently introduced generative models. In particular, we employ images from \cite{bammey2023synthbuster} for Glide~\cite{nichol2021glide}, Stable Diffusion 1.3 (SD1.3) and Stable Diffusion 1.4 (SD1.4), representing some early popular text-to-image synthesis approaches, primarily generating low-resolution images, i.e. $< 0.5$ megapixels. Next, we collect from \cite{bammey2023synthbuster} images for DALLE2~\cite{ramesh2022hierarchical}, Stable Diffusion 2 (SD2) and Stable Diffusion XL (SDXL)~\cite{podell2023sdxl}. Also, we employ the online sources of \cite{fluxdata2024,sd3data2024} to collect images generated by Flux and Stable Diffusion 3 (SD3)~\cite{esser2024scaling} respectively. We also include images generated and provided by GigaGAN~\cite{kang2023scaling}, which constitutes one of the few recently proposed GAN-based generators. These approaches represent a group of recent methods, capable of producing images up to 1 megapixel. Finally, we collect images for Midjourney-v5 (MJv5), DALLE3~\cite{betker2023improving} and Firefly from \cite{bammey2023synthbuster}, as well as for Midjourney-v6.1 (MJv6.1) from \cite{mjv61data2024}. For each model we collect 1k images, with the exception of MJv6.1 for which we use 631 publicly available images. Regarding the real images we employ 1k samples from each of RAISE~\cite{dang2015raise} and FODB~\cite{hadwiger2021forchheim}, as representative sources of images captured by DSLR cameras and smartphones respectively. We include another 1k images from the test sets of ImageNet~\cite{deng2009imagenet}, COCO~\cite{lin2014microsoft} and Open Images~\cite{kuznetsova2020open} that represent a general collection of online images, an object-focused database and a collection of high-resolution Web images respectively. These datasets have been collected before the advent of capable generative AI approaches and thus can be safely assumed to be real.

\begingroup
\renewcommand{\arraystretch}{0.85}
\begin{table}
    \centering
    \scalebox{1.0}{
    \begin{tabular}{lcc}
\toprule
\textbf{Backbone} &\textbf{\# Training Data} &\textbf{AUC} \\
\midrule
CLIP ViT-B/16~\cite{oquab2023dinov2} &400 million & 87.6 \\
DINOv2 ViT-B/14~\cite{radford2021learning} &142 million & 87.5 \\
\midrule
MFM ViT-B/16 (Ours)~\cite{xie2022masked} &1.2 million &\textbf{91.0} \\
\bottomrule
\end{tabular}
    }
    \caption{Evaluation of different backbones. Average AUC over 5 sources of real images and 13 generative models is reported. Best  value is highlighted in bold.}
    \vspace{-2pt}
    \label{table:backbones_comparison}
    \vspace{-6pt}
\end{table}
\endgroup

\textbf{Detection Approaches:} We compare our approach against several recent methods proposed to tackle the issue of generalization on generative models not seen during training. We considered 12 approaches with publicly available code and weights. In particular, we employed CNNDetect~\cite{wang2020cnn}, FreqDetect~\cite{frank2020leveraging}, GramNet~\cite{liu2020global}, Fusing~\cite{ju2022fusing}, LGrad~\cite{tan2023learning}, DMID~\cite{corvi2023detection}, UnivFD~\cite{ojha2023towards}, DeFake~\cite{sha2023fake}, DIRE~\cite{wang2023dire}, PatchCraft~\cite{zhong2024patchcraft}, NPR~\cite{tan2024rethinking} and RINE~\cite{koutlis2024leveraging} on their default setups. To facilitate our experiments we used the SIDBench framework~\cite{schinas2024sidbench}. In total, for the 12 considered methods, we evaluated 25 different weight checkpoints. We report best results per method. 

\begin{figure*}
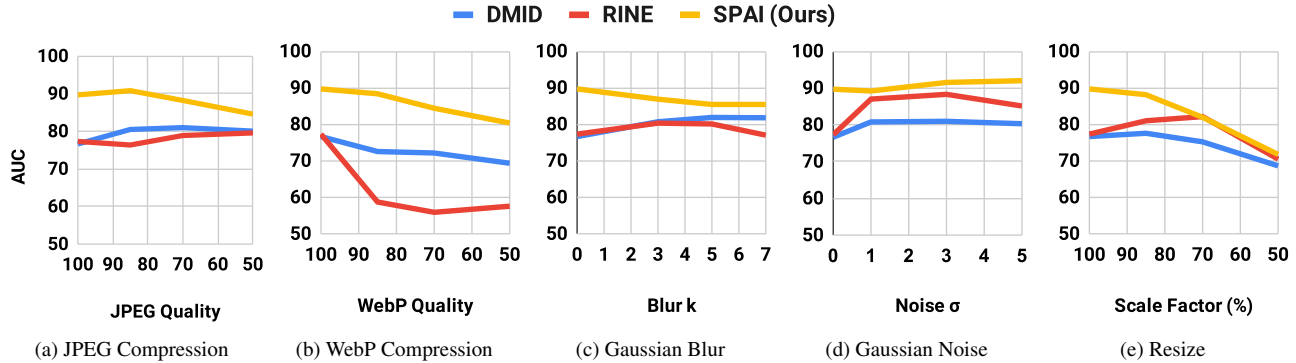

    \centering    \subfloat{\includesvg[width=0.3\textwidth]{figures/perturbations_legend.svg}} \\
    \vspace{-5pt}
    \setcounter{subfigure}{0} 
    \renewcommand{\thesubfigure}{\alph{subfigure}} 
    \subfloat[JPEG Compression]{\includesvg[width=0.211\textwidth]{figures/jpeg_compression.svg}\vspace{-3pt}}
    \subfloat[WebP Compression]{\includesvg[width=0.197\textwidth]{figures/webp_compression.svg}\vspace{-3pt}}
    \subfloat[Gaussian Blur]{\includesvg[width=0.197\textwidth]{figures/blur.svg}\vspace{-3pt}}
    \subfloat[Gaussian Noise]{\includesvg[width=0.197\textwidth]{figures/noise.svg}\vspace{-3pt}}
    \subfloat[Resize]{\includesvg[width=0.197\textwidth]{figures/scale.svg}\vspace{-3pt}}
    \vspace{-6pt}
    \caption{Robustness evaluation on common perturbations. Average AUC is presented over the perturbed versions of two sources of real images from smartphones and DSLR cameras respectively and 13 generative models.}
    \label{fig:robustness}
    \vspace{-10pt}
\end{figure*}

\subsection{Comparison Against State-of-the-Art}

We compare against the considered detectors and report the obtained AUC scores for each of the 13 considered generative models in \cref{table:sota_comparison}, averaged over all the sources of real images. While almost all competitors successfully distinguish AI-generated images from specific generators, in some cases even achieving near-perfect performance, they catastrophically fail to others. For example, while in the case of Firefly images our method achieves the best performance, in the case of DALLE3 images NPR~\cite{tan2024rethinking} performs best. However, NPR underperforms in all the rest generative approaches, overall scoring the worst among the considered detectors. This highlights the inability of detectors that learn to capture specific types of artifacts to generalize to generators that introduce different types of inconsistencies. Instead, our approach, exhibits consistently high detection performance across all the generative models, achieving an absolute  average improvement of 5.5\% over the second best method, even with the later using a much larger ViT-L backbone pre-trained on the 400 million images of CLIP.

\textbf{Robustness Against Online Perturbations:} To understand the robustness of our network against common perturbations encountered online we study five different types of perturbations. In particular, we study the effects of JPEG and WebP compression with quality factors $Q={85, 70, 50}$ as well as Gaussian Blur with kernel size $k={3,5,7}$, Gaussian Noise with standard deviation $\sigma={1,3,5}$ and resizing with scaling factors of ${85,70,50}\%$. To better isolate the effect of each perturbation, in this study we use only the sets of real images originating directly from DSLR and smartphone cameras, i.e. from RAISE and FODB respectively, and present the results in \cref{fig:robustness}. For illustration purposes we limit the results to the two best competitor approaches, as indicated by \cref{table:sota_comparison}. We see that in all cases our approach outperforms the competing methods.

\subsection{Ablation Studies}

\textbf{Pretraining Approach:} We evaluate the importance of the frequency reconstruction pretext task for learning the spectral distribution of real images. To this end, we replace our spectral learned ViT backbone and retrain our network using the CLIP~\cite{radford2021learning} and DINOv2~\cite{oquab2023dinov2}, that have been trained using visual-text alignment and spatial alignment respectively. We report results in \cref{table:backbones_comparison} and find that our ViT leads to clearly higher performance, while being trained only on a fraction of the data used for training CLIP and DINOv2.

\begingroup
\setlength{\tabcolsep}{11pt}
\renewcommand{\arraystretch}{0.91}
\begin{table}
    \centering
    \scalebox{0.94}{
    \begin{tabular}{llc}
\toprule
\multicolumn{2}{l}{\textbf{Ablation}} &\textbf{AUC} \\
\midrule
\midrule
\multicolumn{2}{l}{SPAI (Ours)} &\textbf{91.0} \\
\midrule
\parbox[t]{2mm}{\multirow{6}{*}{\rotatebox[origin=c]{90}{Components}}}
&w/o spectral pretraining & 52.5 \\
&w/o SRS & 71.0 \\
&w/o SCV & 84.9 \\
&w/o SCA & 83.2 \\
&w/o SCA + TenCrop (mean) & 85.3 \\
&w/o SCA + TenCrop (max) & 84.2 \\
\midrule
\parbox[t]{2mm}{\multirow{5}{*}{\rotatebox[origin=c]{90}{Augm.}}}&w/o JPEG compression & 89.1 \\
&w/o distortions & 84.2 \\
&w/o JPEG comp. \& distortions & 87.7 \\
&with WebP compression & 89.3 \\
&with chromatic augmentations & 80.5 \\
\bottomrule
\end{tabular}
    }
    \caption{Ablation studies of the key components and augmentations. Average AUC over 5 sources of real images and 13 generative models is reported. Best  value is highlighted in bold.}
    \label{table:ablations}
\end{table}
\endgroup

\begingroup
\setlength{\tabcolsep}{11pt}
\renewcommand{\arraystretch}{0.89}
\begin{table}
    \centering
    \scalebox{0.93}{
    \begin{tabular}{cc || cc || cc}\toprule
$r$ &\textbf{AUC} &$D$ &\textbf{AUC} &$lr$ &\textbf{AUC} \\
\midrule
8 & 82.4 & 768 & 86.4 & 1e-3 & 84.1 \\
16 &\textbf{91.0} &1024 &\textbf{91.0} &5e-4 &\textbf{91.0} \\
24 & 83.1 & 1536 & 88.8 &1e-4 & 85.6 \\
32 & 85.0 &2048 & 85.3 &5e-5 & 82.7 \\
\bottomrule
\end{tabular}
    }
    \caption{Hyperparameters tuning. Average AUC over 5 sources of real images and 13 generative models is reported. Best  value is highlighted in bold.}
    \label{table:hyperparameter_tuning}
    \vspace{-4pt}
\end{table}
\endgroup

\textbf{Architectural Components:} To verify their significance we remove each of them and retrain the network. We report the respective AUC scores in \cref{table:ablations}. First, we train the backbone along with the rest of the network from scratch, without using pretrained weights. As expected, performance plunges, highlighting the importance of learning the spectral distribution of real images first. Then, we remove SRS, SCV and SCA. For the latter, we evaluate using center cropping of the images as well as ten-crop along with mean and max fusion of the subsequent outputs. In all cases performance drops, highlighting the importance of these components in detecting AI-generated images as out-of-distribution samples to the underlying spectral model.

\textbf{Augmentations:} We evaluate our augmentation policy by removing the noise and blur distortions, the JPEG compression as well as both and report performance in \cref{table:ablations}. Performance drops in all cases, highlighting their value. To evaluate whether further augmentations could benefit performance, we also train our network using WebP compression and chromatic augmentations. However, we found that this led to a decrease in the overall performance. 

\textbf{Hyperparameters Tuning:} We examined different values for the masking radius $r$, the learning rate $lr$ and the size of the latent dimensionality $D$ and we report the corresponding results in \cref{table:hyperparameter_tuning}. We see that $lr=5e-4$ was optimal for our network, as well as $D=1024$. Regarding the masking radius, the optimality of $r=16$ highlights the importance of the alignment between the pretext task and the later use of this underlying spectral model for AID. 

\subsection{Analysis of Spectral Context Attention}

\begin{figure}
    \centering
    \subfloat[6-finger case correctly spotted.] {\includegraphics[width=0.493\columnwidth]{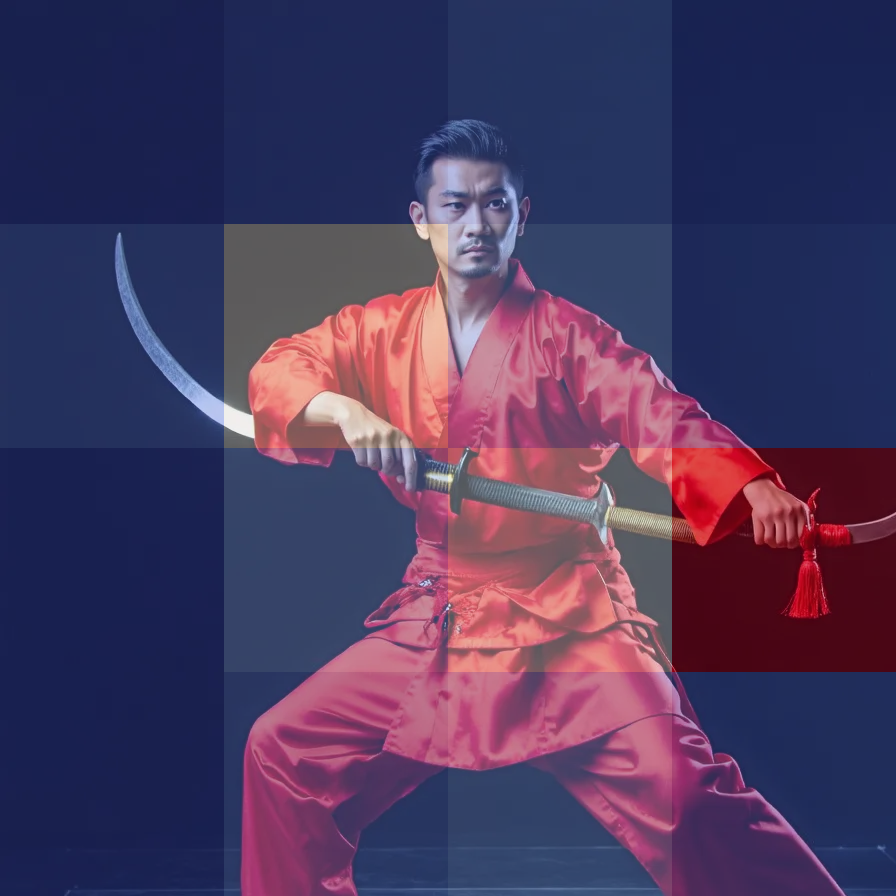}} \hspace{1pt} 
    \subfloat[Attending texture-rich regions.]{\includegraphics[width=0.493\columnwidth]{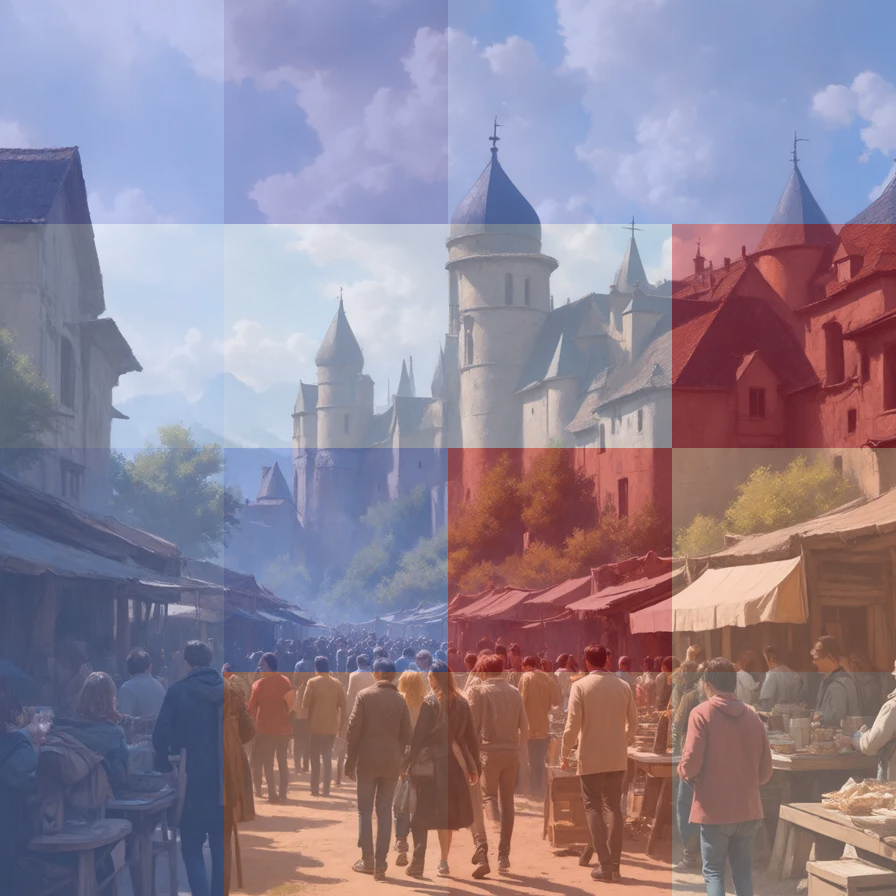}}
    
    \caption{Qualitative analysis of spectral context attention. A cool-warm overlay has been applied on each patch. Red color indicates significant patches for deciding whether the image is AI-generated (high attention values), while blue color indicates irrelevant patches (low attention values). The attention values have been normalized in $[0, 1]$.}
    \label{fig:sca_qualitative}
\end{figure}

We perform a qualitative evaluation of spectral context attention and present two samples in \cref{fig:sca_qualitative}, where we overlay the normalized attention score for each patch of the image. In the first one, depicting an AI-generated image of a human subject possessing a six fingers hand, our attention mechanism identified this patch as significant for the final decision. In the second case, we observe that spectral context attention predicts higher values for patches depicting some complex scenes, in contrast to patches with flat textures, aligning with our original intuition for building it. \looseness=-1 

\subsection{Failure Cases}

While our method, based on spectral learning, exhibits superior generalization performance both w.r.t state-of-the-art approaches as well as on robustness against online perturbations, it still cannot claim to fully solve the AID task. As Dufour et al.~\cite{dufour2024ammeba} highlighted, AI-generated images often appear online in the form of derivative images, i.e. parts of synthetic images appearing in screenshots, memes, or even in photographs of computer screens and in printed material. This intermediate medium, either fully digital, or involving analog parts in the case of external photographs and printings, distorts the spectral distribution of the AI-generated image signal. Thus, the subtle inconsistencies introduced by the generative approaches, become indistinguishable in this later noisy signal. Case in point, we present two AI-generated images in \cref{fig:failures} that have been recently shared online. While our method successfully detects their early-shared copies, it fails to detect two of their derivative copies. This highlights some potential future directions on combining spectral learning with semantic context understanding.

\begin{figure}
  \centering
  \vspace{-9pt}
    \subfloat[Accurately detected AI-generated images shared online. Spectral context attention correctly identifies artifacts in the hand and the texture of the jacket (left), as well as on the face and the ear (right).]{\includesvg[width=0.9999\columnwidth]{figures/detectable_failures.svg}}\\
    \subfloat[Derivative AI-generated images failed to be detected. The distortion introduced by the intermediate medium makes indistinguishable the spectral artifacts that were previously captured correctly.]{\includesvg[width=0.9999\columnwidth]{figures/failures.svg}}
    
  \caption{Failures in detecting derivative AI-generated images. Similar to \cref{fig:sca_qualitative}, spectral context attention is depicted in overlay.}
  \label{fig:failures}
\end{figure}
\section{Conclusion}
\label{sec:conclusion}

In this paper we introduced the key idea that the spectral distribution of real images constitutes an invariant and highly-discriminative pattern for AI-generated image detection and proposed masked spectral learning to model it. To detect generated images as out-of-distributions samples of this model, we introduced the concept of spectral reconstruction similarity for capturing the divergence of the spectral distribution of an image from the one learned by the underlying model. Furthermore, we introduced spectral context attention for detecting subtle spectral inconsistencies in images of any resolution. Using our approach we achieved a significant improvement in state-of-the-art performance, while achieving robustness against common online perturbations. By introducing our work we believe to contribute in reducing the malicious exploitation of generative AI as well as in providing the research community with useful ideas and building blocks for subsequent works in the field.\looseness=-1

\textbf{Limitations:} Detecting subtle spectral inconsistencies in AI-generated images requires this information to pass through the medium transmitting the image. Instead, compression algorithms that discard information not stimulating the human eye as well as noisy digital and analog channels corrupt this useful information. This ultimately constrains what any detector relying on the image signal can detect. Yet, we believe that our work constitutes a contribution towards better approaching these theoretical limitations.

\clearpage

\section*{Acknowledgments}
This work was partly supported by the Horizon Europe projects ELIAS (grant no. 101120237) and vera.ai (grant no. 101070093). The computational resources were granted with the support of GRNET.   

{
    \small
    \bibliographystyle{ieeenat_fullname}
    \bibliography{main}
}

\clearpage
\setcounter{page}{1}
\maketitlesupplementary

\section{Evaluation Setup in Details}

To better facilitate the reproducibility of our results, in this paragraph, we provide a more extensive description of the implementation of our evaluation setup.

\subsection{Projection Operators}

$\mathbf{P}_{n}$: In our architecture we employed $N$ projection operators $\mathbf{P}_{n}: R^d \rightarrow R^D$ for projecting the representations computed by each block of the vision transformer~\cite{dosovitskiy2020image} $\mathcal{G}$ into a space that facilitates the operations of spectral reconstruction similarity. These operators apply a LayerNorm~\cite{lei2016layer} operation to their input and then process the normalized results using a sequence of a linear, a GELU~\cite{hendrycks2016gaussian} and another linear operation. Finally, a LayerNorm operation is applied to their output.

\textbf{$\mathcal{P}_{1}$ and $\mathcal{P}_{2}$}: To build the spectral context vector based on the learnable spectral map $C$ we used the projection operators $\mathcal{P}_1(\cdot): R^{2 D} \rightarrow R^{D}$ and $\mathcal{P}_1(\cdot): R^{D} \rightarrow R^{D}$ that share the same architecture. In particular, they first process input features using two linear layers with GELU~\cite{hendrycks2016gaussian} activations and then normalize their output using  LayerNorm~\cite{lei2016layer}. \looseness=-1

\subsection{Implementation and Training Details}

We implement our approach using PyTorch and train it for 35 epochs using the AdamW~\cite{loshchilov2017decoupled} optimizer. In the first five epochs we perform a linear warmup of the learning rate ($lr$) and increase it from $2.5e-7$ to $5e-4$. Then, from the 5th epoch we apply a per-step cosine decay to ultimately decrease it to $2.5e-7$ at the last step of the 35th epoch. We set the latent dimensionality $D=1024$ and the masking radius $r=16$ according to the hyperparameter tuning procedure that we presented in Sec. 4.3 of the main paper. The vision transformer (ViT)~\cite{dosovitskiy2020image} that we employ in our model of real images $\mathcal{G}$ uses a patch size $p=16$ and includes $N=12$ transformer~\cite{vaswani2017attention} blocks with a latent dimensionality $d=768$. To compute the 2D Discrete Fourier Transform we use the Fast Fourier Transform algorithm. 

We split each image into $K$ patches of size $h=w=224$, while we empirically set the latent dimensionality of the spectral context attention to $D_{h}=1536$. During training we employ $K_{training}=4$ patches. These patches are generated as augmented views of the input image, using random resizing, cropping, rotation, Gaussian blur, Gaussian noise and JPEG compression augmentations. The training of spectral reconstruction similarity, spectral context vector and spectral context attention is performed on a single Nvidia L40S 48GB GPU, using mixed-precision arithmetic, and takes about 50 hours.

Ultimately, we select the best epoch as the one that minimizes loss on the validation split of our training dataset~\cite{corvi2023detection}. However, we noticed that the performance on these validation samples would saturate very quickly, reaching a near-optimal level during the first few epochs, without further indicating whether the model would learn useful patterns. To make validation setup more challenging, without using any external data, we applied once our augmentation policy to the validation data. Then, we used this augmented validation split for selecting the best epoch. We provide an overview of our training and validation data in \cref{table:train_datasets} as well as our test data in \cref{table:test_datasets}. 

\begingroup
\setlength{\tabcolsep}{6pt}
\begin{table}
    \centering
    \scalebox{1.0}{
    \begin{tabular}{lcccc}
\toprule
\textbf{Approach} & \textbf{Split} & \textbf{Gen. \#} & \textbf{Real \#} & \textbf{Size} \\ 
\midrule
Latent Diff.~\cite{corvi2023detection} & train & 180k & 180k & 0.1 MP \\
Latent Diff.~\cite{corvi2023detection} & val. & 20k & 20k & 0.1 MP \\
\bottomrule
\end{tabular}
    }
    \caption{Training and validation data. The average image size in megapixels is presented for each split.}
    \label{table:train_datasets}
    \vspace{-6pt}
\end{table}
\endgroup

\begingroup
\setlength{\tabcolsep}{10pt}
\begin{table}
    \centering
    \scalebox{1.0}{
    \begin{tabular}{c|lcc}
\toprule
\multicolumn{2}{c}{\textbf{Origin}} & \textbf{Images \#} & \textbf{Size} \\ 
\midrule
\parbox[t]{7pt}{\multirow{13}{*}{\rotatebox[origin=c]{90}{AI-Generated}}} & Glide~\cite{bammey2023synthbuster} & 1k & 0.1 MP \\
& SD1.3~\cite{bammey2023synthbuster} & 1k & 0.3 MP\\
& SD1.4~\cite{bammey2023synthbuster} & 1k & 0.3 MP \\
& Flux~\cite{fluxdata2024} & 1k & 0.8 MP \\
& DALLE2~\cite{bammey2023synthbuster} & 1k & 1.0 MP \\
& SD2~\cite{bammey2023synthbuster} & 1k & 1.0 MP \\
& SDXL~\cite{bammey2023synthbuster} & 1k & 1.0 MP \\
& SD3~\cite{sd3data2024} & 1k & 1.0 MP \\
& GigaGAN~\cite{kang2023scaling} & 1k & 1.0 MP \\
& MJv5~\cite{bammey2023synthbuster} & 1k & 1.2 MP \\
& MJv6.1~\cite{mjv61data2024} & 631 & 1.2 MP \\
& DALLE3~\cite{bammey2023synthbuster} & 1k & 1.2 MP \\
& Firefly~\cite{bammey2023synthbuster} & 1k & 4.1 MP \\
\midrule
\parbox[t]{7pt}{\multirow{5}{*}{\rotatebox[origin=c]{90}{Real}}} & ImageNet~\cite{deng2009imagenet} & 1k & 0.2 MP\\ 
& COCO~\cite{lin2014microsoft} & 1k & 0.3 MP \\ 
& OpenImages~\cite{kuznetsova2020open} & 1k & 0.8 MP \\ 
& FODB~\cite{hadwiger2021forchheim} & 1k & 2.8 MP \\ 
& RAISE-1k~\cite{dang2015raise} & 1k & 15 MP\\ 

\bottomrule
\end{tabular}
    }
    \caption{Test data. The average image size in megapixels is presented for each generative model and source of real images. }
    \label{table:test_datasets}
\end{table}
\endgroup

\section{Runtime Analysis}

To evaluate the computational performance of our approach we analyze its runtime using the proposed spectral context attention as well as by solely relying on the scaled dot product self-attention of the vision transformer~\cite{dosovitskiy2020image}. We report the runtime across images of different resolution in \cref{fig:runtime}. As we see, the runtime increases rapidly when relying on the self-attention of quadratic computational complexity, while, due to memory constraints, the maximum image resolution we could process using an Nvidia L40S 48GB GPU was limited to 18 megapixels. Instead, employing spectral context attention enables our approach to scale linearly w.r.t the size of the image, without being limited by the available memory. To better highlight the difference in memory requirements,  we present in \cref{table:memory} the required GPU memory for different image sizes. We see that for a 10 megapixels image self-attention requires almost 18 gigabytes of memory. Instead, when using spectral context attention less than 1 gigabyte of memory is required. Finally, using spectral context attention we managed to analyze with our architecture images of 1 gigapixel, under a single-GPU setup, exceeding by a large margin the size of the images produced by commercial cameras at the time of writing this manuscript. To eliminate any possible inconsistencies due to inefficient implementations, in our runtime analysis we employed the cuda-optimized FlashAttention~\cite{dao2022flashattention}.    

\setlength{\textfloatsep}{12pt}{
\begin{figure}\vspace{-4pt}
  \centering
  \includesvg[width=0.99999\columnwidth]{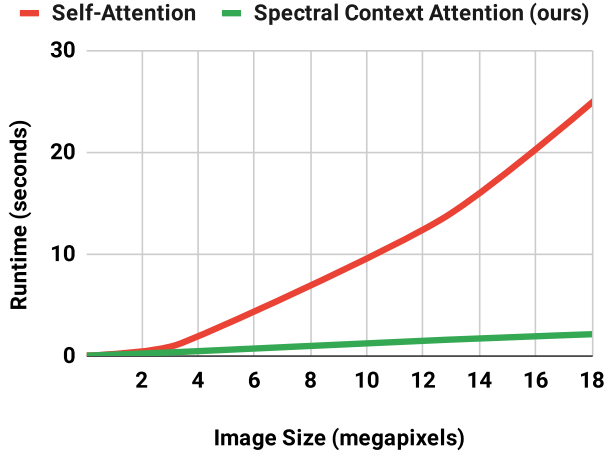}
  \vspace{-22pt}
  \caption{ Runtime comparison between vision transformer's self-attention and spectral context attention for different image resolutions. Lower is better. 18 megapixels was the maximum possible size to scale self-attention due to memory constraints, while spectral context attention did not face similar issues. }
  \label{fig:runtime}
\end{figure}
}

\begingroup
\setlength{\tabcolsep}{4.5pt}
\begin{table}
    \centering
    \scalebox{1.0}{
    \begin{tabular}{lcccc}
\toprule
\textbf{Attention} &\textbf{0.1 MP} &\textbf{1 MP} &\textbf{10 MP} &\textbf{1000 MP} 
\\\midrule
Self-Att. &0.68 GB &2.05 GB &17.7 GB &N/A \\
SCA (ours) &0.62 GB &0.68 GB &0.97 GB & 38.6 GB \\
\bottomrule
\end{tabular}
    }
    \caption{Comparison of the required GPU memory to process images of different size between vision transformer's self-attention and spectral context attention. Using spectral context attention we managed to analyze gigapixel sized images using a single GPU. MP stands for megapixel and GB for gigabyte. Lower is better. }
    \label{table:memory}
\end{table}
\endgroup

\section{Feature Space Analysis}

\begin{figure*}
    \centering
    \begin{minipage}{0.05\textwidth}
    \rotatebox{90}{Stable Diffusion 1.4}
    \end{minipage}%
    \begin{minipage}{0.95\textwidth}
    \subfloat[SCV] {\includegraphics[width=0.27\textwidth]{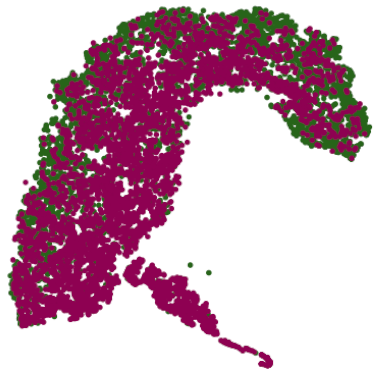}} \hfil 
    \subfloat[SRS] 
    {\includegraphics[width=0.27\textwidth]{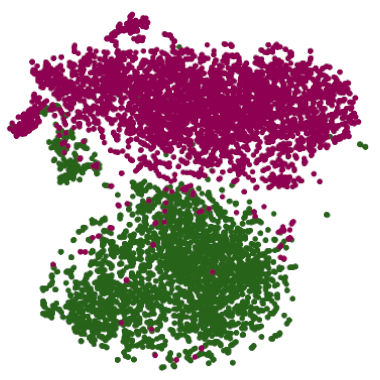}} \hfil 
    \subfloat[SCA] 
    {\includegraphics[width=0.27\textwidth]{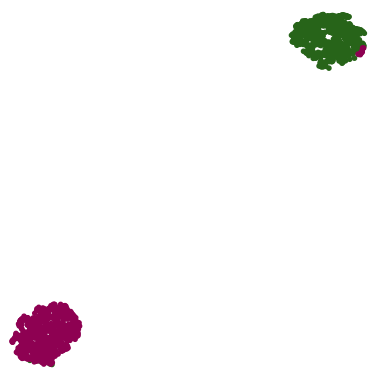}}
    \end{minipage}
    \\
    \vspace{8pt}
    \setcounter{subfigure}{0} 
    \renewcommand{\thesubfigure}{\alph{subfigure}} 
    \begin{minipage}{0.05\textwidth}
    \rotatebox{90}{Stable Diffusion XL}
    \end{minipage}%
    \begin{minipage}{0.95\textwidth}
    \subfloat[SCV] {\includegraphics[width=0.27\textwidth]{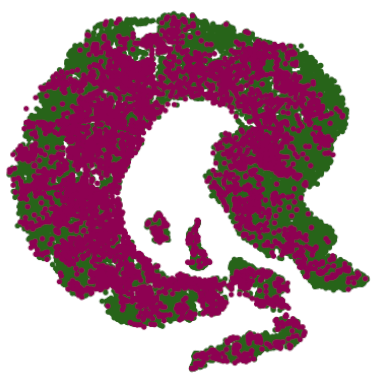}} \hfil 
    \subfloat[SRS] 
    {\includegraphics[width=0.27\textwidth]{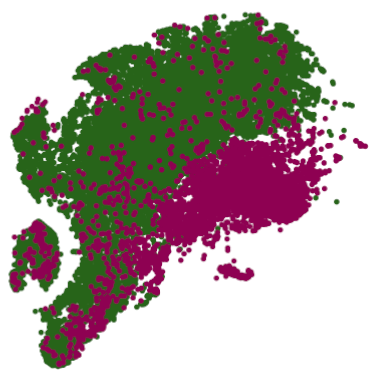}} \hfil 
    \subfloat[SCA] 
    {\includegraphics[width=0.27\textwidth]{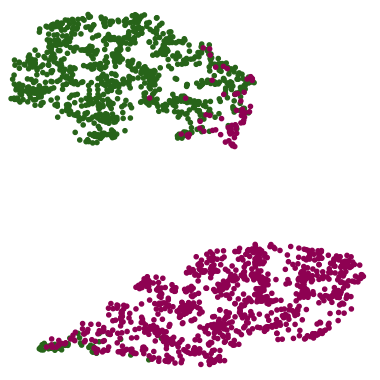}}
    \end{minipage}
    \\
    \vspace{8pt}
    \setcounter{subfigure}{0} 
    \renewcommand{\thesubfigure}{\alph{subfigure}} 
    \begin{minipage}{0.05\textwidth}
    \rotatebox{90}{MidJourney-v5}
    \end{minipage}%
    \begin{minipage}{0.95\textwidth}
    \subfloat[SCV] {\includegraphics[width=0.27\textwidth]{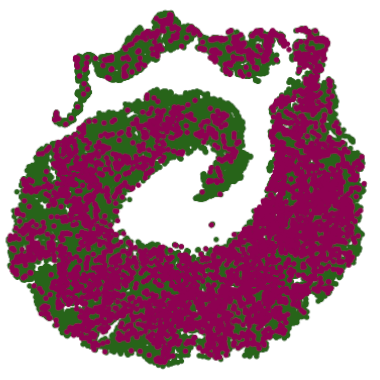}} 
    \hfil 
    \subfloat[SRS] 
    {\includegraphics[width=0.27\textwidth]{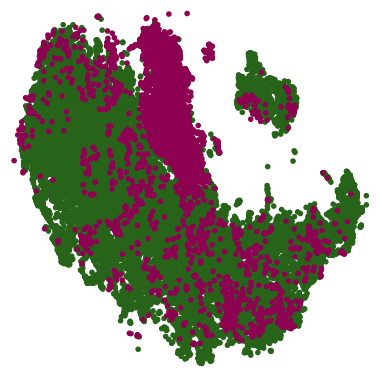}} 
    \hfil 
    \subfloat[SCA] 
    {\includegraphics[width=0.27\textwidth]{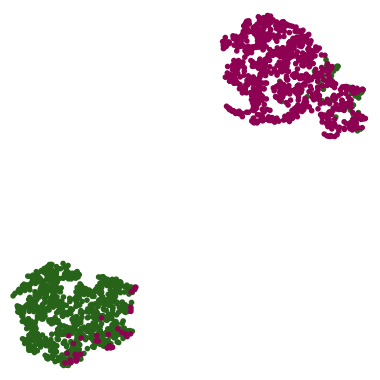}}
    \end{minipage}
    \caption{t-SNE embeddings for the spectral context vector (SCV) $z^C$, the spectral reconstruction similarity (SRS) values $z^{\lambda}$ and the image-level spectral vector $\mathbf{z}^S$ generated by the spectral context attention (SCA) for AI-generated images from three generative models and a common source of real images. Each dot corresponds to the embeddings of different image patches in the case of (a) SCV and (b) SRS and to different images in the case of (c) SCA. Embeddings for AI-generated samples are denoted in green, while for real ones in purple. The spectral context itself cannot discriminate between real and AI-generated samples (a). While spectral reconstruction similarity values provide significant discriminative ability, lots of noise is involved (b). Instead, using the spectral context attention to combine the most discriminative of the SRS values, according to the spectral context of each patch, produces highly separable image-level embeddings (c). \looseness=-1}
    \label{fig:tsne}
    \vspace{-4pt}
\end{figure*}

To study the effect of our key architectural components in the feature space we embed the spectral context vector (SCV) $z^{C}$, the spectral reconstruction similarity (SRS) values $z^{\lambda}$ and the image-level spectral vector $\mathbf{z}^{S}$ generated by spectral context attention (SCA) using t-SNE~\cite{van2008visualizing} and present the results in \cref{fig:tsne}. We embed these three latent representations for AI-generated images of different resolution, originating from Stable Diffusion 1.4, Stable Diffusion XL and MidJourney-v5, while, for illustration purposes, limiting our source of real images to COCO. As we see, the spectral context vector itself provides minimal discriminative capability, highlighting that our approach does not rely upon the context of the images. On the other hand, while SRS values provide significant discriminative capability, there is lots of noise involved, as different SRS values are useful for image patches with different spectral context. Finally, using the SCA to combine the most discriminative of the SRS values according to the spectral context of each patch produces highly separable image-level embeddings, verifying our original intuition for building this mechanism.

\begingroup
\setlength{\tabcolsep}{2.75pt}
\renewcommand{\arraystretch}{1.11}
\begin{table*}
    \centering
    \scalebox{0.94}{
    \begin{tabular}{lcccccccccccccc}\toprule
\textbf{Image Size} &\multicolumn{3}{c}{< 0.5 MPixels} &\multicolumn{6}{c}{0.5 - 1.0 MPixels} &\multicolumn{4}{c}{> 1.0 MPixels} & \textbf{Acc.} \\
\cmidrule(lr){2-4} \cmidrule(lr){5-10} \cmidrule(lr){11-14}
\textbf{Approach} &\textbf{Glide} &\textbf{SD1.3} &\textbf{SD1.4} &\textbf{Flux} &\textbf{DALLE2} &\textbf{SD2} &\textbf{SDXL} &\textbf{SD3} &\textbf{GigaGAN} &\textbf{MJv5} &\textbf{MJv6.1} &\textbf{DALLE3} &\textbf{Firefly} & \textbf{AVG} \\
\cmidrule(rr){1-1} \cmidrule(rr){2-4} \cmidrule(rr){5-10} \cmidrule(rr){11-14} \cmidrule(rl){15-15}
Dire~\cite{wang2023dire} &\cellcolor[HTML]{eaa49e}38.2 &\cellcolor[HTML]{e9f0ec}57.7 &\cellcolor[HTML]{e6efea}58.5 &\cellcolor[HTML]{eec9c6}46.0 &\cellcolor[HTML]{f1e4e3}51.5 &\cellcolor[HTML]{dbebe3}61.7 &\cellcolor[HTML]{efcfcd}47.3 &\cellcolor[HTML]{f0dad8}49.6 &\cellcolor[HTML]{ebb1ac}40.9 &\cellcolor[HTML]{edbfbb}43.8 &\cellcolor[HTML]{f0dedd}50.4 &\cellcolor[HTML]{d8eae1}62.6 &\cellcolor[HTML]{f1e2e0}51.1 &50.7 \\
CNNDet.~\cite{wang2020cnn} &\cellcolor[HTML]{f1e6e5}52.0 &\cellcolor[HTML]{f1e7e6}52.2 &\cellcolor[HTML]{f1e8e7}52.3 &\cellcolor[HTML]{f0d8d6}49.1 &\cellcolor[HTML]{e8efec}58.0 &\cellcolor[HTML]{f1e7e6}52.2 &\cellcolor[HTML]{edf1ef}56.6 &\cellcolor[HTML]{efd0ce}47.5 &\cellcolor[HTML]{d6e9e0}63.0 &\cellcolor[HTML]{f0e0df}50.8 &\cellcolor[HTML]{f1e9e8}52.6 &\cellcolor[HTML]{efd1ce}47.6 &\cellcolor[HTML]{f0f2f1}55.5 &53.0 \\
NPR~\cite{tan2024rethinking} &\cellcolor[HTML]{9bd4b8}79.9 &\cellcolor[HTML]{9bd4b8}79.9 &\cellcolor[HTML]{9dd4b9}79.4 &\cellcolor[HTML]{e67c73}29.9 &\cellcolor[HTML]{e67c73}29.9 &\cellcolor[HTML]{e67c73}30.1 &\cellcolor[HTML]{e67c73}29.9 &\cellcolor[HTML]{9bd4b8}79.9 &\cellcolor[HTML]{9bd4b8}79.9 &\cellcolor[HTML]{e6837a}31.4 &\cellcolor[HTML]{e67c73}29.9 &\cellcolor[HTML]{9bd4b8}79.9 &\cellcolor[HTML]{e67c73}29.9 &53.1 \\
Fusing~\cite{ju2022fusing} &\cellcolor[HTML]{f2f0f0}54.1 &\cellcolor[HTML]{f2eaea}52.9 &\cellcolor[HTML]{f2ebea}53.0 &\cellcolor[HTML]{f0dfdd}50.5 &\cellcolor[HTML]{e3ede8}59.4 &\cellcolor[HTML]{f1e9e8}52.6 &\cellcolor[HTML]{f1e3e2}51.4 &\cellcolor[HTML]{efd4d1}48.2 &\cellcolor[HTML]{d8e9e1}62.6 &\cellcolor[HTML]{f2eceb}53.2 &\cellcolor[HTML]{e6eeea}58.6 &\cellcolor[HTML]{efd4d1}48.2 &\cellcolor[HTML]{eff2f1}55.9 &53.9 \\
FreqDet.~\cite{frank2020leveraging} &\cellcolor[HTML]{eecdca}46.9 &\cellcolor[HTML]{99d3b7}80.4 &\cellcolor[HTML]{99d3b7}80.6 &\cellcolor[HTML]{ecb9b4}42.6 &\cellcolor[HTML]{ecbab5}42.8 &\cellcolor[HTML]{eecac7}46.3 &\cellcolor[HTML]{daeae3}61.8 &\cellcolor[HTML]{cfe6db}65.1 &\cellcolor[HTML]{ddece5}60.9 &\cellcolor[HTML]{edbfbb}44.0 &\cellcolor[HTML]{ecb4af}41.6 &\cellcolor[HTML]{eecdca}46.9 &\cellcolor[HTML]{a5d7bf}77.2 &56.7 \\
UnivFD~\cite{ojha2023towards} &\cellcolor[HTML]{f0e0df}50.8 &\cellcolor[HTML]{c5e3d4}67.9 &\cellcolor[HTML]{c7e3d5}67.4 &\cellcolor[HTML]{eec9c6}45.9 &\cellcolor[HTML]{97d2b5}81.0 &\cellcolor[HTML]{b3dcc8}73.2 &\cellcolor[HTML]{cde6da}65.5 &\cellcolor[HTML]{edc4c1}45.0 &\cellcolor[HTML]{b1dcc7}73.6 &\cellcolor[HTML]{f0dcdb}50.0 &\cellcolor[HTML]{f2efef}53.8 &\cellcolor[HTML]{edc4c0}44.9 &\cellcolor[HTML]{7ec9a5}88.2 &62.1 \\
LGrad~\cite{tan2023learning} &\cellcolor[HTML]{c0e1d1}69.3 &\cellcolor[HTML]{a0d5bb}78.6 &\cellcolor[HTML]{9fd5bb}78.8 &\cellcolor[HTML]{c2e2d2}68.9 &\cellcolor[HTML]{9ed5ba}79.2 &\cellcolor[HTML]{f3f3f3}54.6 &\cellcolor[HTML]{d2e7dd}64.3 &\cellcolor[HTML]{e88f88}34.0 &\cellcolor[HTML]{9ad3b7}80.4 &\cellcolor[HTML]{d5e9df}63.3 &\cellcolor[HTML]{b1dcc7}73.6 &\cellcolor[HTML]{e99f99}37.2 &\cellcolor[HTML]{ecb6b2}42.1 &63.4 \\
GramNet~\cite{liu2020global} &\cellcolor[HTML]{b4ddc9}72.7 &\cellcolor[HTML]{9fd5bb}78.7 &\cellcolor[HTML]{9ed5ba}79.1 &\cellcolor[HTML]{b0dbc6}73.9 &\cellcolor[HTML]{9ad3b7}80.4 &\cellcolor[HTML]{e6efea}58.5 &\cellcolor[HTML]{b6ddca}72.3 &\cellcolor[HTML]{e7887f}32.4 &\cellcolor[HTML]{99d3b7}80.4 &\cellcolor[HTML]{e7efeb}58.2 &\cellcolor[HTML]{9ad3b7}80.3 &\cellcolor[HTML]{e99e97}37.0 &\cellcolor[HTML]{e8938c}34.8 &64.5 \\
DeFake~\cite{sha2023fake} &\cellcolor[HTML]{a8d8c1}76.3 &\cellcolor[HTML]{e3ede8}59.5 &\cellcolor[HTML]{e4eee9}59.0 &\cellcolor[HTML]{9cd4b9}79.6 &\cellcolor[HTML]{eec5c2}45.2 &\cellcolor[HTML]{dbebe3}61.7 &\cellcolor[HTML]{f1e4e3}51.7 &\cellcolor[HTML]{a1d6bc}78.3 &\cellcolor[HTML]{c8e4d6}67.0 &\cellcolor[HTML]{d9eae2}62.2 &\cellcolor[HTML]{a5d7bf}77.1 &\cellcolor[HTML]{96d2b4}81.5 &\cellcolor[HTML]{edc0bc}44.0 &64.8 \\
DMID~\cite{corvi2023detection} &\cellcolor[HTML]{f1e8e7}52.4 &\cellcolor[HTML]{57bb8a}99.3 &\cellcolor[HTML]{57bb8a}99.3 &\cellcolor[HTML]{94d1b3}82.1 &\cellcolor[HTML]{f0dad8}49.5 &\cellcolor[HTML]{5dbd8e}97.9 &\cellcolor[HTML]{61bf91}96.5 &\cellcolor[HTML]{e8efec}57.9 &\cellcolor[HTML]{f2f1f1}54.3 &\cellcolor[HTML]{5bbd8d}98.4 &\cellcolor[HTML]{a2d6bc}78.1 &\cellcolor[HTML]{f0d9d7}49.3 &\cellcolor[HTML]{e9f0ed}57.5 &74.8 \\
PatchCr.~\cite{zhong2024patchcraft} &\cellcolor[HTML]{bfe1d0}69.6 &\cellcolor[HTML]{84cba8}86.6 &\cellcolor[HTML]{83cba8}86.7 &\cellcolor[HTML]{9cd4b9}79.7 &\cellcolor[HTML]{b8decb}71.7 &\cellcolor[HTML]{83cba8}86.8 &\cellcolor[HTML]{7ac8a2}89.4 &\cellcolor[HTML]{ebada8}40.2 &\cellcolor[HTML]{7bc8a3}89.0 &\cellcolor[HTML]{b5ddc9}72.6 &\cellcolor[HTML]{7fcaa5}87.9 &\cellcolor[HTML]{ebafaa}40.6 &\cellcolor[HTML]{b5ddca}72.5 &74.9 \\
RINE~\cite{koutlis2024leveraging} &\cellcolor[HTML]{7dc9a4}88.5 &\cellcolor[HTML]{61bf91}96.6 &\cellcolor[HTML]{62bf91}96.4 &\cellcolor[HTML]{8dcfae}84.1 &\cellcolor[HTML]{94d1b3}82.0 &\cellcolor[HTML]{79c8a1}89.6 &\cellcolor[HTML]{65c094}95.5 &\cellcolor[HTML]{efcecb}47.0 &\cellcolor[HTML]{8fd0b0}83.3 &\cellcolor[HTML]{78c7a0}90.1 &\cellcolor[HTML]{c0e1d1}69.2 &\cellcolor[HTML]{efd1cf}47.6 &\cellcolor[HTML]{c8e4d6}67.0 &\underline{79.8} \\
\midrule
SPAI (Ours) &\cellcolor[HTML]{96d2b5}81.3 &\cellcolor[HTML]{70c49b}92.2 &\cellcolor[HTML]{70c49b}92.3 &\cellcolor[HTML]{b7decb}71.9 &\cellcolor[HTML]{8fcfb0}83.4 &\cellcolor[HTML]{7cc9a3}88.8 &\cellcolor[HTML]{76c79f}90.5 &\cellcolor[HTML]{dcebe4}61.3 &\cellcolor[HTML]{a8d8c1}76.2 &\cellcolor[HTML]{82cba7}87.2 &\cellcolor[HTML]{afdbc5}74.3 &\cellcolor[HTML]{98d3b6}80.8 &\cellcolor[HTML]{77c7a0}90.3 &\textbf{82.3} \\
\bottomrule
\end{tabular}
    }
    \caption{Comparison against state-of-the-art. Average accuracy over 5 sources of real images is reported. Lower values are highlighted in red, while higher values are highlighted in green. Best overall average value is highlighted in bold, while second best is underlined.}
    \label{table:sota_bacc}
\end{table*}
\endgroup

\begingroup
\setlength{\tabcolsep}{2.75pt}
\renewcommand{\arraystretch}{1.11}
\begin{table*}
    \centering
    \scalebox{0.94}{
    \begin{tabular}{lcccccccccccccc}
\toprule
\textbf{Image Size} &\multicolumn{3}{c}{< 0.5 MPixels} &\multicolumn{6}{c}{0.5 - 1.0 MPixels} &\multicolumn{4}{c}{> 1.0 MPixels} & \textbf{AP} \\
\cmidrule(lr){2-4} \cmidrule(lr){5-10} \cmidrule(lr){11-14}
\textbf{Approach} &\textbf{Glide} &\textbf{SD1.3} &\textbf{SD1.4} &\textbf{Flux} &\textbf{DALLE2} &\textbf{SD2} &\textbf{SDXL} &\textbf{SD3} &\textbf{GigaGAN} &\textbf{MJv5} &\textbf{MJv6.1} &\textbf{DALLE3} &\textbf{Firefly} & \textbf{AVG} \\
\cmidrule(rr){1-1} \cmidrule(rr){2-4} \cmidrule(rr){5-10} \cmidrule(rr){11-14} \cmidrule(rl){15-15}
Dire~\cite{wang2023dire} &\cellcolor[HTML]{ec9c95}39.5 &\cellcolor[HTML]{f7d9d6}56.6 &\cellcolor[HTML]{f8dcd9}57.4 &\cellcolor[HTML]{f0b0ab}45.2 &\cellcolor[HTML]{f4c9c5}52.2 &\cellcolor[HTML]{f0f9f5}70.2 &\cellcolor[HTML]{f1bab5}47.8 &\cellcolor[HTML]{f1bab6}48.0 &\cellcolor[HTML]{eda49e}41.7 &\cellcolor[HTML]{efaca7}44.1 &\cellcolor[HTML]{ec9e97}40.1 &\cellcolor[HTML]{f9e4e2}59.7 &\cellcolor[HTML]{f2bcb8}48.6 &50.1 \\
CNNDet.~\cite{wang2020cnn} &\cellcolor[HTML]{f8dedc}58.0 &\cellcolor[HTML]{f8ddda}57.6 &\cellcolor[HTML]{f9e3e1}59.4 &\cellcolor[HTML]{efaca6}44.1 &\cellcolor[HTML]{edf8f2}70.8 &\cellcolor[HTML]{f8dbd9}57.2 &\cellcolor[HTML]{fefefe}67.0 &\cellcolor[HTML]{ea968e}37.8 &\cellcolor[HTML]{d4eee1}75.6 &\cellcolor[HTML]{f3c3bf}50.6 &\cellcolor[HTML]{f1b7b2}47.2 &\cellcolor[HTML]{e98d86}35.5 &\cellcolor[HTML]{f5fbf8}69.2 &56.1 \\
FreqDet.~\cite{frank2020leveraging} &\cellcolor[HTML]{f0b4af}46.3 &\cellcolor[HTML]{88cfac}90.6 &\cellcolor[HTML]{86ceab}90.8 &\cellcolor[HTML]{eea7a1}42.6 &\cellcolor[HTML]{f0b0ab}45.2 &\cellcolor[HTML]{f1b5b0}46.7 &\cellcolor[HTML]{fae7e5}60.5 &\cellcolor[HTML]{fdf6f5}64.7 &\cellcolor[HTML]{fcf2f1}63.7 &\cellcolor[HTML]{eeaba5}43.6 &\cellcolor[HTML]{e67c73}30.5 &\cellcolor[HTML]{f1b8b4}47.5 &\cellcolor[HTML]{e8f6ef}71.7 &57.3 \\
NPR~\cite{tan2024rethinking} &\cellcolor[HTML]{d5eee2}75.4 &\cellcolor[HTML]{97d5b7}87.6 &\cellcolor[HTML]{e5f5ed}72.3 &\cellcolor[HTML]{ec9b94}39.3 &\cellcolor[HTML]{e67f76}31.3 &\cellcolor[HTML]{e98d85}35.2 &\cellcolor[HTML]{eb9790}38.3 &\cellcolor[HTML]{e5f5ed}72.4 &\cellcolor[HTML]{a9ddc3}84.0 &\cellcolor[HTML]{eb9790}38.2 &\cellcolor[HTML]{e67d74}31.0 &\cellcolor[HTML]{65c194}97.4 &\cellcolor[HTML]{f3c3bf}50.4 &57.9 \\
Fusing~\cite{ju2022fusing} &\cellcolor[HTML]{fdf6f5}64.7 &\cellcolor[HTML]{fcf2f1}63.6 &\cellcolor[HTML]{fcf0ef}63.0 &\cellcolor[HTML]{f8ddda}57.7 &\cellcolor[HTML]{ccebdc}77.1 &\cellcolor[HTML]{fdf9f9}65.6 &\cellcolor[HTML]{fbeceb}61.9 &\cellcolor[HTML]{eda59f}42.1 &\cellcolor[HTML]{bae3cf}80.7 &\cellcolor[HTML]{fdf6f5}64.6 &\cellcolor[HTML]{fdfefe}67.6 &\cellcolor[HTML]{ea918a}36.6 &\cellcolor[HTML]{dcf1e7}74.0 &63.0 \\
LGrad~\cite{tan2023learning} &\cellcolor[HTML]{bde5d1}80.1 &\cellcolor[HTML]{ceebdd}76.8 &\cellcolor[HTML]{c6e8d7}78.4 &\cellcolor[HTML]{e8f6ef}71.7 &\cellcolor[HTML]{aedec6}83.1 &\cellcolor[HTML]{f9e3e1}59.3 &\cellcolor[HTML]{f5fbf8}69.2 &\cellcolor[HTML]{e7837b}32.6 &\cellcolor[HTML]{8cd1af}89.7 &67.2 &\cellcolor[HTML]{f4fbf8}69.3 &\cellcolor[HTML]{eb9890}38.3 &\cellcolor[HTML]{efaea8}44.6 &66.2 \\
GramNet~\cite{liu2020global} &\cellcolor[HTML]{d8efe4}74.9 &\cellcolor[HTML]{c1e6d4}79.4 &\cellcolor[HTML]{bfe5d3}79.8 &\cellcolor[HTML]{d9f0e5}74.6 &\cellcolor[HTML]{bce4d1}80.3 &\cellcolor[HTML]{fdf8f7}65.2 &\cellcolor[HTML]{dcf1e7}74.0 &\cellcolor[HTML]{e99089}36.2 &\cellcolor[HTML]{bde5d1}80.1 &\cellcolor[HTML]{fcf2f1}63.5 &\cellcolor[HTML]{e0f3e9}73.3 &\cellcolor[HTML]{f1bab6}48.0 &\cellcolor[HTML]{f0b1ac}45.5 &67.3 \\
UnivFD~\cite{ojha2023towards} &\cellcolor[HTML]{fbeeed}62.6 &\cellcolor[HTML]{b3e1ca}82.1 &\cellcolor[HTML]{b4e1cb}81.9 &\cellcolor[HTML]{eea9a4}43.3 &\cellcolor[HTML]{84cda9}91.4 &\cellcolor[HTML]{9ed8bc}86.1 &\cellcolor[HTML]{bee5d2}79.9 &\cellcolor[HTML]{eb9992}38.6 &\cellcolor[HTML]{9ad6b9}87.0 &\cellcolor[HTML]{f8dddb}57.9 &\cellcolor[HTML]{f5ccc8}52.9 &\cellcolor[HTML]{ec9c95}39.6 &\cellcolor[HTML]{6dc499}95.9 &69.2 \\
DeFake~\cite{sha2023fake} &\cellcolor[HTML]{9ad6b9}87.0 &\cellcolor[HTML]{fcf0ef}63.0 &\cellcolor[HTML]{fbeeed}62.5 &\cellcolor[HTML]{86ceab}90.8 &\cellcolor[HTML]{efafaa}44.9 &\cellcolor[HTML]{fefdfd}66.7 &\cellcolor[HTML]{f6d2cf}54.7 &\cellcolor[HTML]{98d6b7}87.4 &\cellcolor[HTML]{ddf1e7}73.9 &\cellcolor[HTML]{fefdfd}66.6 &\cellcolor[HTML]{aaddc4}83.8 &\cellcolor[HTML]{7acaa3}93.2 &\cellcolor[HTML]{eea8a2}42.9 &70.6 \\
PatchCr.~\cite{zhong2024patchcraft} &\cellcolor[HTML]{bfe5d3}79.7 &\cellcolor[HTML]{6ac398}96.3 &\cellcolor[HTML]{68c296}96.7 &\cellcolor[HTML]{9bd7b9}86.8 &\cellcolor[HTML]{bde5d1}80.0 &\cellcolor[HTML]{6bc498}96.1 &\cellcolor[HTML]{70c59b}95.3 &\cellcolor[HTML]{eda39d}41.5 &\cellcolor[HTML]{63c093}97.7 &\cellcolor[HTML]{b8e3ce}81.1 &\cellcolor[HTML]{76c8a0}94.1 &\cellcolor[HTML]{eb9a93}39.0 &\cellcolor[HTML]{cbeadb}77.5 &81.7 \\
DMID~\cite{corvi2023detection} &\cellcolor[HTML]{e9f6f0}71.5 &\cellcolor[HTML]{57bb8a}100.0 &\cellcolor[HTML]{58bc8b}100.0 &\cellcolor[HTML]{66c194}97.2 &\cellcolor[HTML]{f6d3d0}54.9 &\cellcolor[HTML]{59bc8b}99.7 &\cellcolor[HTML]{59bc8c}99.7 &\cellcolor[HTML]{e8f6ef}71.8 &\cellcolor[HTML]{f0f9f4}70.2 &\cellcolor[HTML]{58bc8b}99.9 &\cellcolor[HTML]{7acaa3}93.2 &\cellcolor[HTML]{f0b0ab}45.3 &\cellcolor[HTML]{96d5b6}87.7 &83.9 \\
RINE~\cite{koutlis2024leveraging} &\cellcolor[HTML]{6fc59b}95.5 &\cellcolor[HTML]{58bc8b}99.9 &\cellcolor[HTML]{58bc8b}99.9 &\cellcolor[HTML]{76c8a0}94.0 &\cellcolor[HTML]{7acaa3}93.2 &\cellcolor[HTML]{68c296}96.8 &\cellcolor[HTML]{5abd8d}99.4 &\cellcolor[HTML]{f1b6b1}46.9 &\cellcolor[HTML]{79c9a2}93.4 &\cellcolor[HTML]{67c295}97.0 &\cellcolor[HTML]{c8e9d9}78.0 &\cellcolor[HTML]{f2bcb8}48.6 &\cellcolor[HTML]{aedec7}83.1 &\underline{86.6} \\
\midrule
SPAI (Ours) &\cellcolor[HTML]{86ceab}90.9 &\cellcolor[HTML]{5bbd8d}99.3 &\cellcolor[HTML]{5bbd8d}99.4 &\cellcolor[HTML]{b3e1ca}82.0 &\cellcolor[HTML]{87cfac}90.7 &\cellcolor[HTML]{68c296}96.8 &\cellcolor[HTML]{66c195}97.2 &\cellcolor[HTML]{e4f5ed}72.4 &\cellcolor[HTML]{a6dbc1}84.7 &\cellcolor[HTML]{71c69c}95.1 &\cellcolor[HTML]{c2e7d5}79.1 &\cellcolor[HTML]{90d3b2}88.8 &\cellcolor[HTML]{72c69d}94.9 &\textbf{90.1} \\
\bottomrule
\end{tabular}
    }
    \caption{Comparison against state-of-the-art. Average precision over 5 sources of real images is reported. Lower values are highlighted in red, while higher values are highlighted in green. Best overall average value is highlighted in bold, while second best is underlined.}
    \label{table:sota_ap}
    \vspace{-6pt}
\end{table*}
\endgroup

\section{Additional Metrics}

To study the calibration of our approach with respect to the state-of-the-art detectors as well as to facilitate comparison across popular metrics in the field, we expand the analysis of the main paper by computing the balanced accuracy on the $0.5$ threshold and the average precision metrics. We report the results across our test set of 5 sources of real images and 13 generative models in \cref{table:sota_bacc} and \cref{table:sota_ap} respectively. As we see in the former, our approach, based on spectral learning, not only provides significant discriminative ability, but is also better calibrated around the common $0.5$ threshold, providing an absolute increase of 2.5\% in balanced accuracy. Moreover, SPAI achieves an absolute increase of 3.5\% in terms of average precision, reinstating its superior discriminative ability.

\section{Ethical Considerations}

Introducing an approach for distinguishing AI-generated content from real one intends to prevent the malicious exploitation of generative AI. However, any detection method, will inevitably fail to correctly predict some cases, allowing malicious actors to exploit such results to either promote generated content or discredit real one. Yet, we believe that the improved generalization performance of our approach across several generative models as well as its superior robustness against several common attacks, ultimately decrease the potential of exploitation.

\section{Qualitative Evaluation}

\begin{figure*}
    \centering
    \subfloat[Detection: 100\%] {\includegraphics[width=0.243\textwidth]{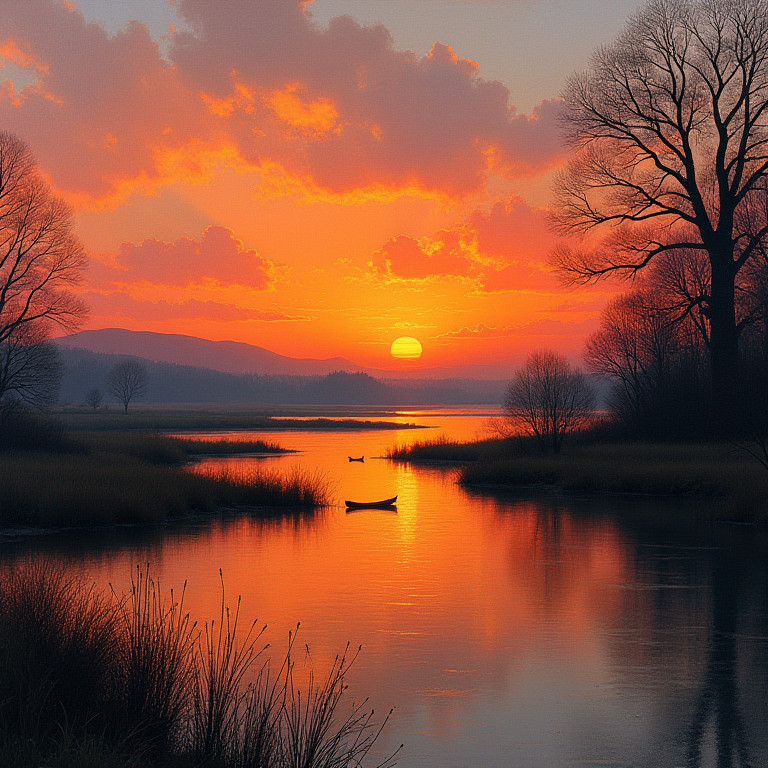}} \hspace{1pt} 
    \subfloat[Detection: 100\%] 
    {\includegraphics[width=0.243\textwidth]{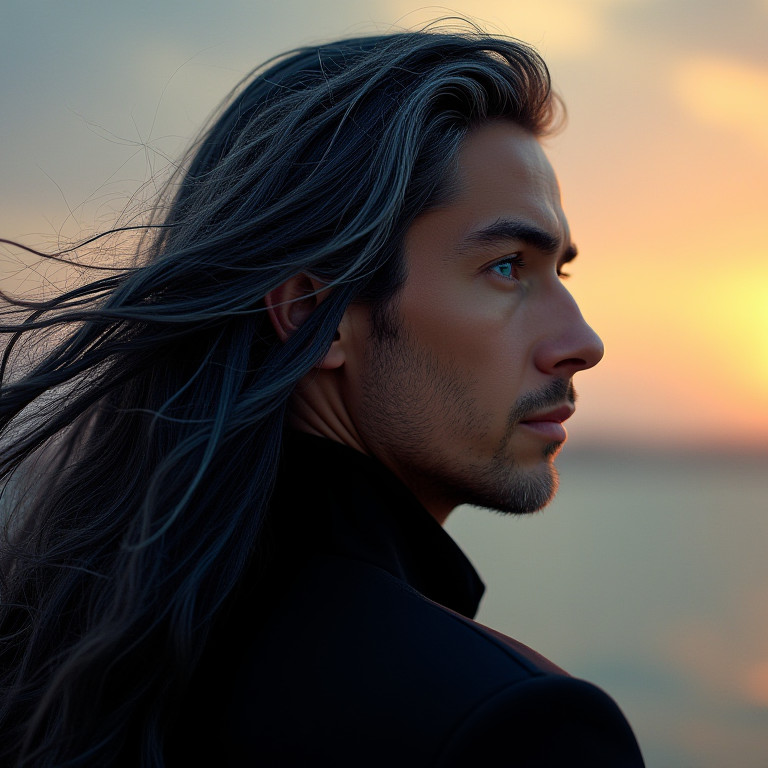}} \hspace{1pt} 
    \subfloat[Detection: 100\%] 
    {\includegraphics[width=0.243\textwidth]{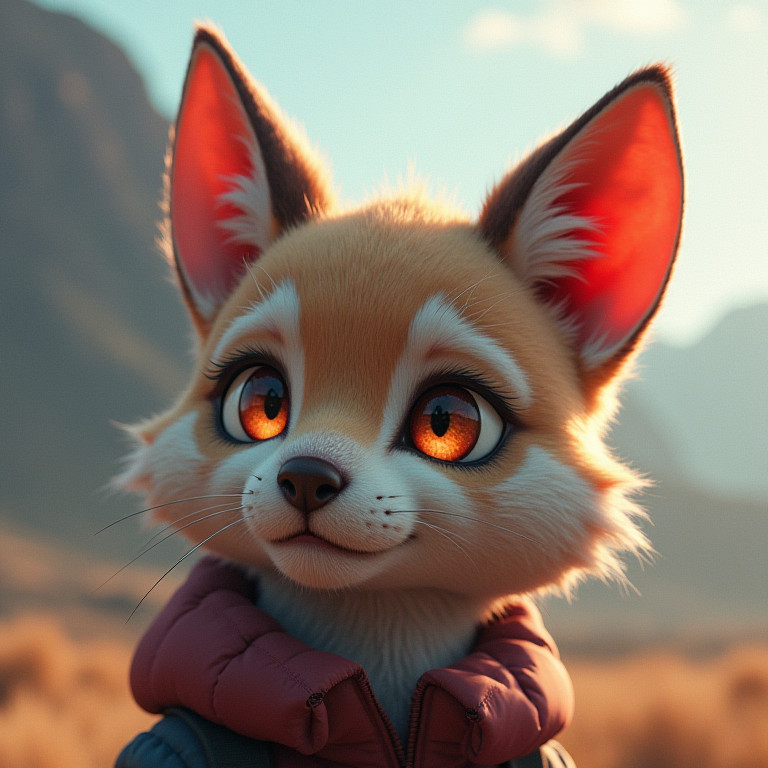}} \hspace{1pt} 
    \subfloat[Detection: 99\%] 
    {\includegraphics[width=0.243\textwidth]{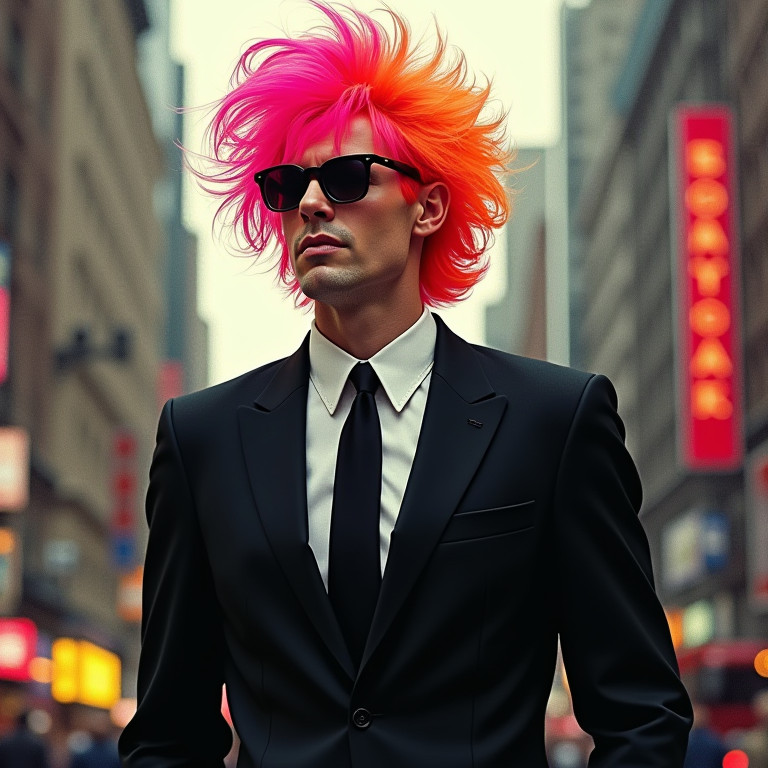}} 
    \caption{Accurately detected Flux images. For illustration purposes cropped to a square aspect ratio.}
    \label{fig:qualitative_flux}
\end{figure*}

\begin{figure*}
    \centering
    \subfloat[Detection: 100\%] {\includegraphics[width=0.243\textwidth]{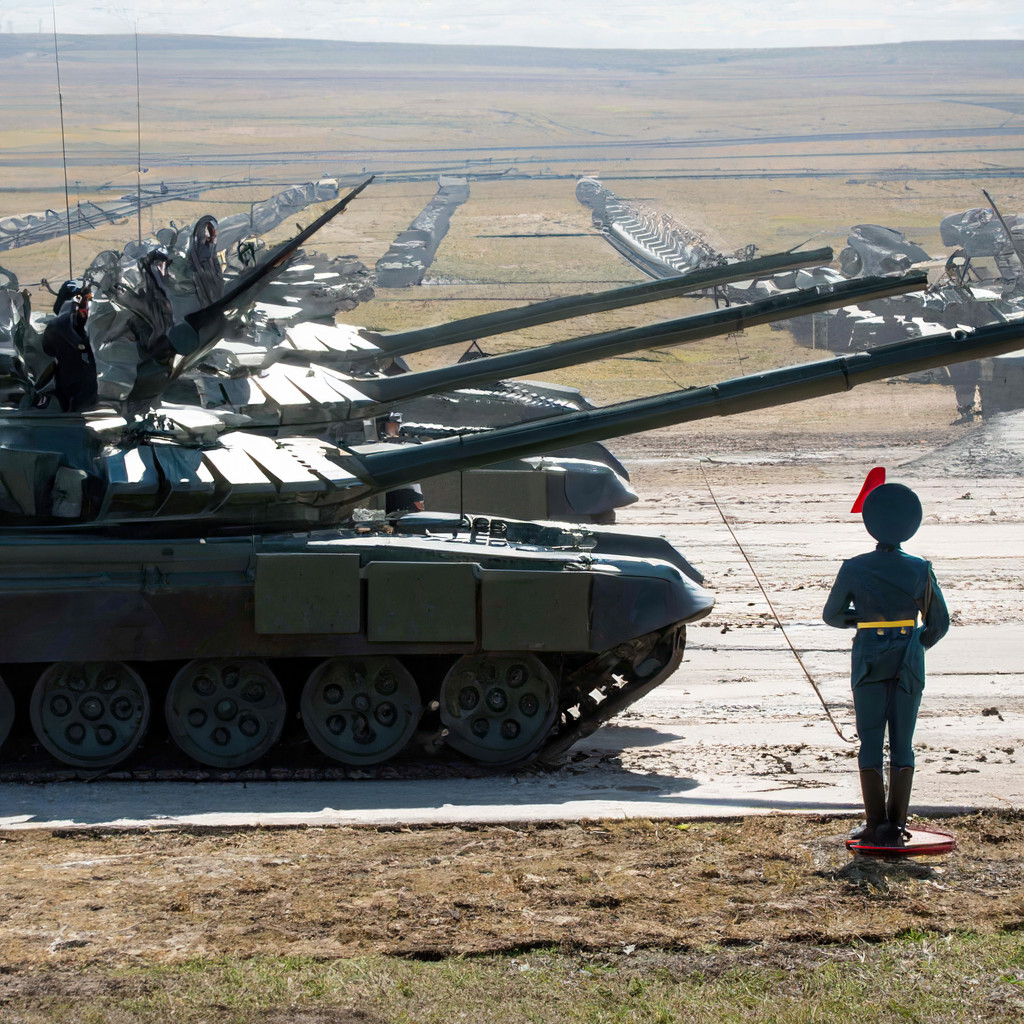}} \hspace{1pt} 
    \subfloat[Detection: 100\%] 
    {\includegraphics[width=0.243\textwidth]{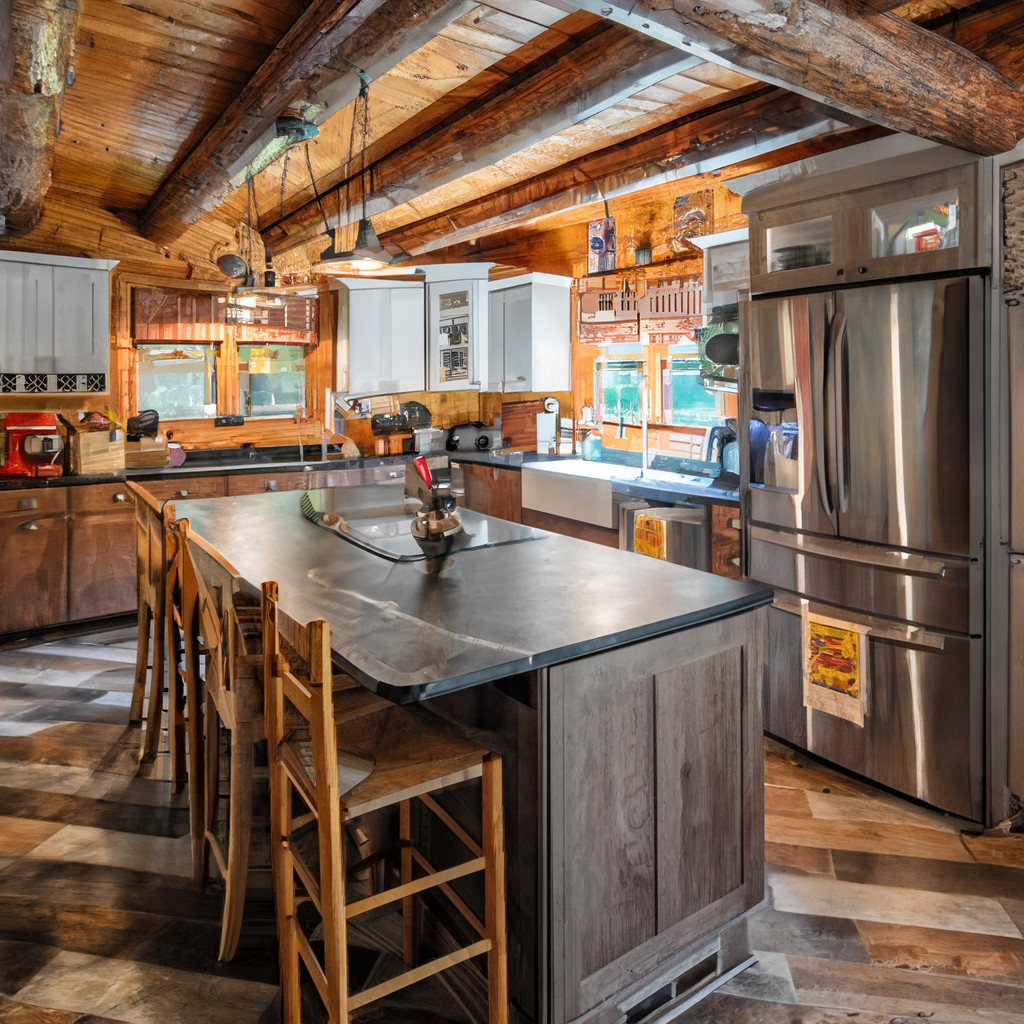}} \hspace{1pt} 
    \subfloat[Detection: 100\%] 
    {\includegraphics[width=0.243\textwidth]{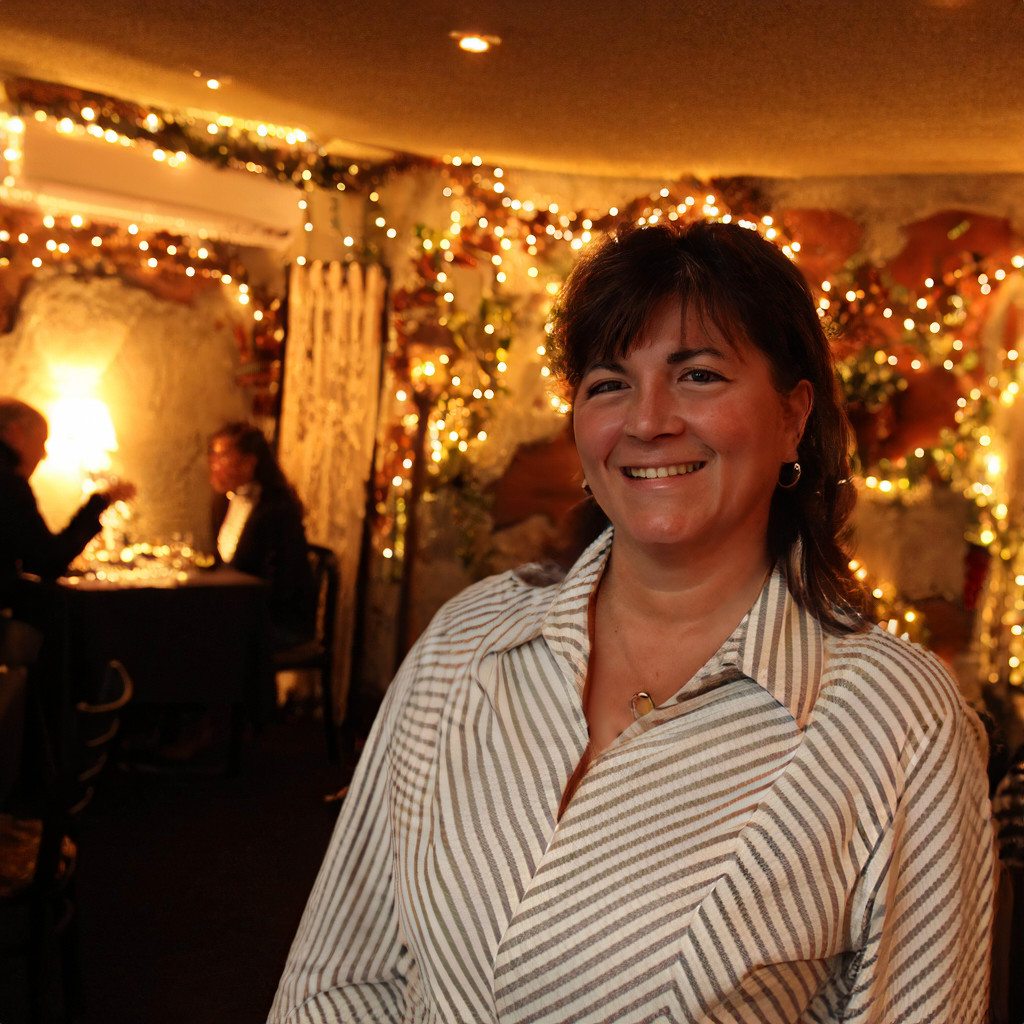}} \hspace{1pt} 
    \subfloat[Detection: 100\%] 
    {\includegraphics[width=0.243\textwidth]{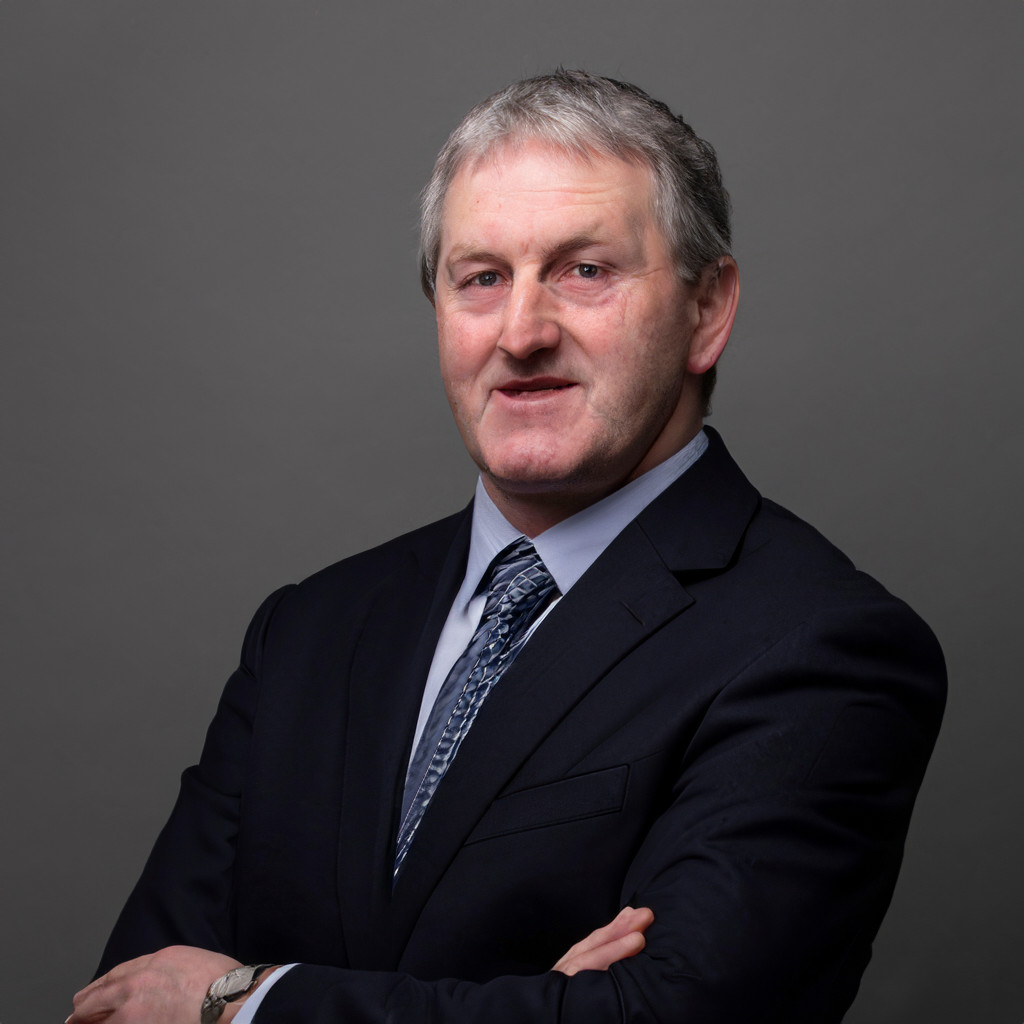}} 
    \caption{Accurately detected GigaGAN images. For illustration purposes cropped to a square aspect ratio.}
    \label{fig:qualitative_gigagan}
\end{figure*}

\begin{figure*}
    \centering
    \subfloat[Detection: 75\%] {\includegraphics[width=0.243\textwidth]{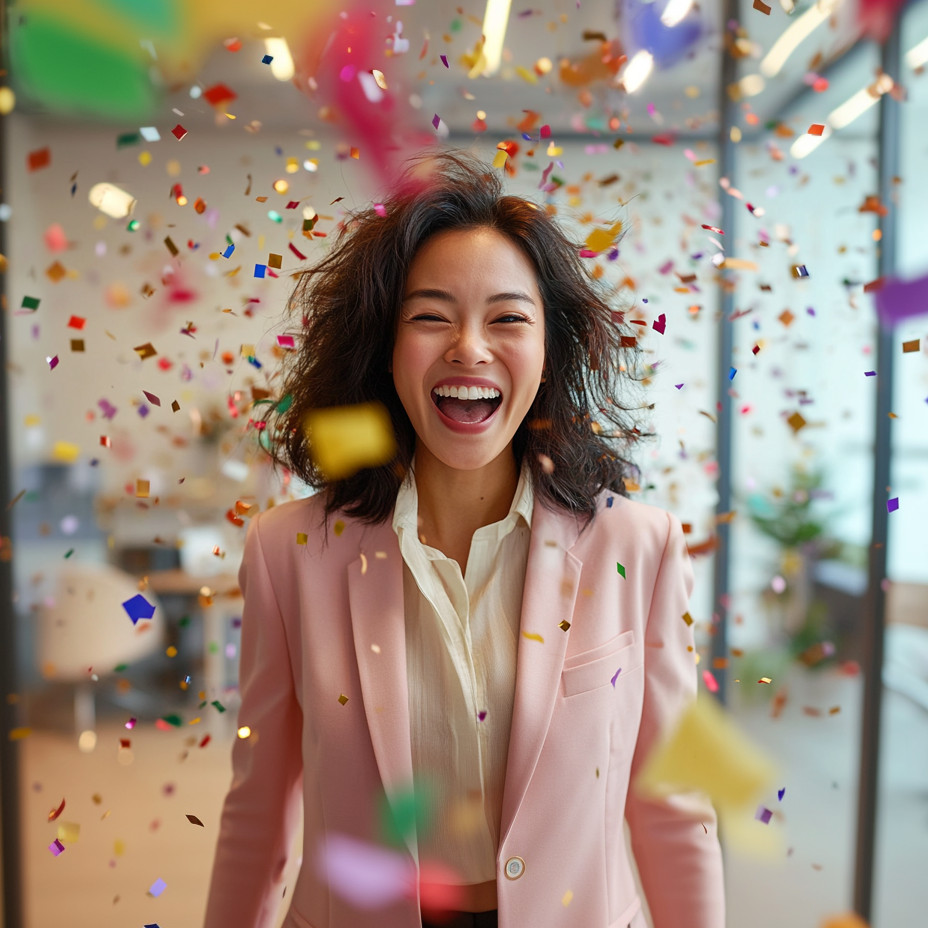}} \hspace{1pt} 
    \subfloat[Detection: 97\%] 
    {\includegraphics[width=0.243\textwidth]{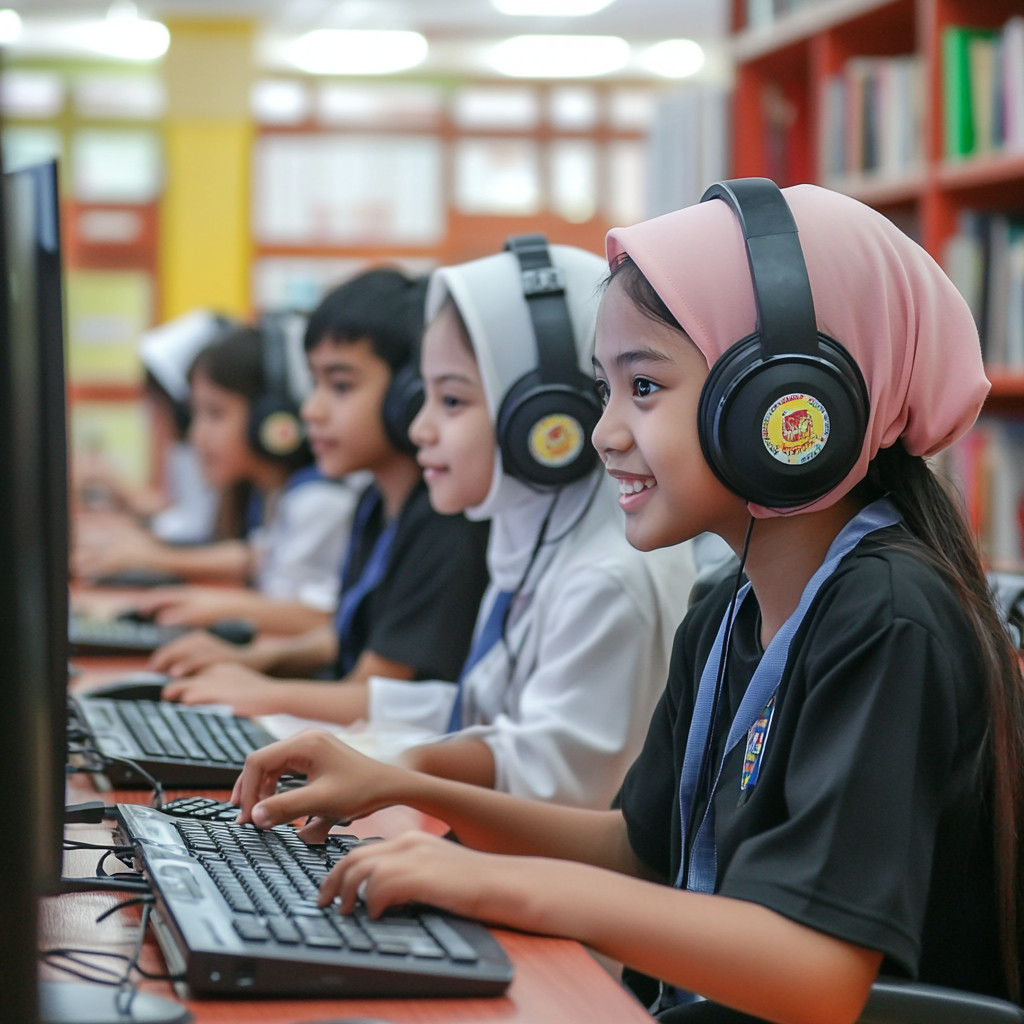}} \hspace{1pt} 
    \subfloat[Detection: 100\%] 
    {\includegraphics[width=0.243\textwidth]{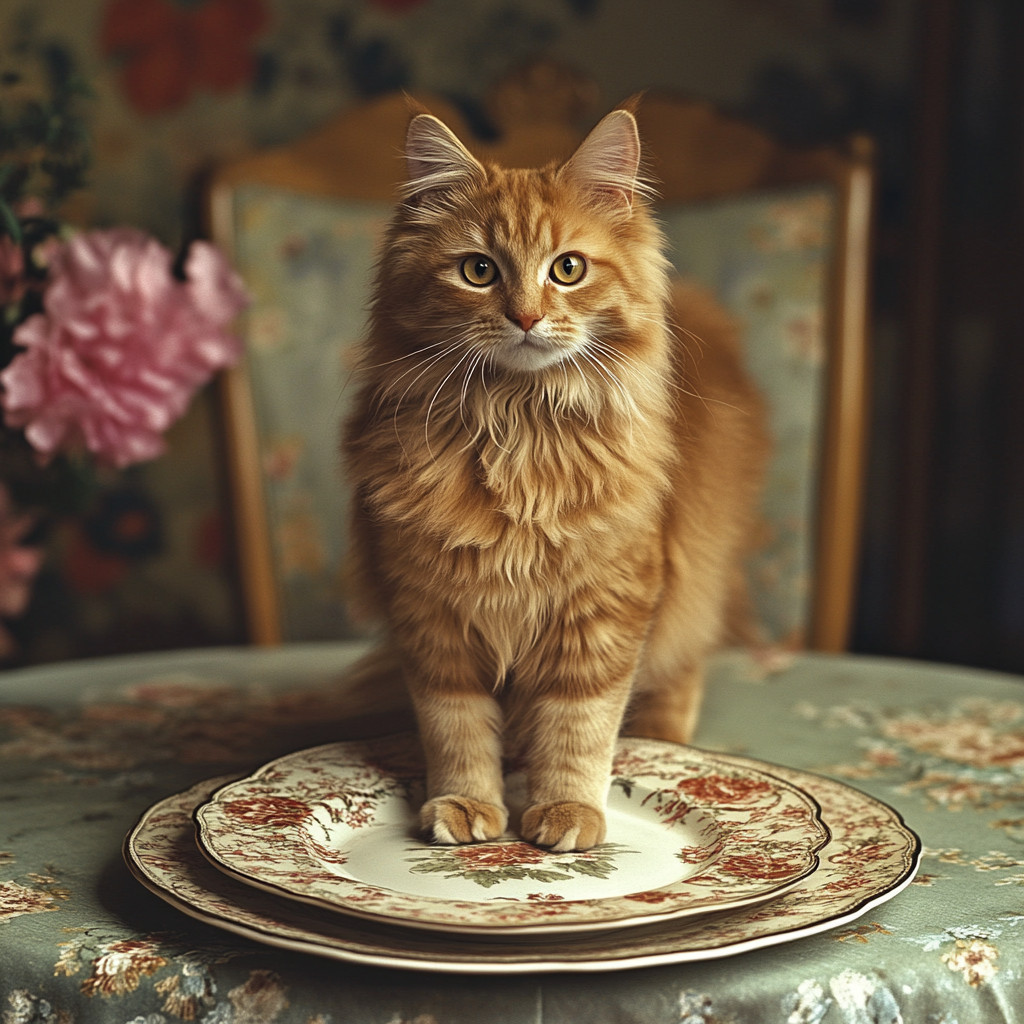}} \hspace{1pt} 
    \subfloat[Detection: 86\%] 
    {\includegraphics[width=0.243\textwidth]{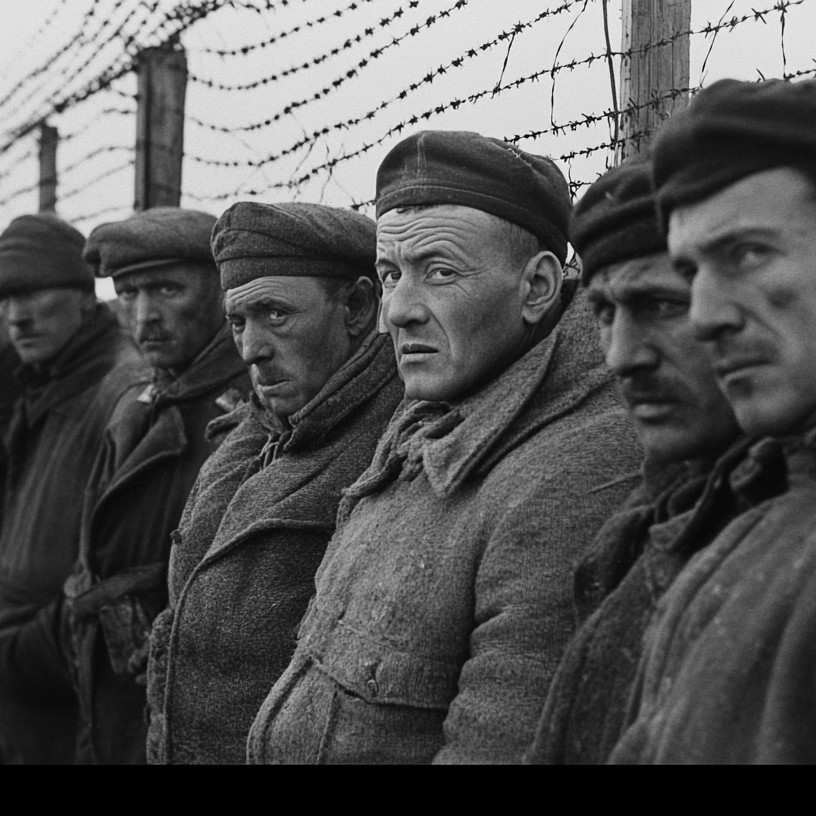}} 
    \caption{Accurately detected MidJourney-v6.1 images. For illustration purposes cropped to a square aspect ratio.}
    \label{fig:qualitative_mj61}
\end{figure*}

\begin{figure*}
    \centering
    \subfloat[Detection: 87\%] {\includegraphics[width=0.243\textwidth]{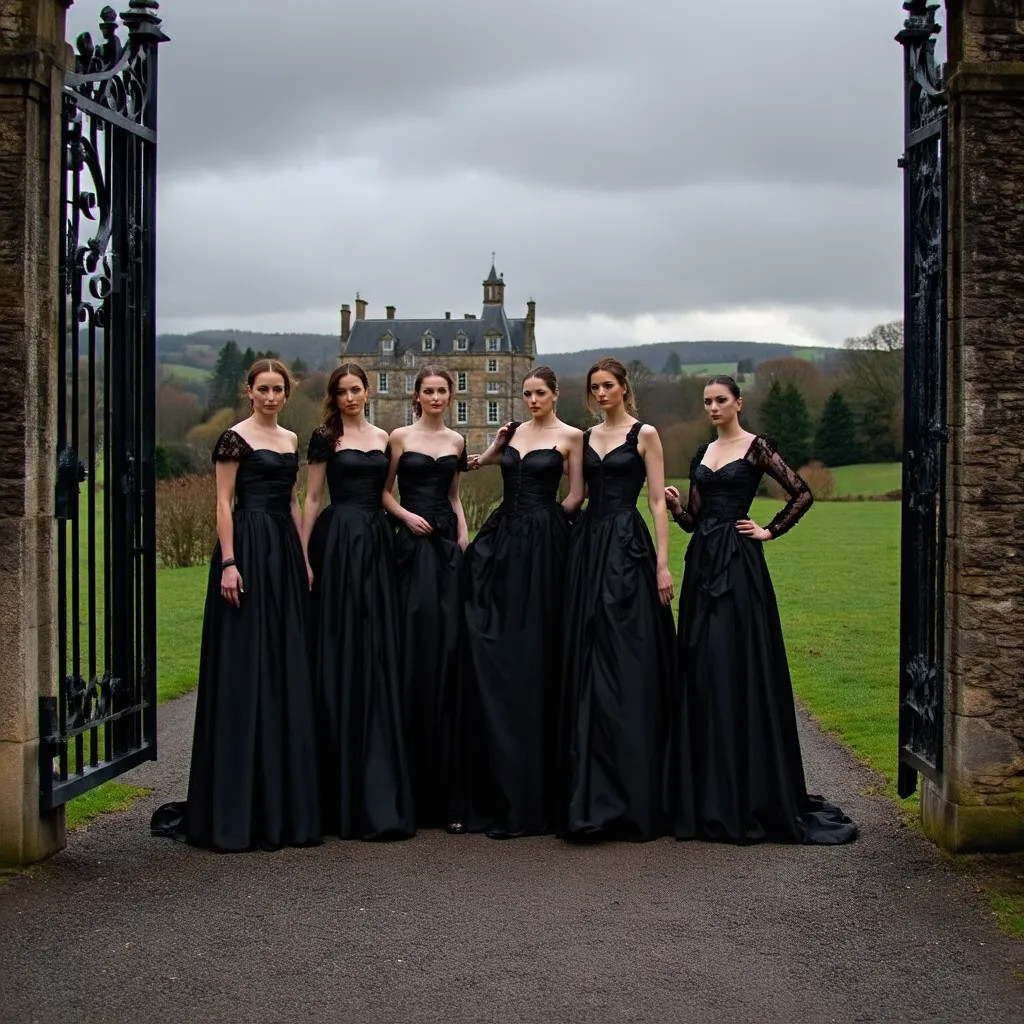}} \hspace{1pt} 
    \subfloat[Detection: 81\%] 
    {\includegraphics[width=0.243\textwidth]{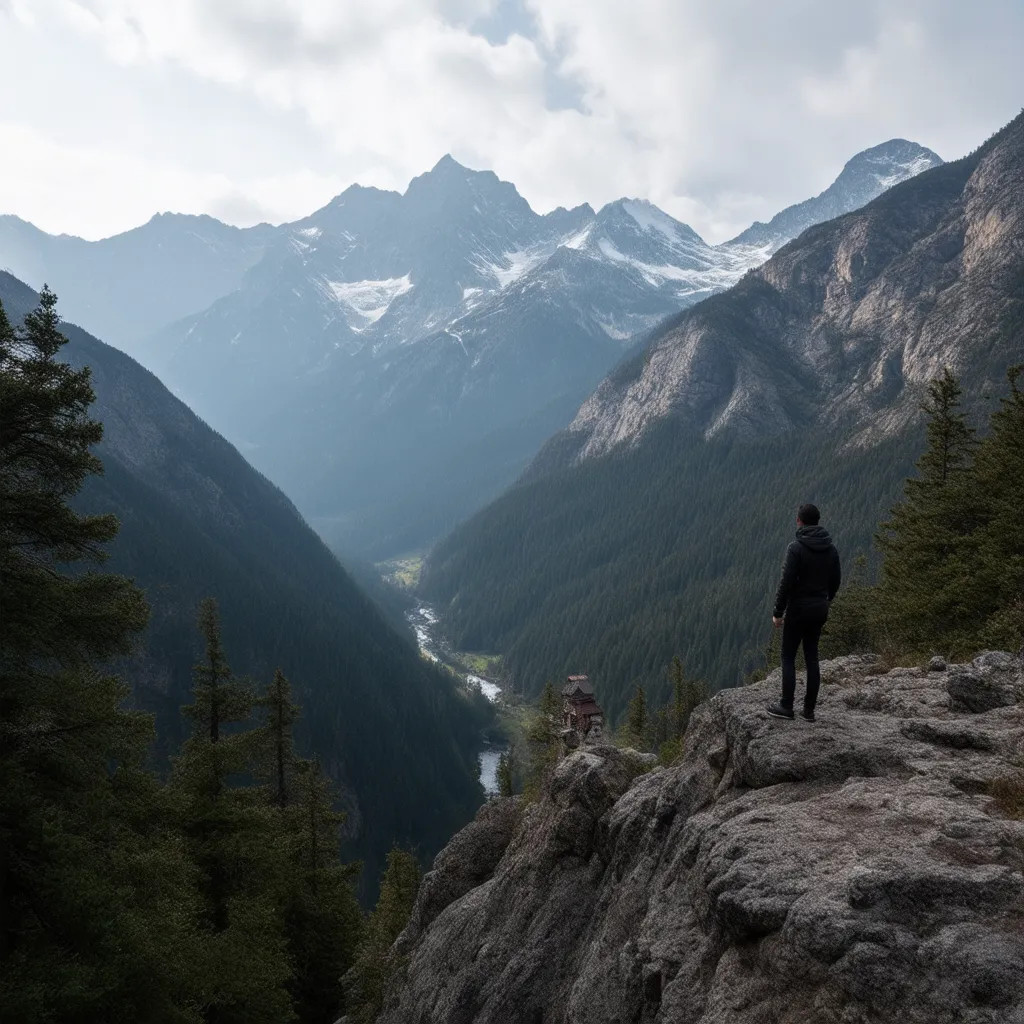}} \hspace{1pt} 
    \subfloat[Detection: 100\%] 
    {\includegraphics[width=0.243\textwidth]{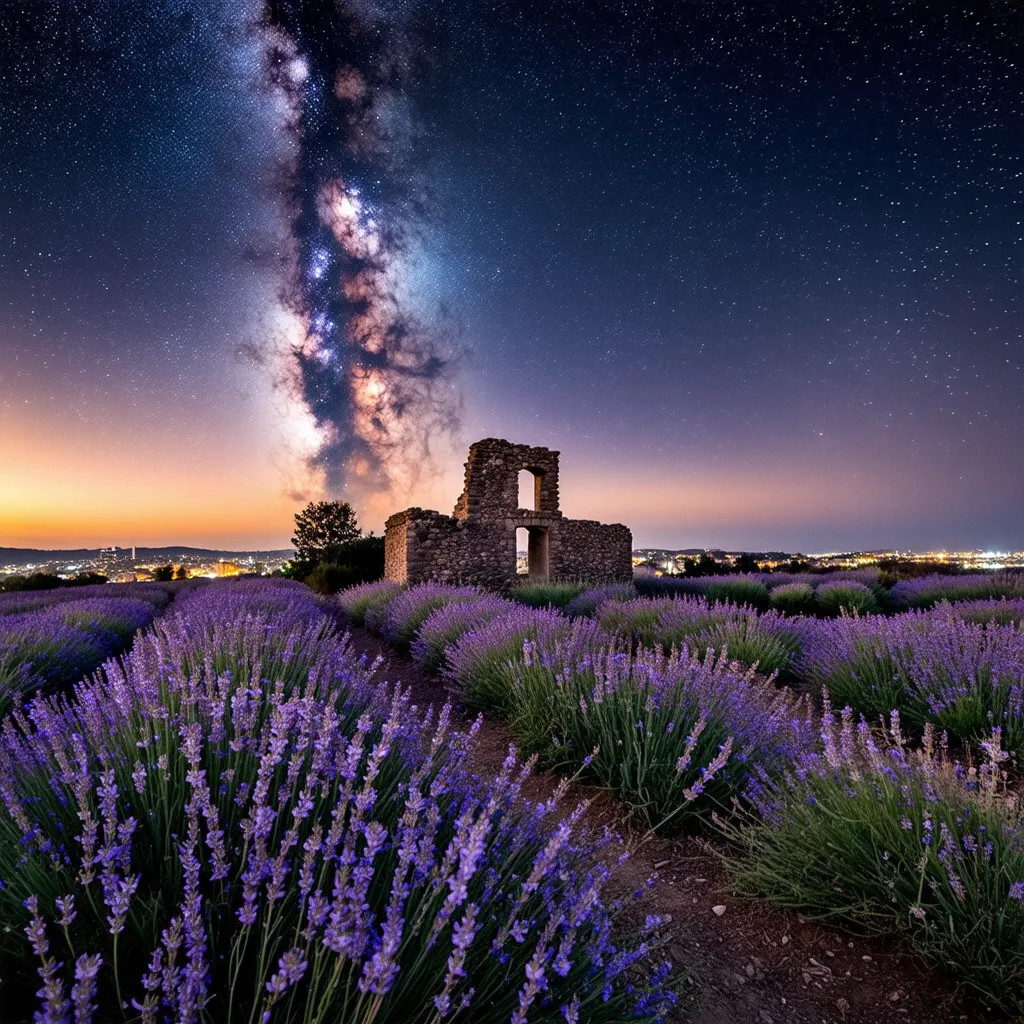}} \hspace{1pt} 
    \subfloat[Detection: 100\%] 
    {\includegraphics[width=0.243\textwidth]{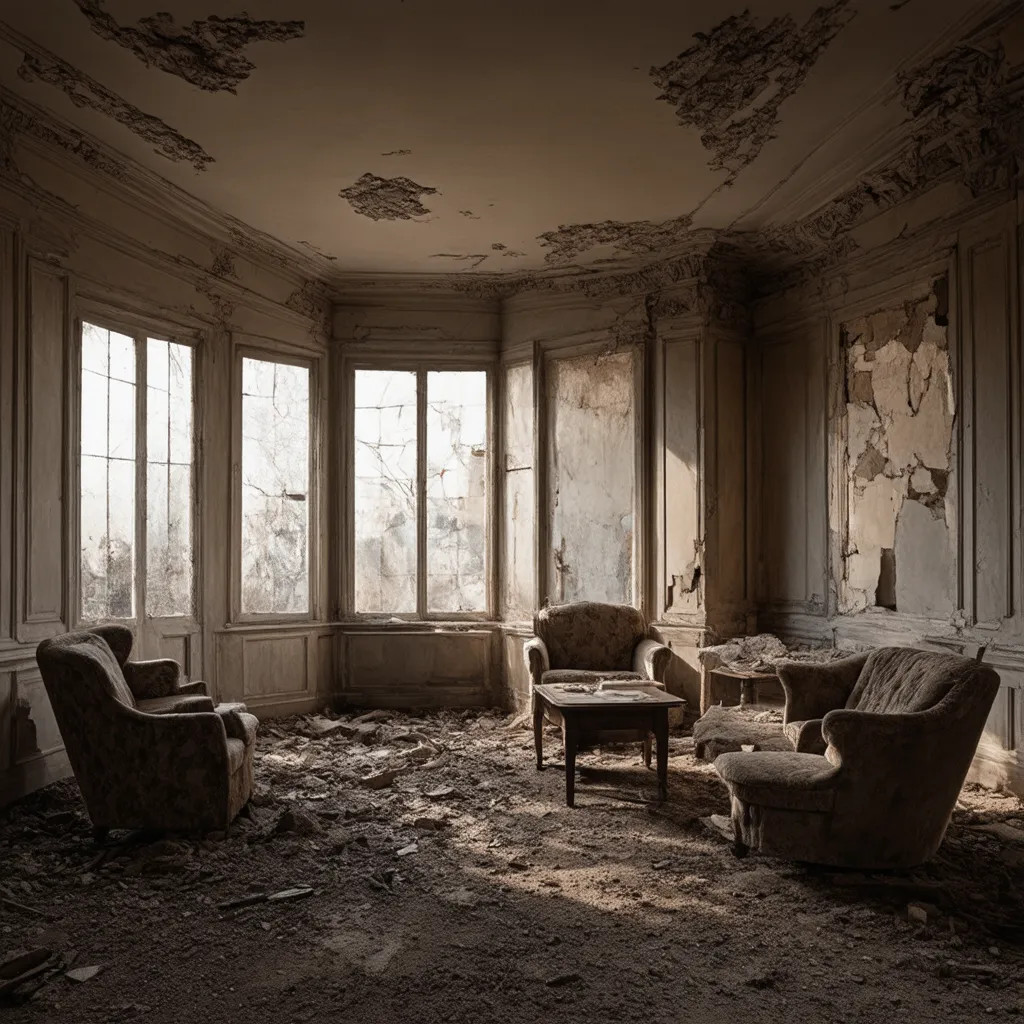}} 
    \caption{Accurately detected Stable Diffusion 3 images. For illustration purposes cropped to a square aspect ratio.}
    \label{fig:qualitative_sd3}
\end{figure*}

\begin{figure*}
    \centering
    \subfloat[Detection: 99\%] {\includegraphics[width=0.243\textwidth]{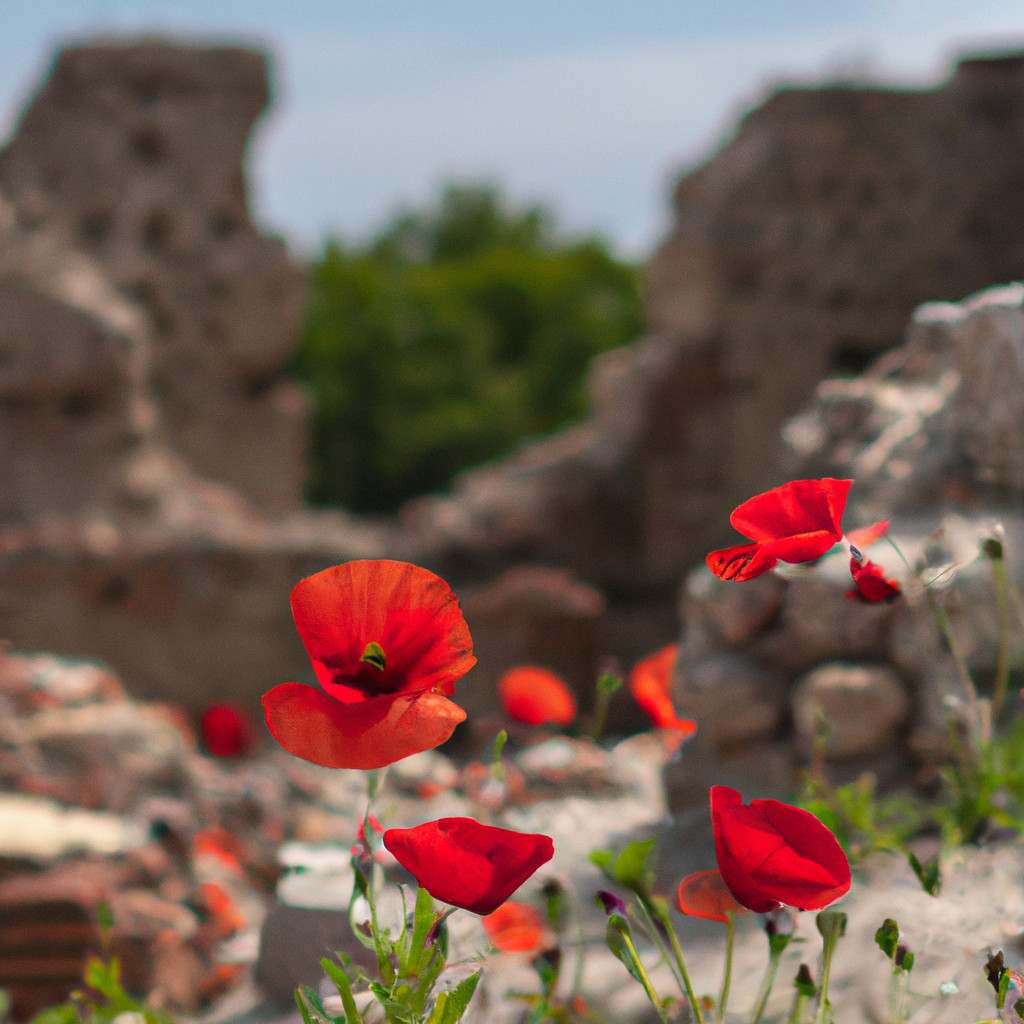}} \hspace{1pt} 
    \subfloat[Detection: 100\%] 
    {\includegraphics[width=0.243\textwidth]{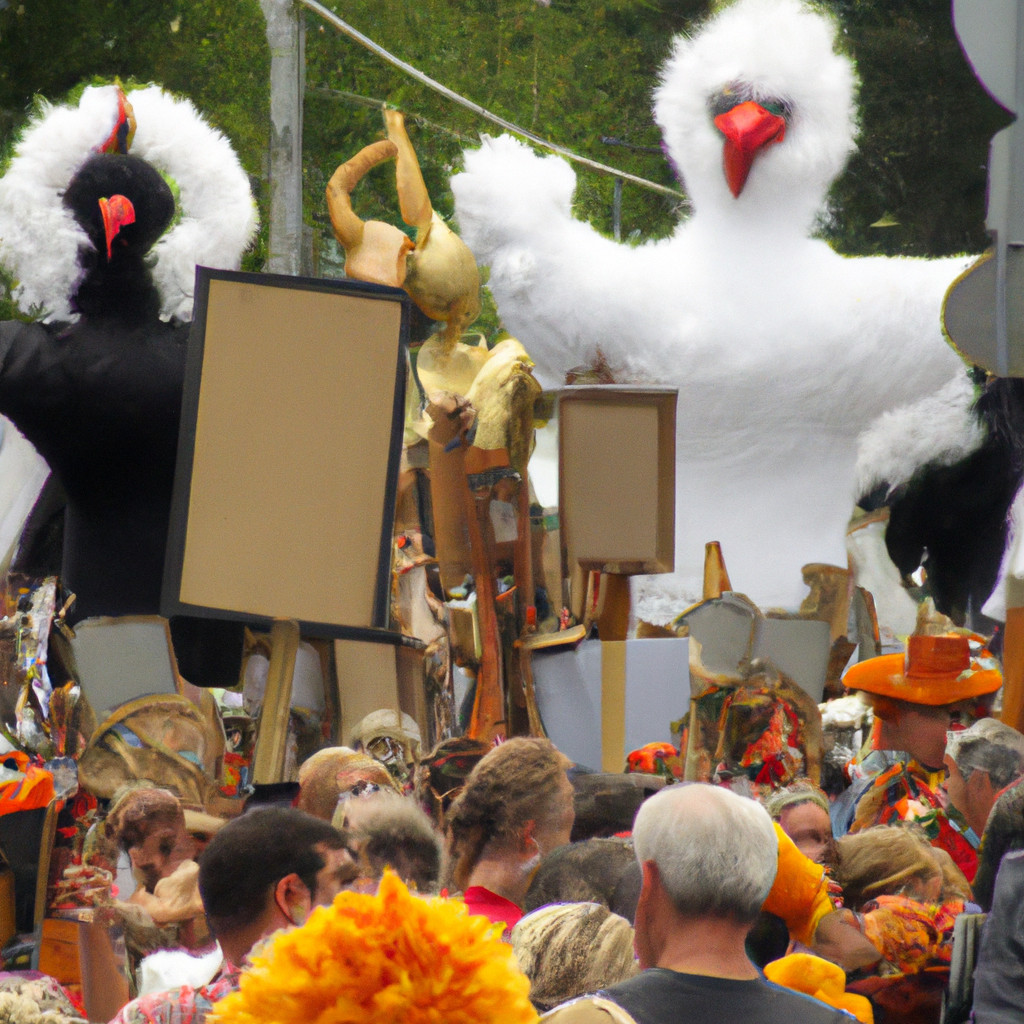}} \hspace{1pt} 
    \subfloat[Detection: 100\%] 
    {\includegraphics[width=0.243\textwidth]{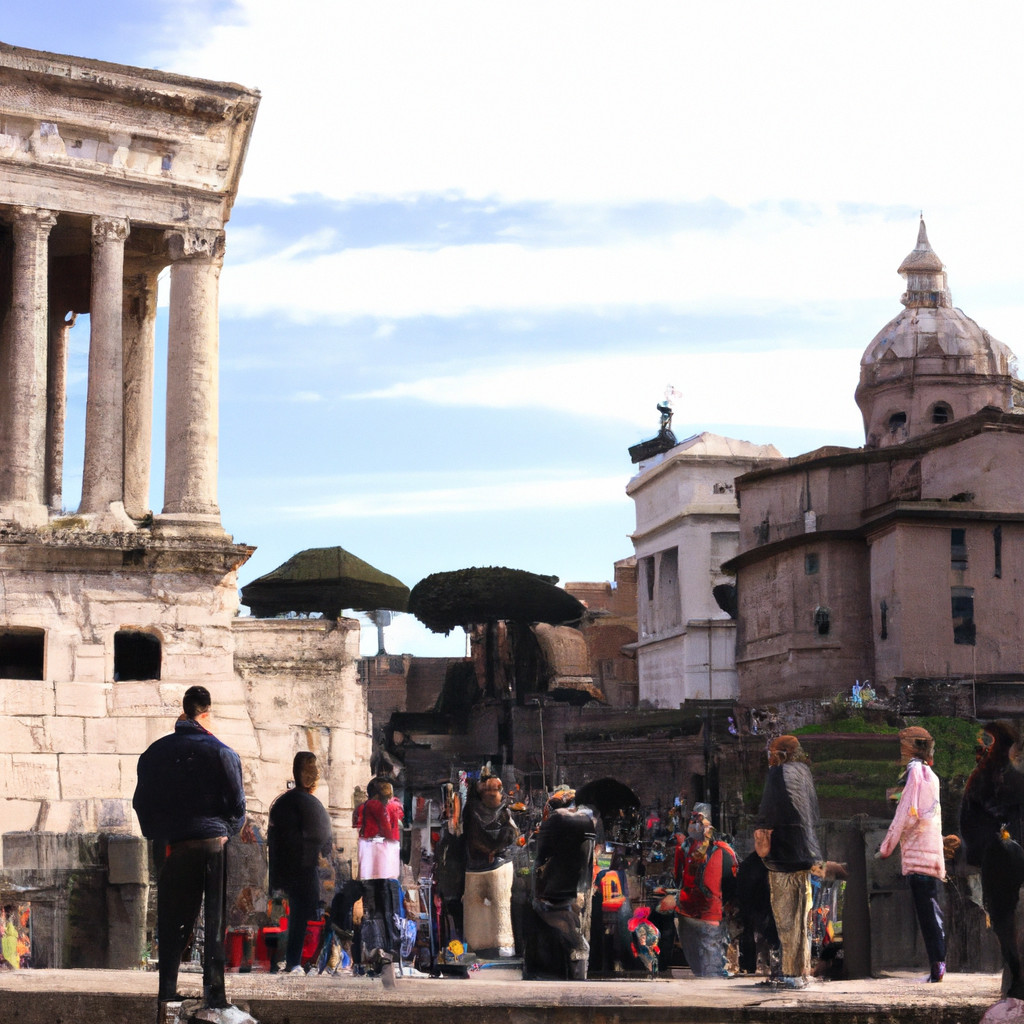}} \hspace{1pt} 
    \subfloat[Detection: 99\%] 
    {\includegraphics[width=0.243\textwidth]{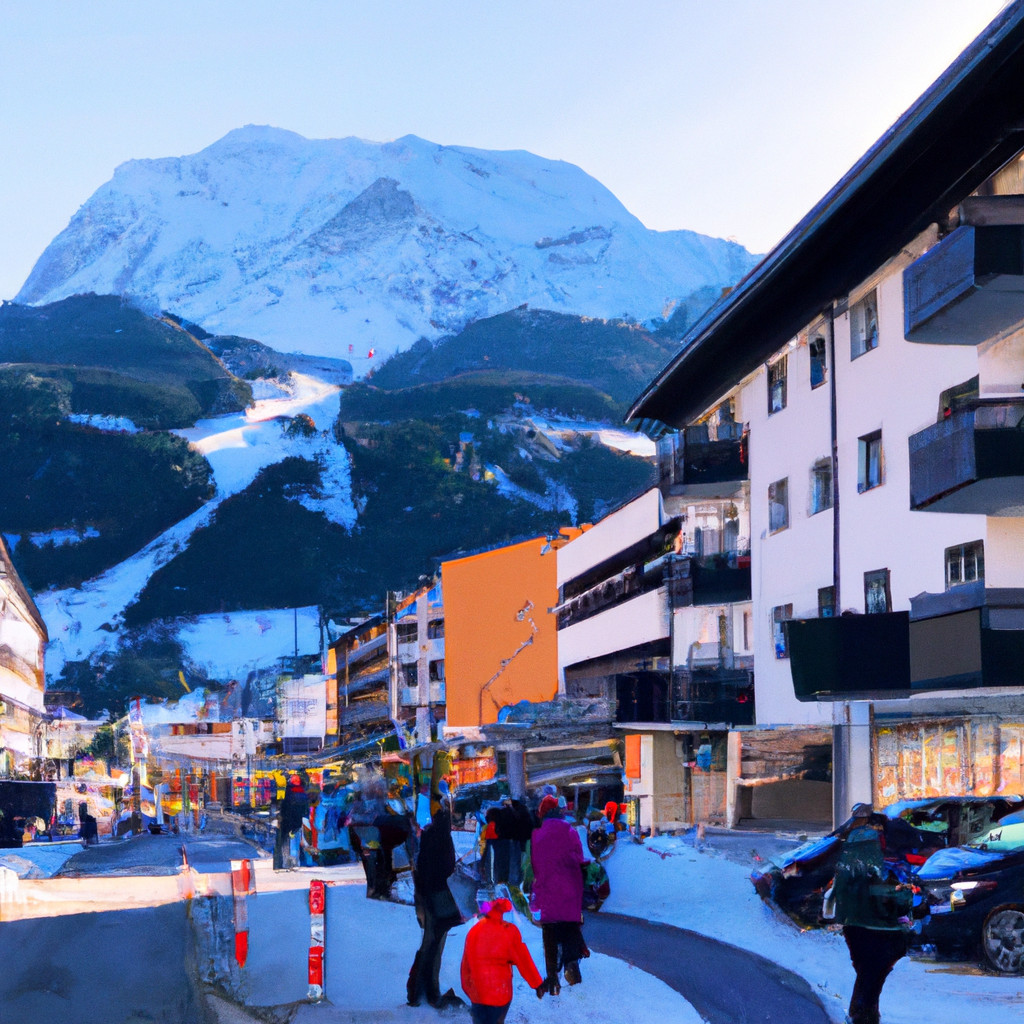}} 
    \caption{Accurately detected DALLE2 images. For illustration purposes cropped to a square aspect ratio.}
    \label{fig:qualitative_dalle2}
\end{figure*}

\begin{figure*}
    \centering
    \subfloat[Detection: 90\%] {\includegraphics[width=0.243\textwidth]{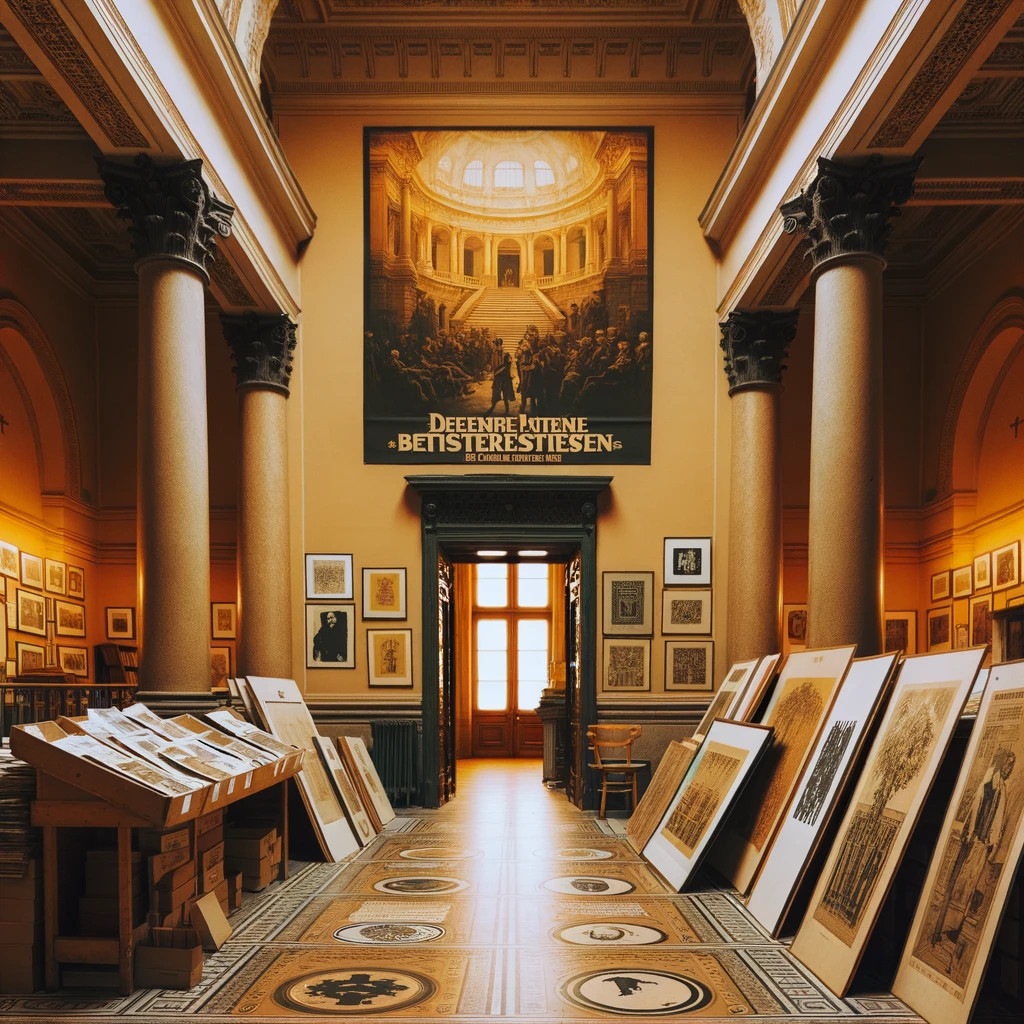}} \hspace{1pt} 
    \subfloat[Detection: 100\%] 
    {\includegraphics[width=0.243\textwidth]{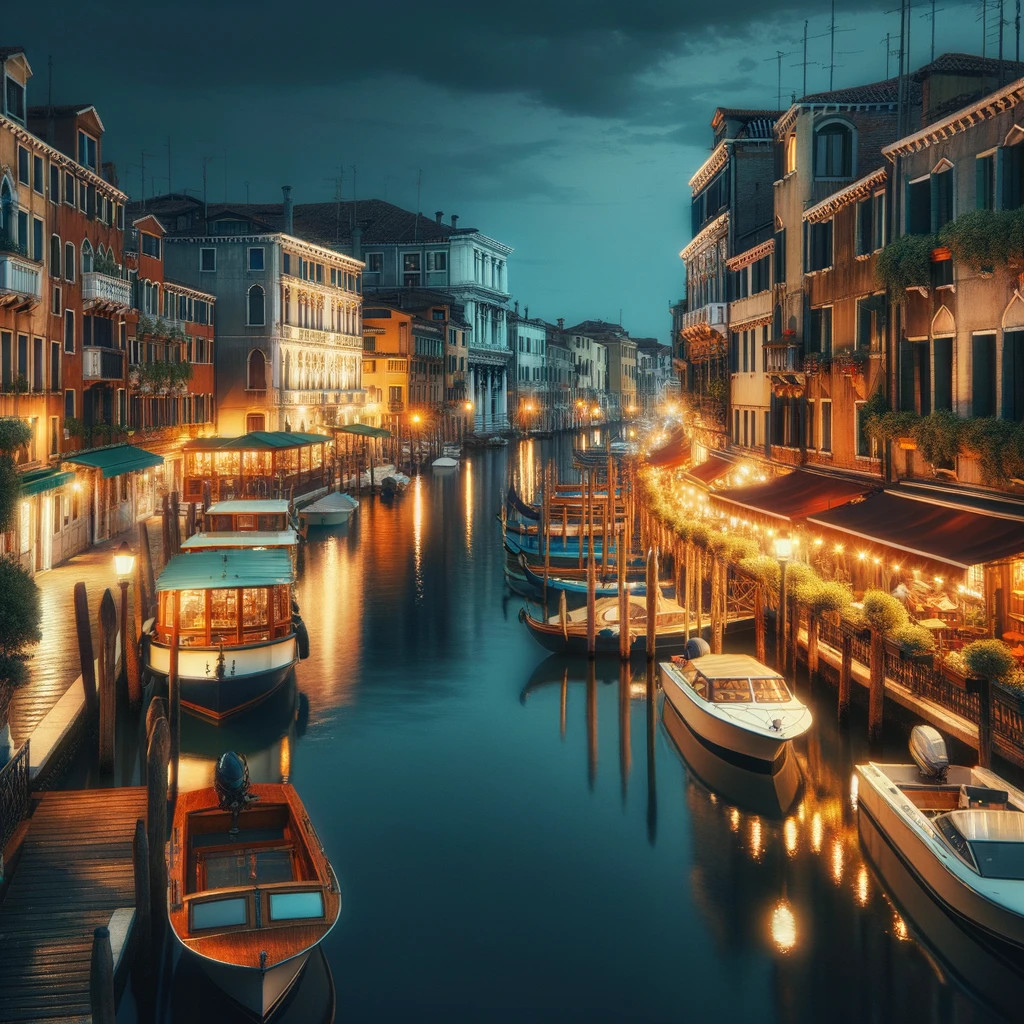}} \hspace{1pt} 
    \subfloat[Detection: 99\%] 
    {\includegraphics[width=0.243\textwidth]{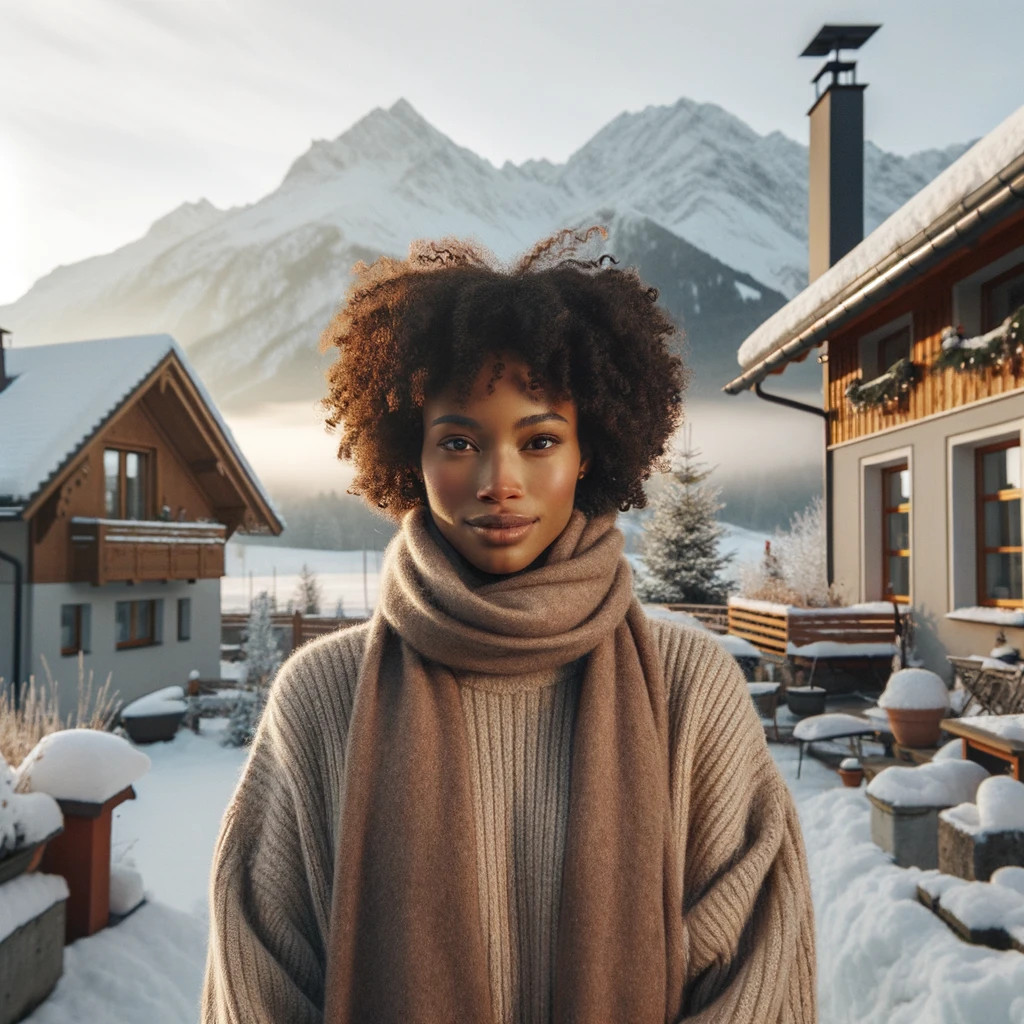}} \hspace{1pt} 
    \subfloat[Detection: 87\%] 
    {\includegraphics[width=0.243\textwidth]{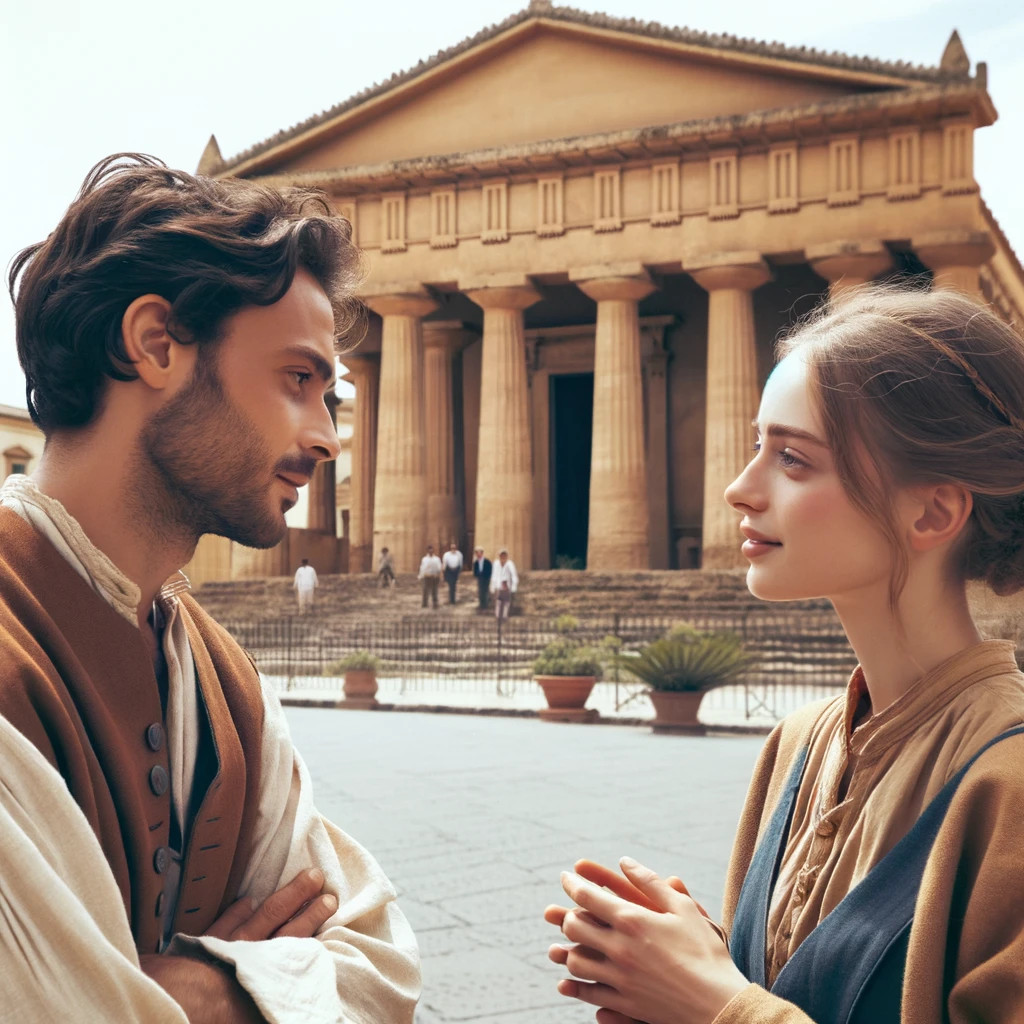}} 
    \caption{Accurately detected DALLE3 images. For illustration purposes cropped to a square aspect ratio.}
    \label{fig:qualitative_dalle3}
\end{figure*}

\begin{figure*}
    \centering
    \subfloat[Detection: 100\%] {\includegraphics[width=0.243\textwidth]{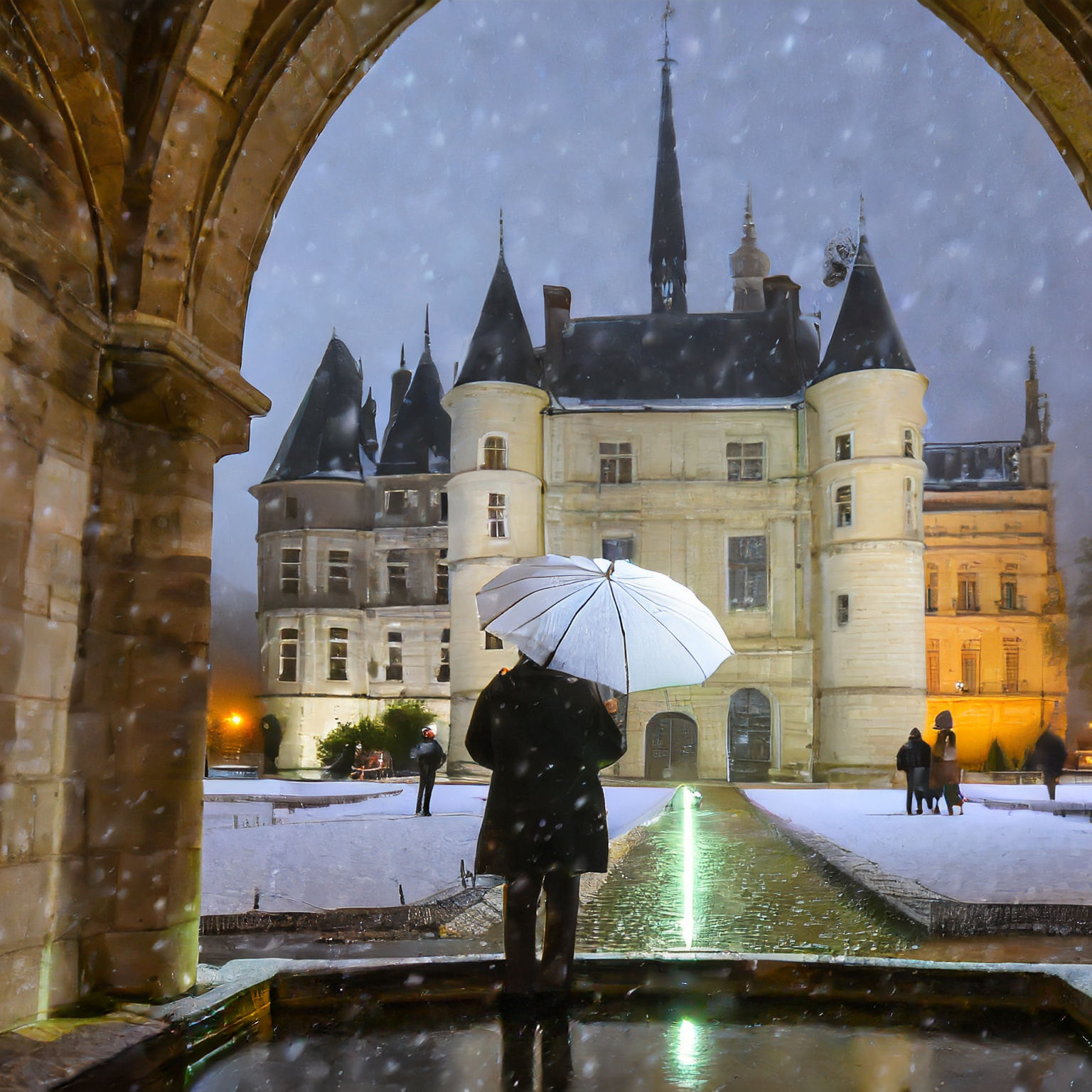}} \hspace{1pt} 
    \subfloat[Detection: 86\%] 
    {\includegraphics[width=0.243\textwidth]{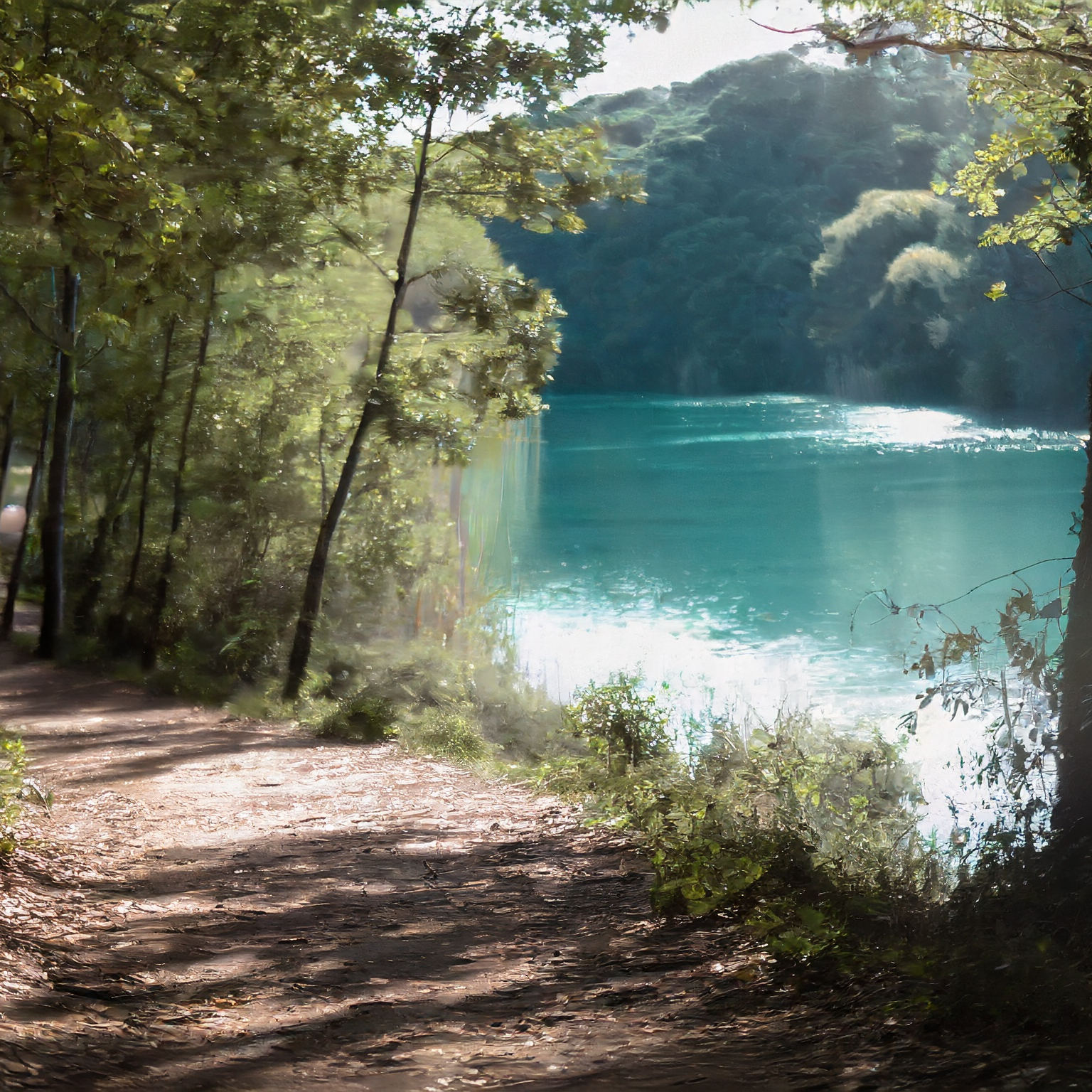}} \hspace{1pt} 
    \subfloat[Detection: 100\%] 
    {\includegraphics[width=0.243\textwidth]{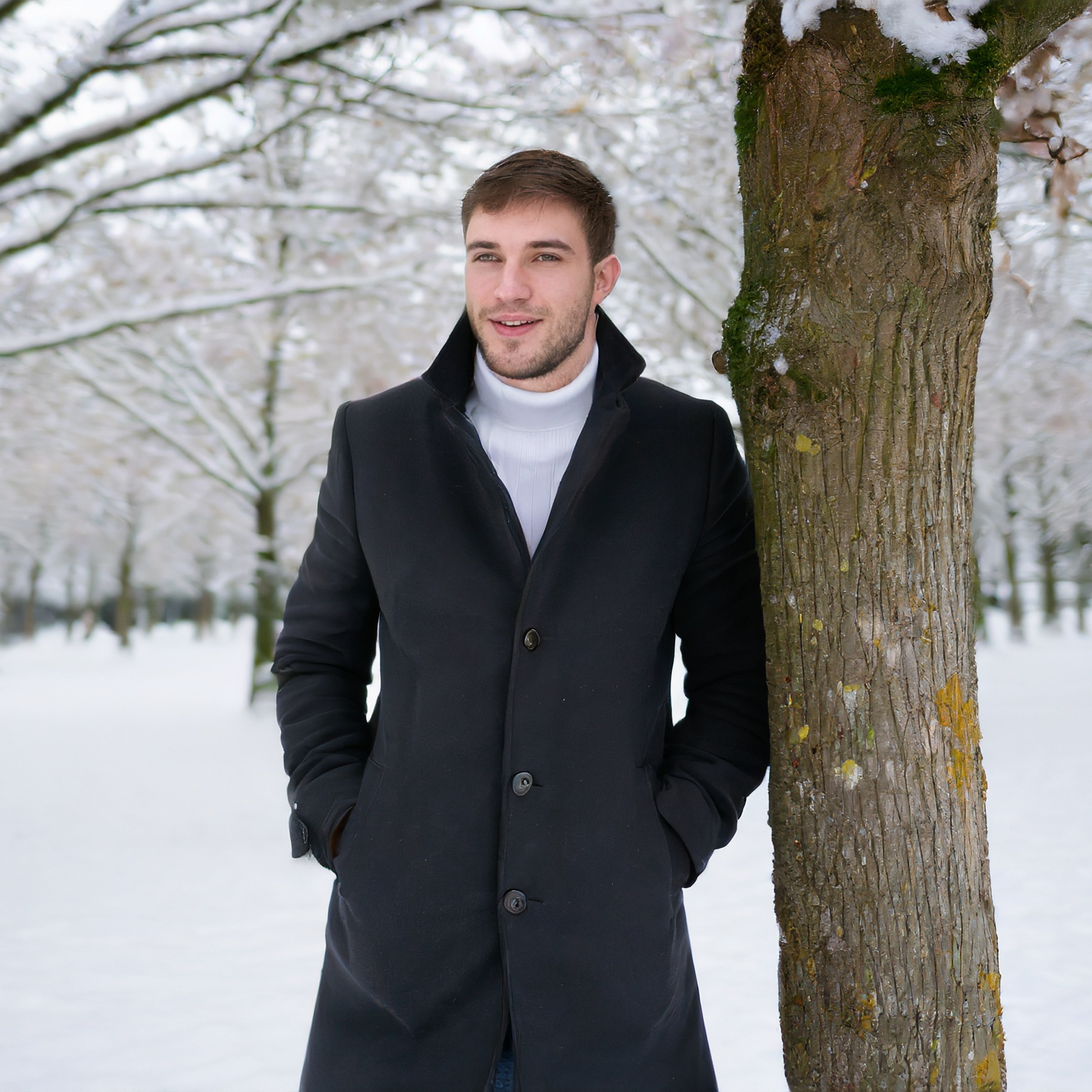}} \hspace{1pt} 
    \subfloat[Detection: 100\%] 
    {\includegraphics[width=0.243\textwidth]{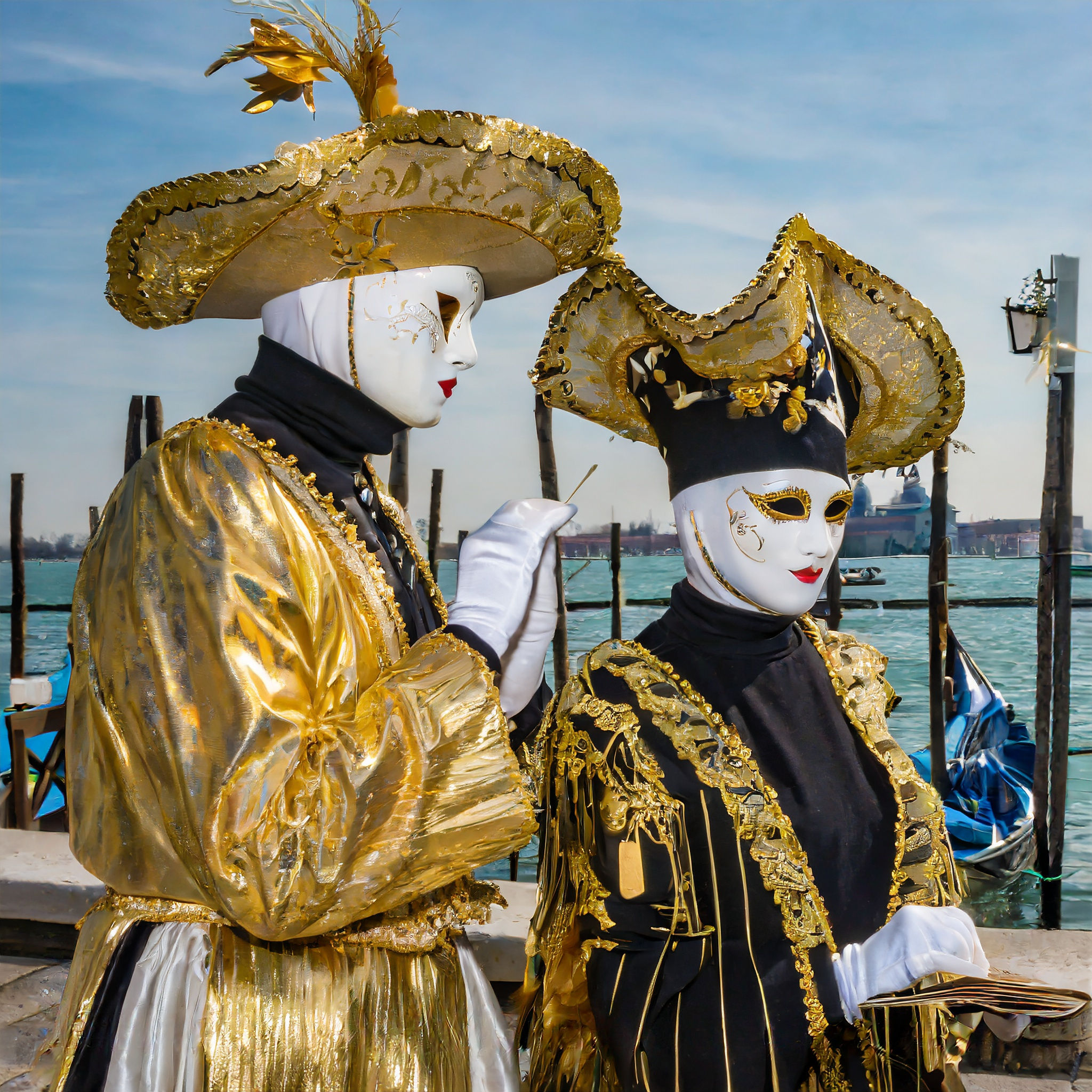}} 
    \caption{Accurately detected Firefly images. For illustration purposes cropped to a square aspect ratio.}
    \label{fig:qualitative_firefly}
\end{figure*}

\begin{figure*}
    \centering
    \subfloat[Detection: 100\%] {\includegraphics[width=0.243\textwidth]{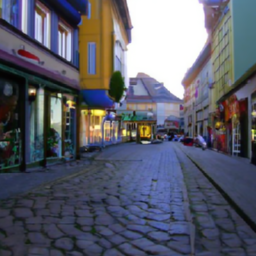}} \hspace{1pt} 
    \subfloat[Detection: 100\%] 
    {\includegraphics[width=0.243\textwidth]{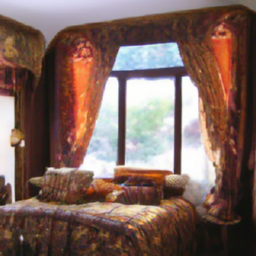}} \hspace{1pt} 
    \subfloat[Detection: 100\%] 
    {\includegraphics[width=0.243\textwidth]{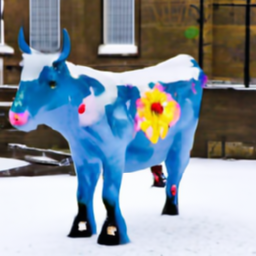}} \hspace{1pt} 
    \subfloat[Detection: 100\%] 
    {\includegraphics[width=0.243\textwidth]{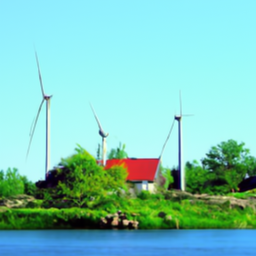}} 
    \caption{Accurately detected Glide images. For illustration purposes cropped to a square aspect ratio.}
    \label{fig:qualitative_glide}
\end{figure*}

\begin{figure*}
    \centering
    \subfloat[Detection: 100\%] {\includegraphics[width=0.243\textwidth]{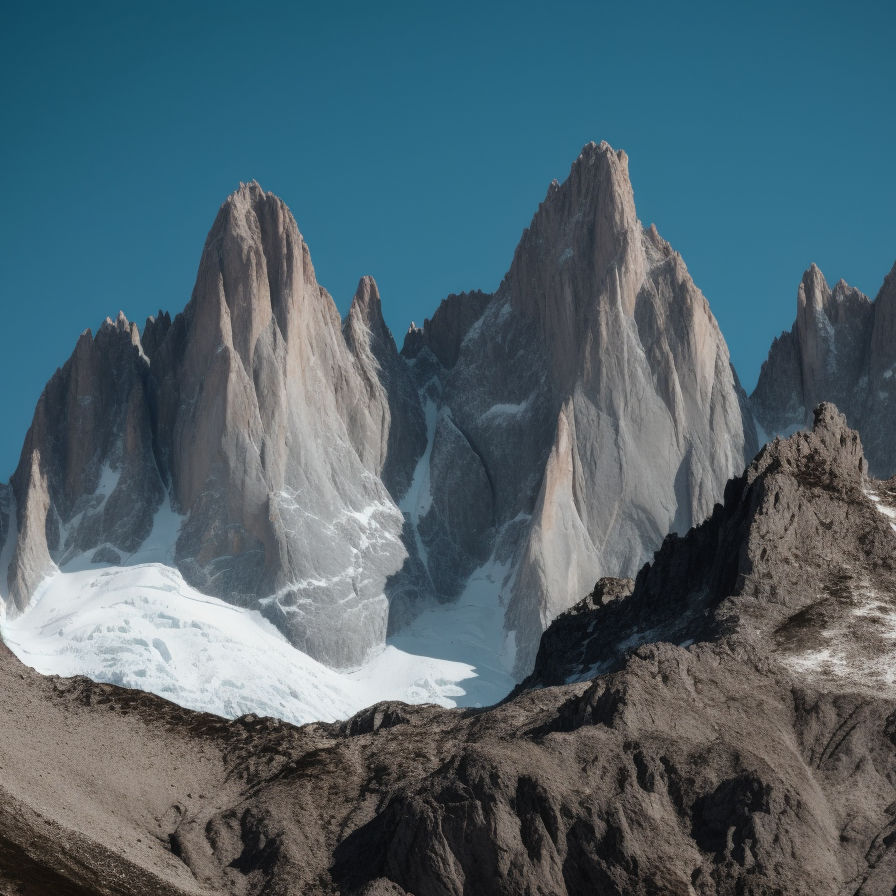}} \hspace{1pt} 
    \subfloat[Detection: 100\%] 
    {\includegraphics[width=0.243\textwidth]{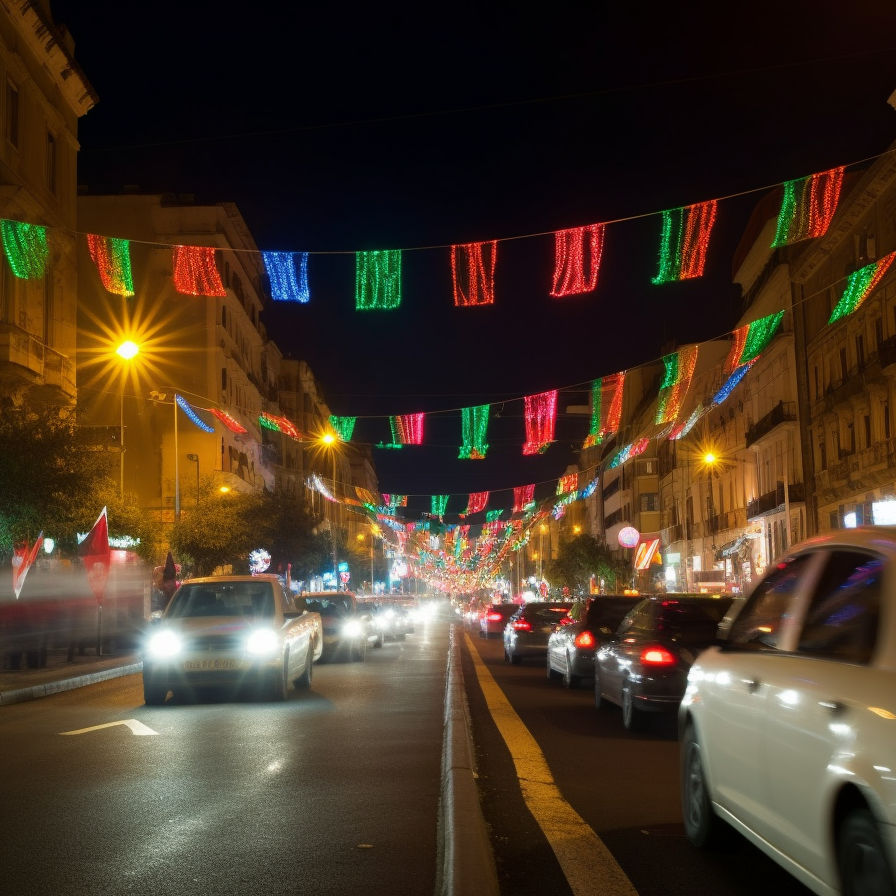}} \hspace{1pt} 
    \subfloat[Detection: 100\%] 
    {\includegraphics[width=0.243\textwidth]{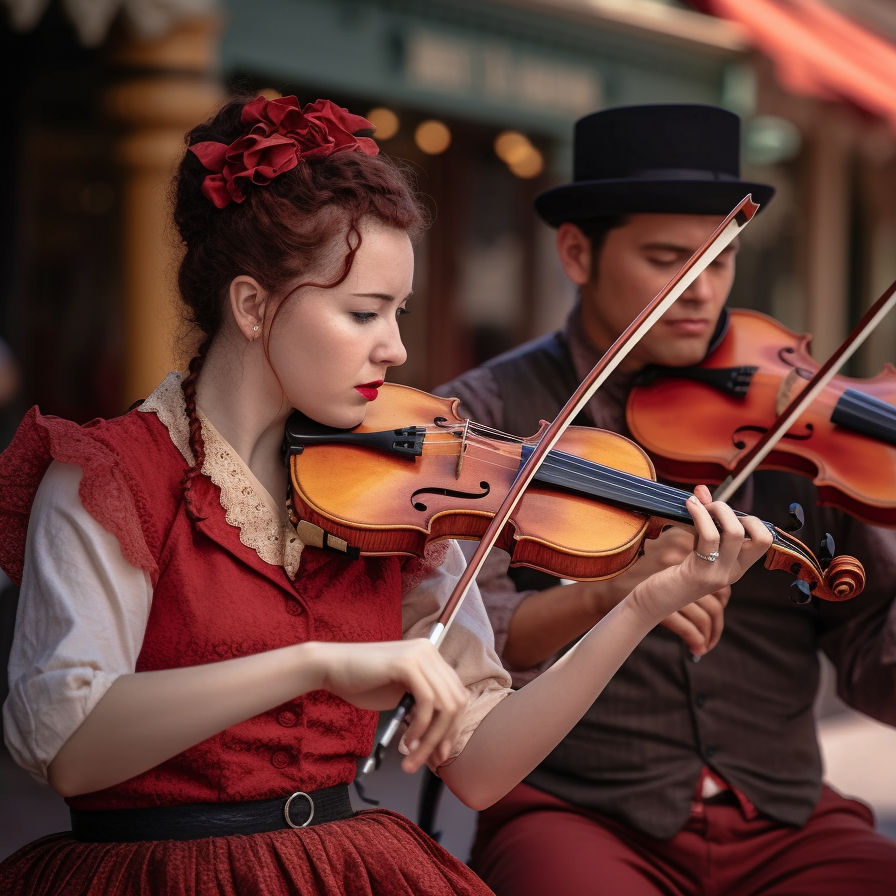}} \hspace{1pt} 
    \subfloat[Detection: 100\%] 
    {\includegraphics[width=0.243\textwidth]{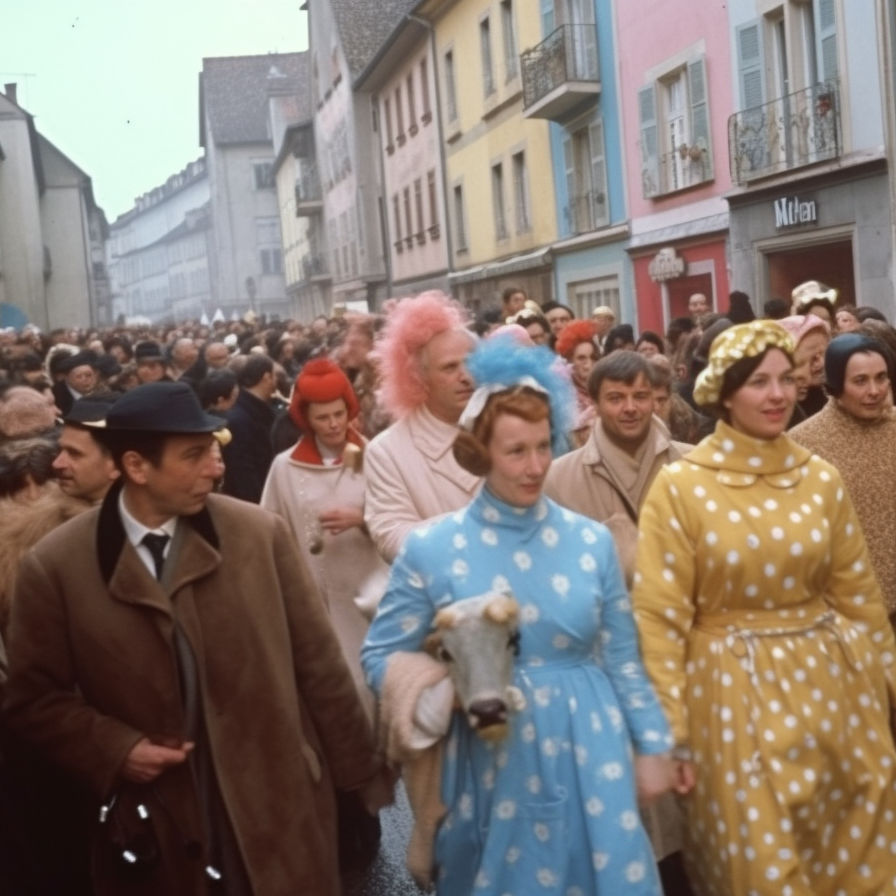}} 
    \caption{Accurately detected MidJourney-v5 images. For illustration purposes cropped to a square aspect ratio.}
    \label{fig:qualitative_mj5}
\end{figure*}

\begin{figure*}
    \centering
    \subfloat[Detection: 100\%] {\includegraphics[width=0.243\textwidth]{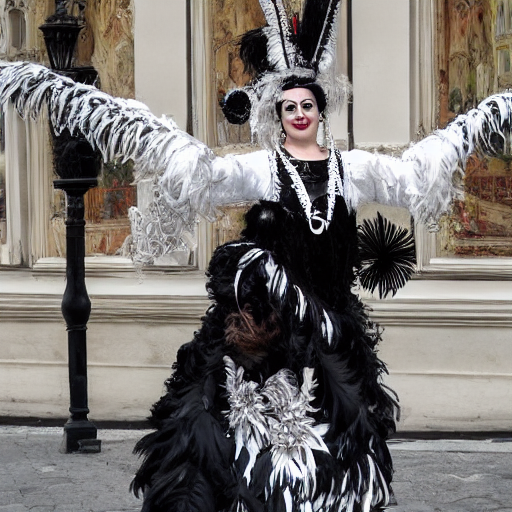}} \hspace{1pt} 
    \subfloat[Detection: 100\%] 
    {\includegraphics[width=0.243\textwidth]{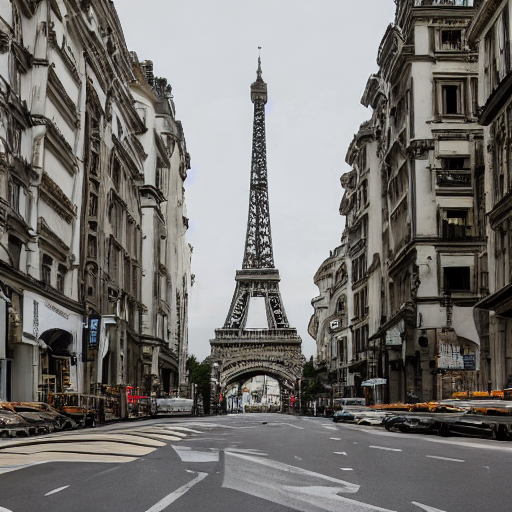}} \hspace{1pt} 
    \subfloat[Detection: 100\%] 
    {\includegraphics[width=0.243\textwidth]{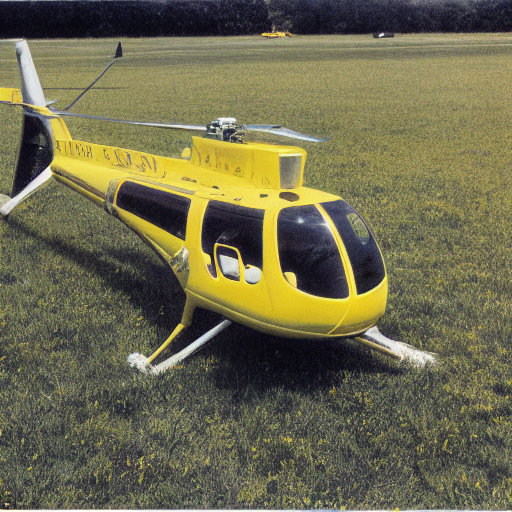}} \hspace{1pt} 
    \subfloat[Detection: 100\%] 
    {\includegraphics[width=0.243\textwidth]{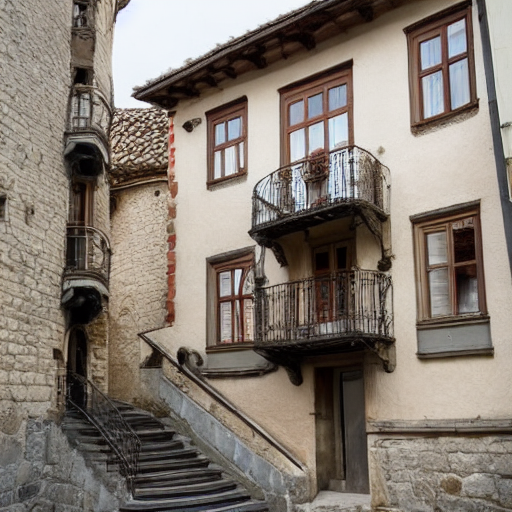}} 
    \caption{Accurately detected Stable Diffusion 1.3 images. For illustration purposes cropped to a square aspect ratio.}
    \label{fig:qualitative_sd13}
\end{figure*}

\begin{figure*}
    \centering
    \subfloat[Detection: 100\%] {\includegraphics[width=0.243\textwidth]{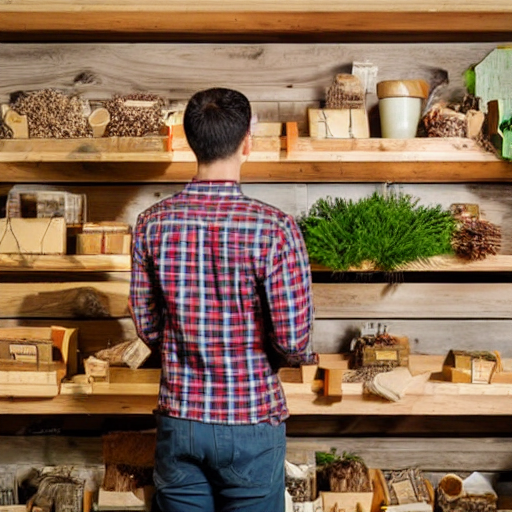}} \hspace{1pt} 
    \subfloat[Detection: 100\%] 
    {\includegraphics[width=0.243\textwidth]{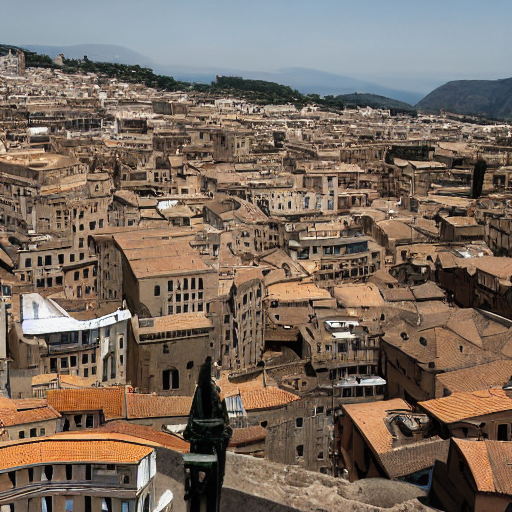}} \hspace{1pt} 
    \subfloat[Detection: 100\%] 
    {\includegraphics[width=0.243\textwidth]{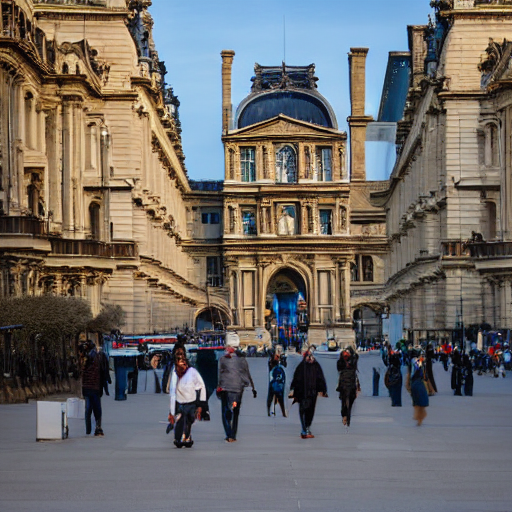}} \hspace{1pt} 
    \subfloat[Detection: 100\%] 
    {\includegraphics[width=0.243\textwidth]{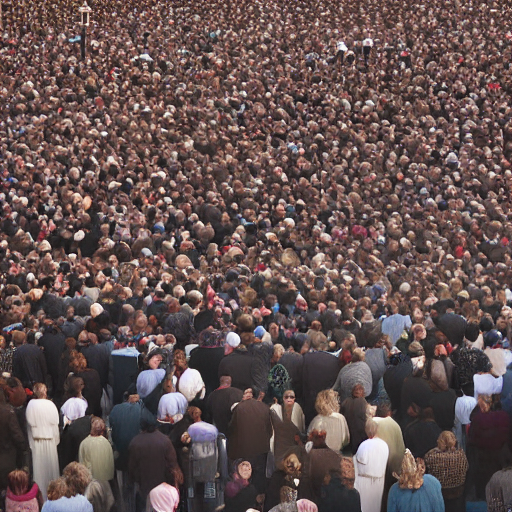}} 
    \caption{Accurately detected Stable Diffusion 1.4 images. For illustration purposes cropped to a square aspect ratio.}
    \label{fig:qualitative_sd14}
\end{figure*}

\begin{figure*}
    \centering
    \subfloat[Detection: 100\%] {\includegraphics[width=0.243\textwidth]{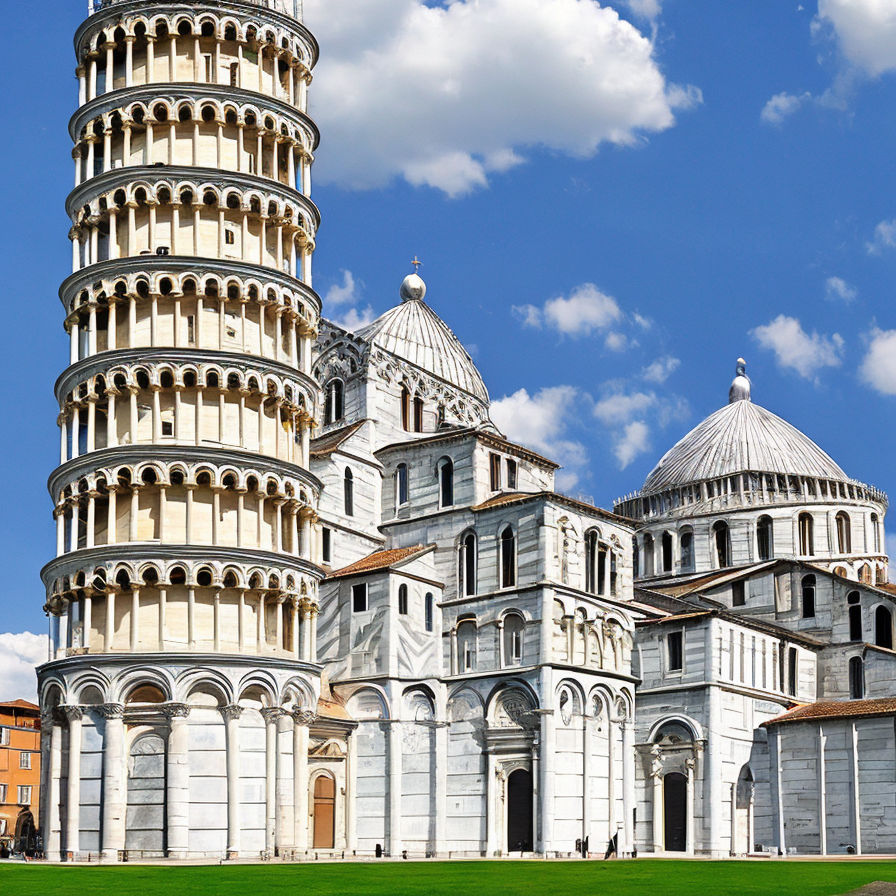}} \hspace{1pt} 
    \subfloat[Detection: 98\%] 
    {\includegraphics[width=0.243\textwidth]{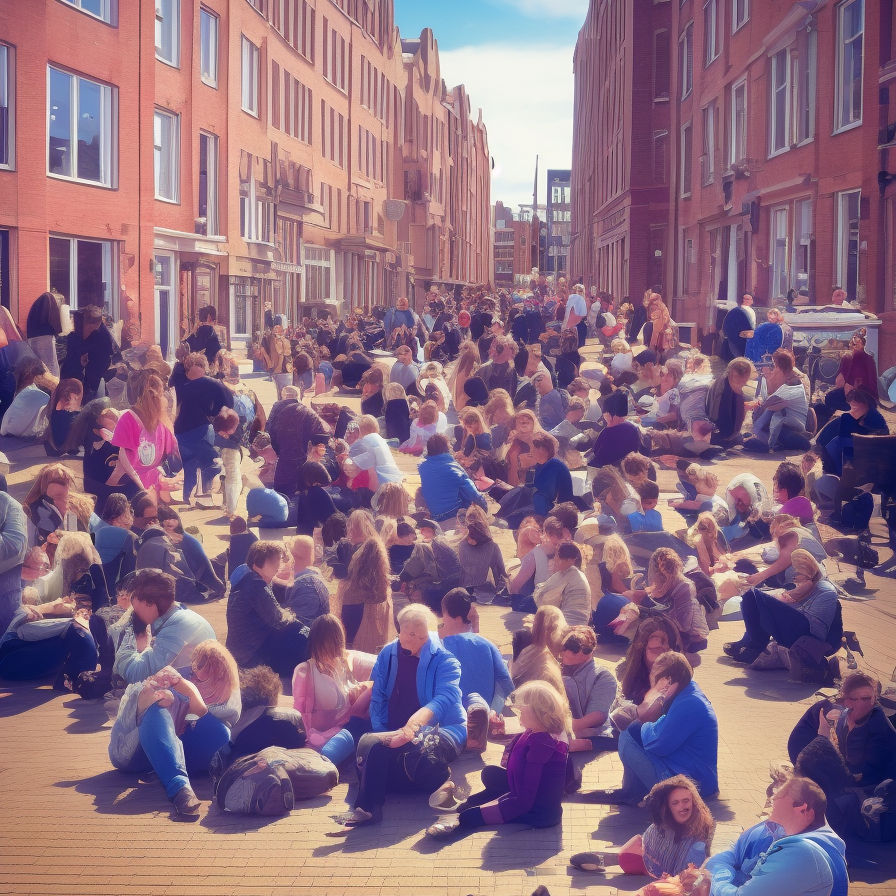}} \hspace{1pt} 
    \subfloat[Detection: 100\%] 
    {\includegraphics[width=0.243\textwidth]{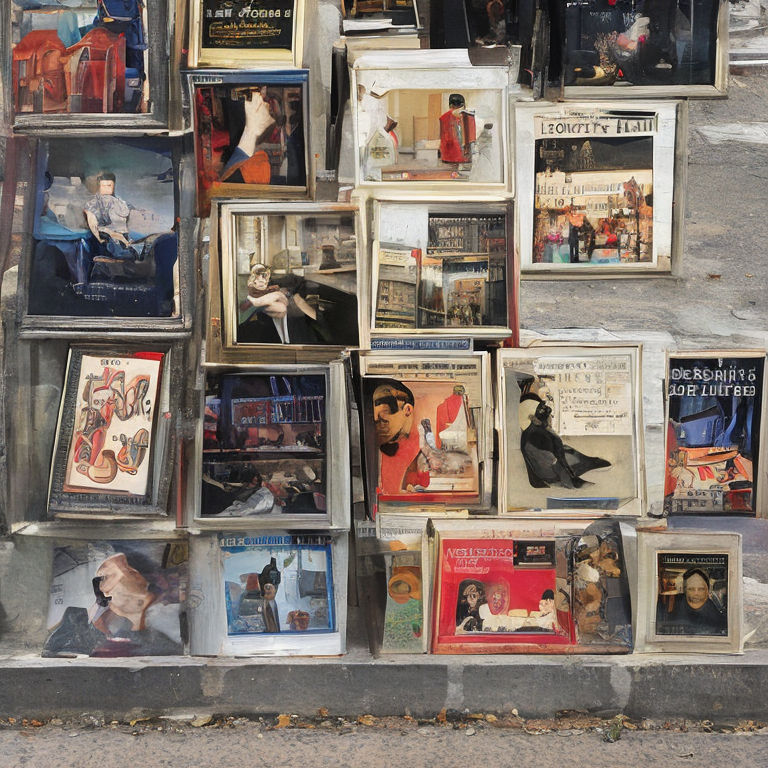}} \hspace{1pt} 
    \subfloat[Detection: 100\%] 
    {\includegraphics[width=0.243\textwidth]{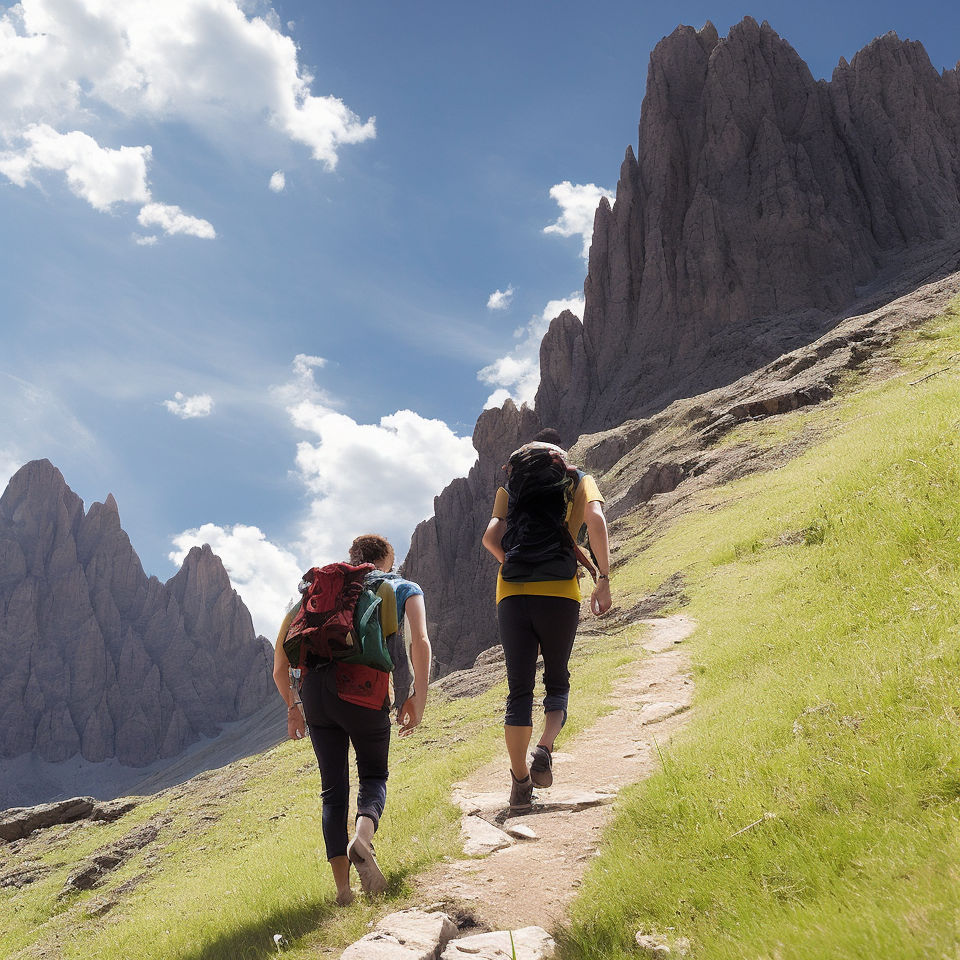}} 
    \caption{Accurately detected Stable Diffusion 2 images. For illustration purposes cropped to a square aspect ratio.}
    \label{fig:qualitative_sd2}
\end{figure*}

\begin{figure*}
    \centering 
    \subfloat[Detection: 100\%] 
    {\includegraphics[width=0.243\textwidth]{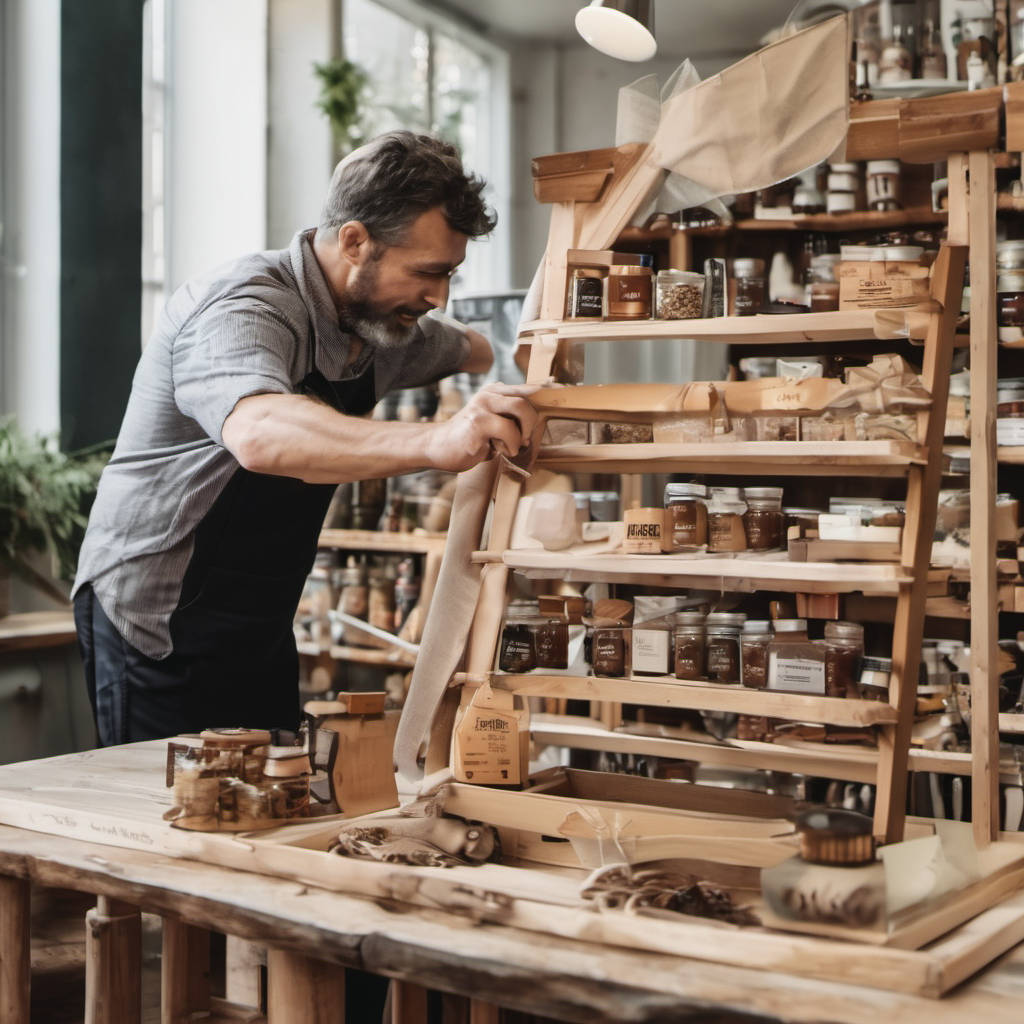}} \hspace{1pt} 
    \subfloat[Detection: 100\%] 
    {\includegraphics[width=0.243\textwidth]{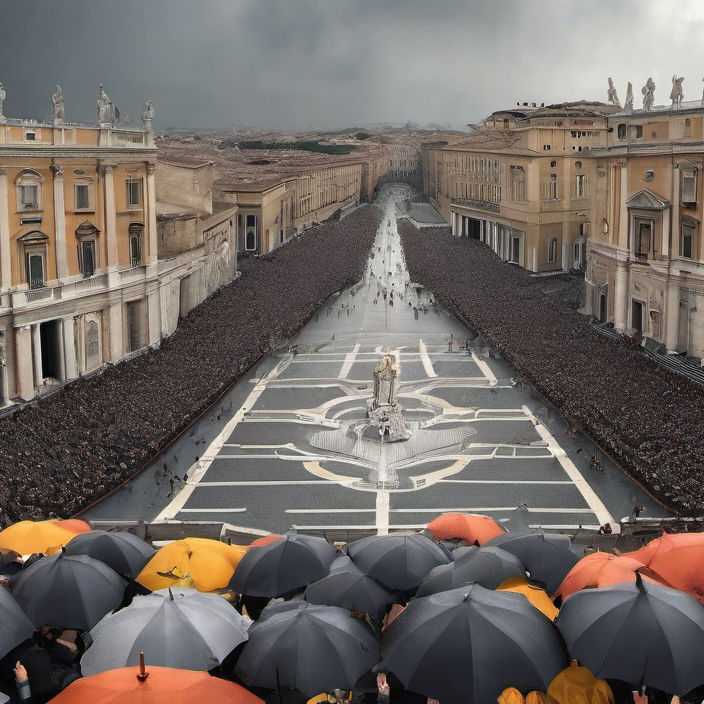}} \hspace{1pt} 
    \subfloat[Detection: 82\%] {\includegraphics[width=0.243\textwidth]{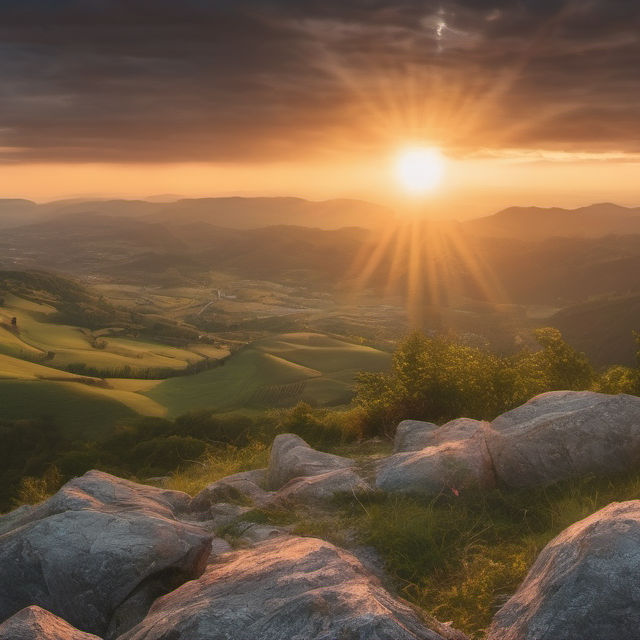}} \hspace{1pt}
    \subfloat[Detection: 100\%] 
    {\includegraphics[width=0.243\textwidth]{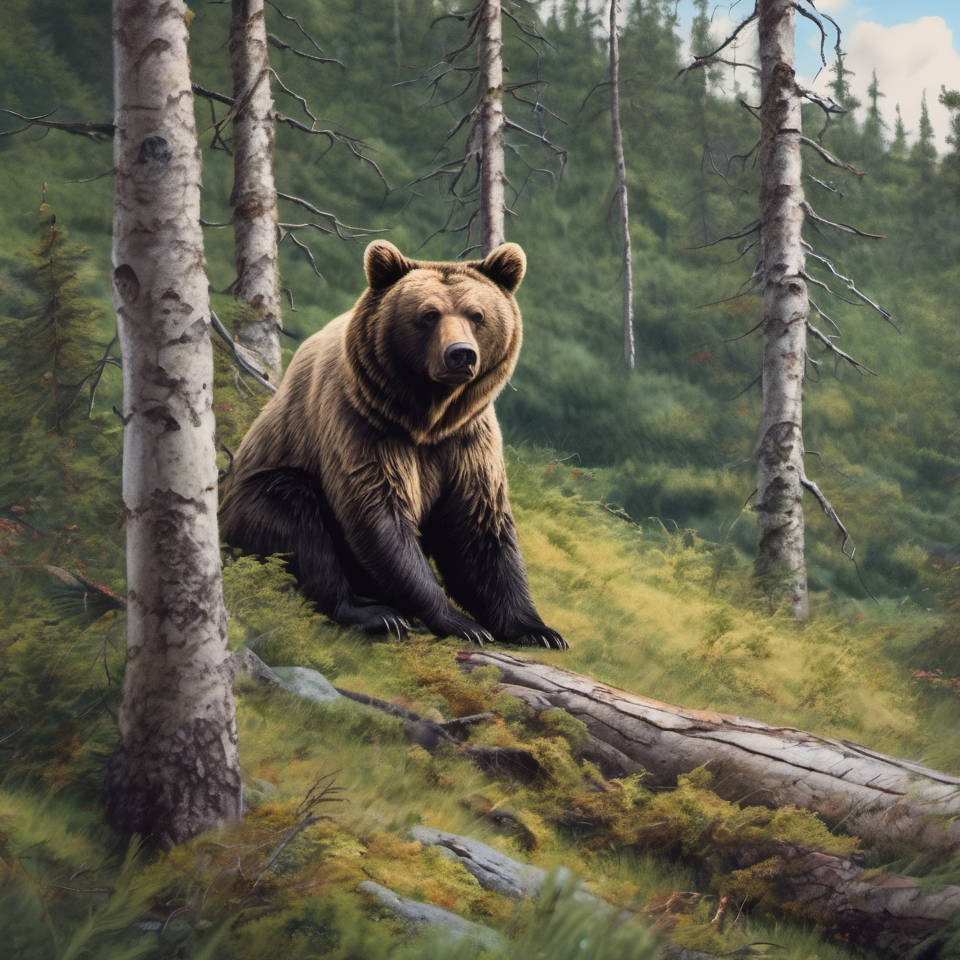}} 
    \caption{Accurately detected Stable Diffusion XL images. For illustration purposes cropped to a square aspect ratio.}
    \label{fig:qualitative_sdxl}
\end{figure*}

\begin{figure*}
    \centering 
    \subfloat[Detection: 0\%] 
    {\includegraphics[width=0.243\textwidth]{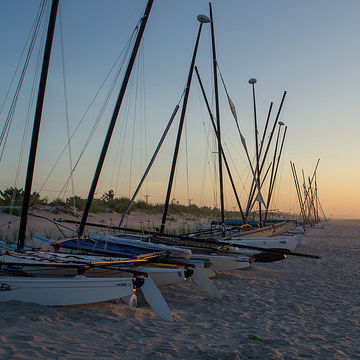}} \hspace{1pt} 
    \subfloat[Detection: 0\%] 
    {\includegraphics[width=0.243\textwidth]{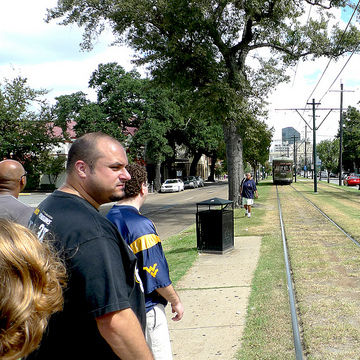}} \hspace{1pt} 
    \subfloat[Detection: 0\%] {\includegraphics[width=0.243\textwidth]{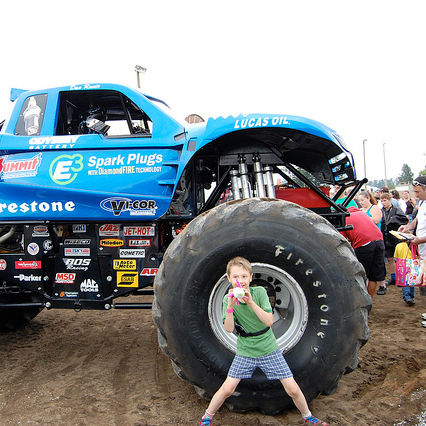}} \hspace{1pt}
    \subfloat[Detection: 0\%] 
    {\includegraphics[width=0.243\textwidth]{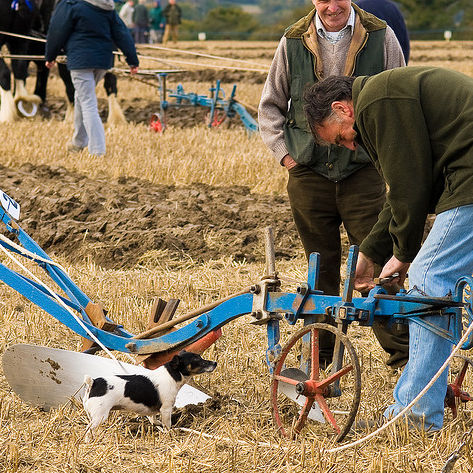}} 
    \caption{Accurately classified real images from COCO. For illustration purposes cropped to a square aspect ratio.}
    \label{fig:qualitative_coco}
\end{figure*}

\begin{figure*}
    \centering 
    \subfloat[Detection: 0\%] 
    {\includegraphics[width=0.243\textwidth]{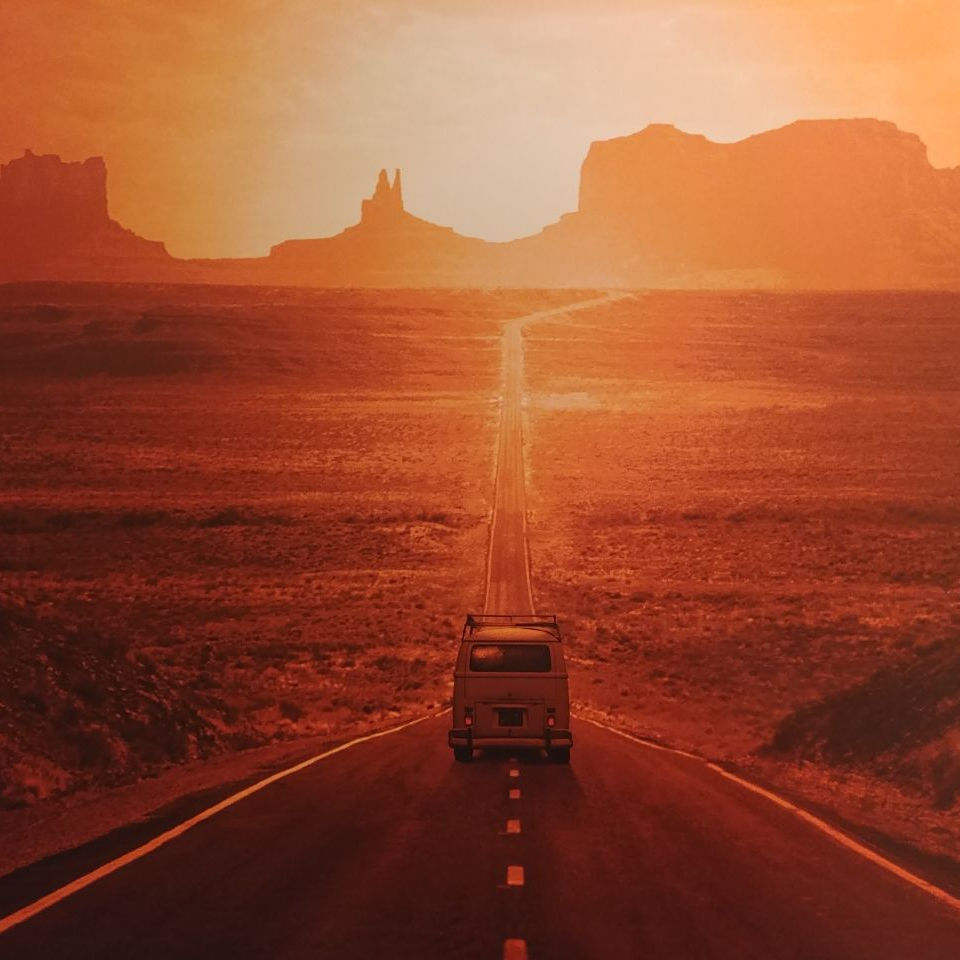}} \hspace{1pt} 
    \subfloat[Detection: 0\%] 
    {\includegraphics[width=0.243\textwidth]{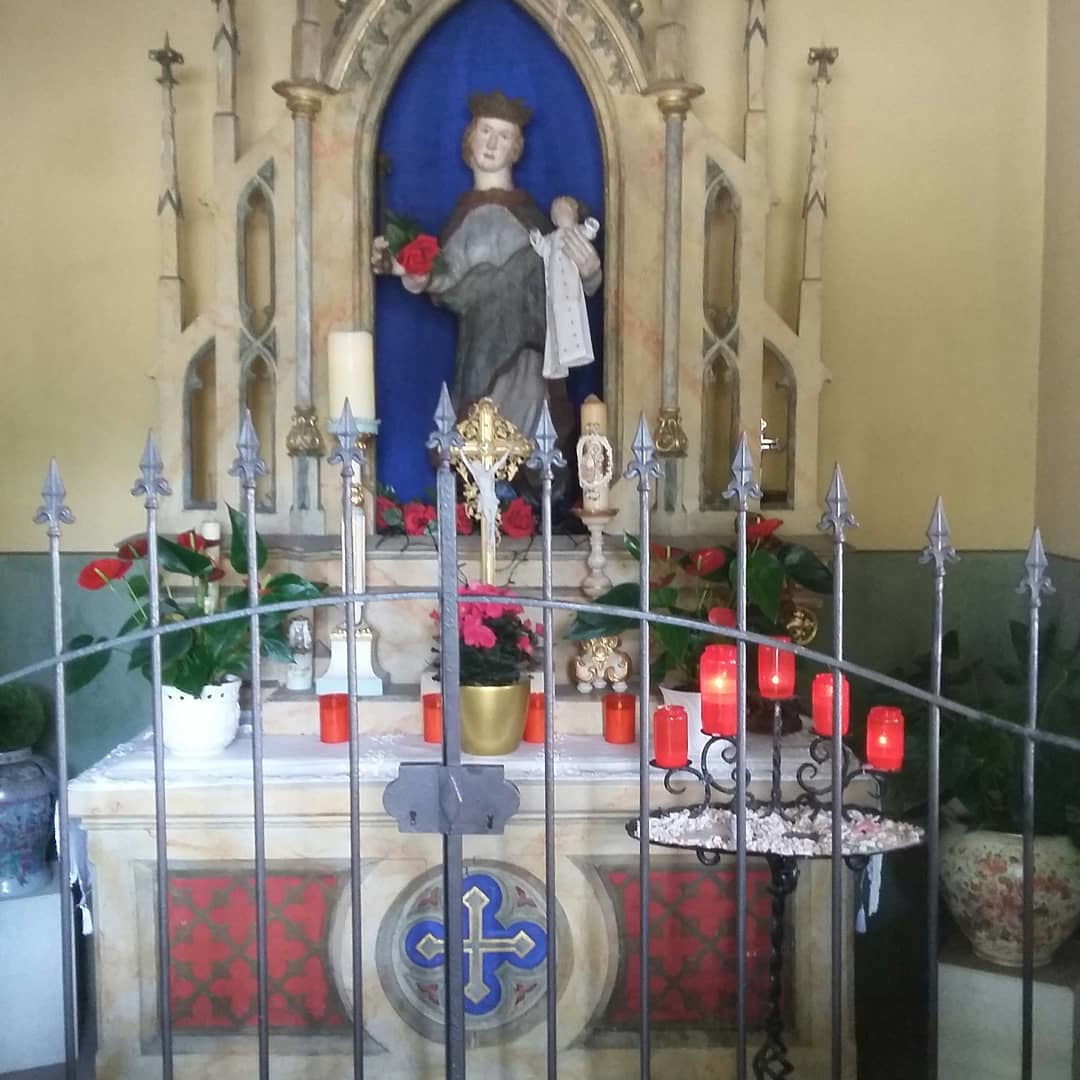}} \hspace{1pt} 
    \subfloat[Detection: 0\%] {\includegraphics[width=0.243\textwidth]{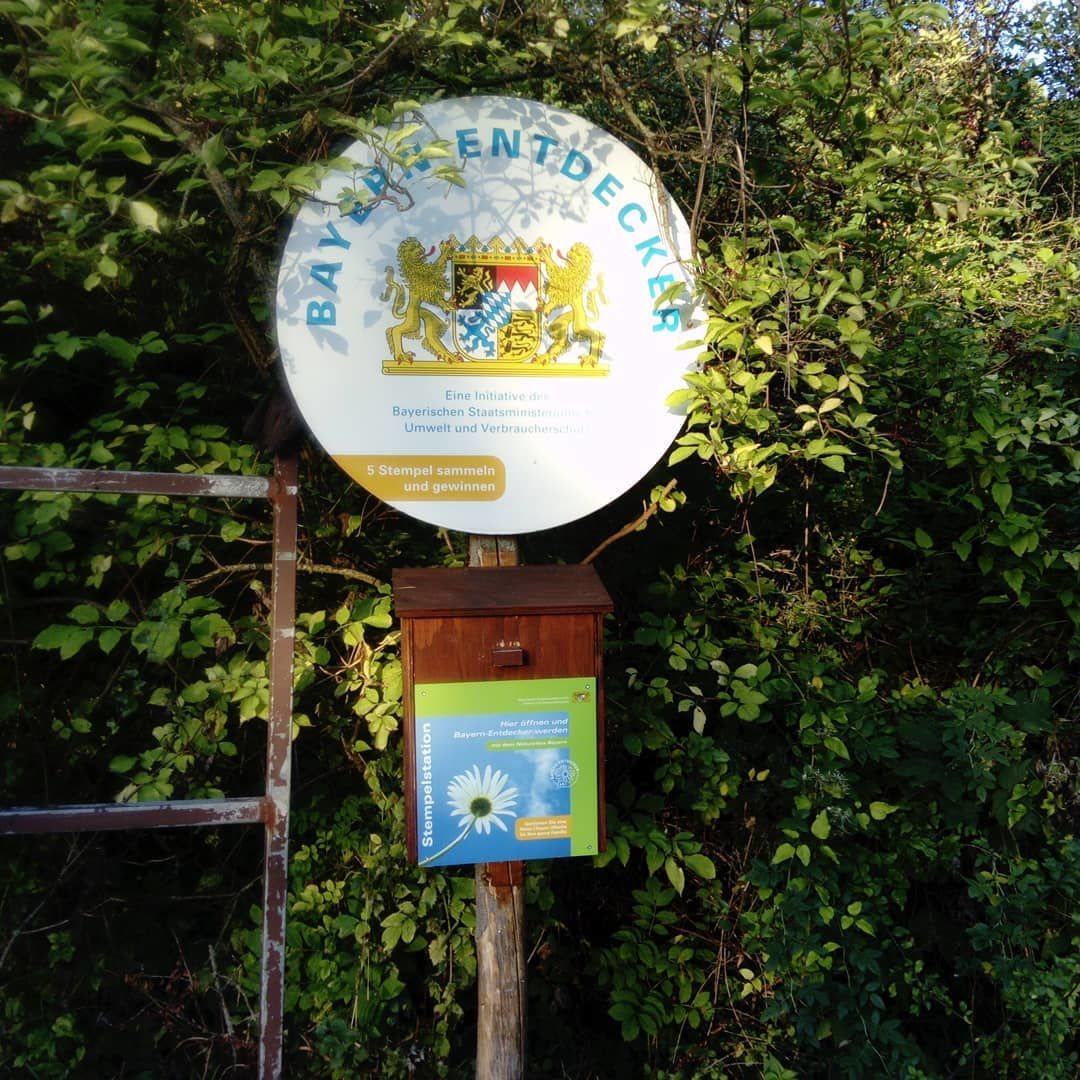}} \hspace{1pt}
    \subfloat[Detection: 0\%] 
    {\includegraphics[width=0.243\textwidth]{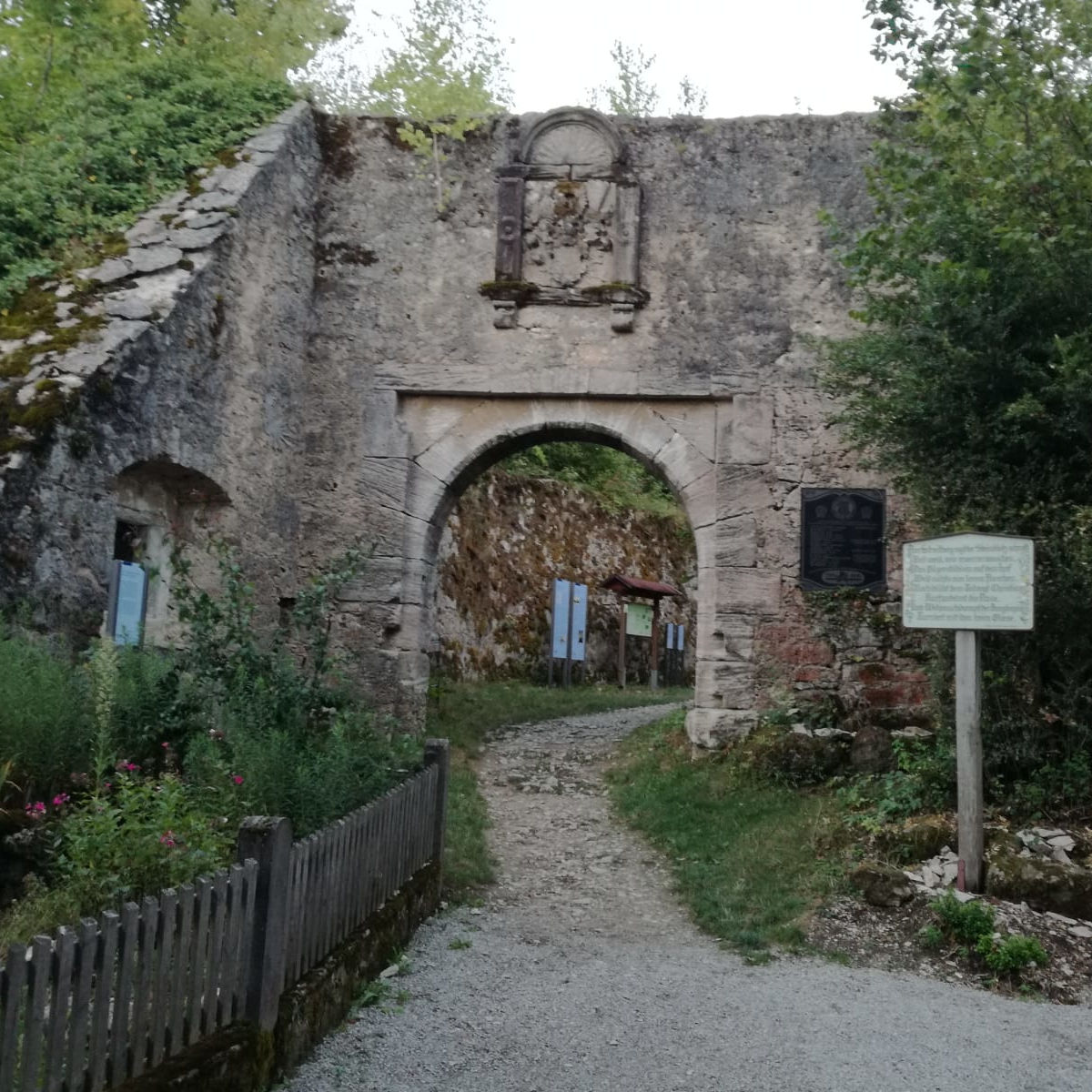}} 
    \caption{Accurately classified real images from FODB. For illustration purposes cropped to a square aspect ratio.}
    \label{fig:qualitative_fodb}
\end{figure*}

\begin{figure*}
    \centering 
    \subfloat[Detection: 0\%] 
    {\includegraphics[width=0.243\textwidth]{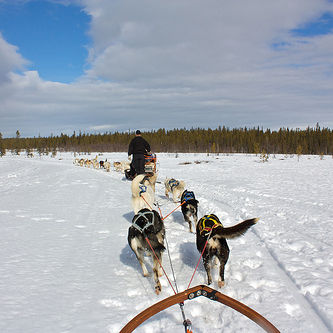}} \hspace{1pt} 
    \subfloat[Detection: 0\%] 
    {\includegraphics[width=0.243\textwidth]{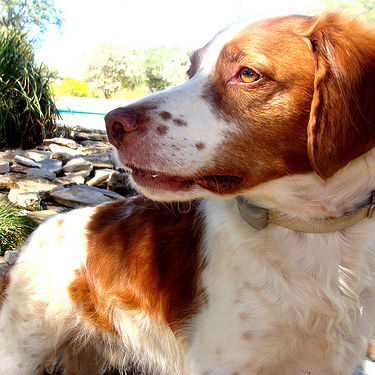}} \hspace{1pt} 
    \subfloat[Detection: 0\%] {\includegraphics[width=0.243\textwidth]{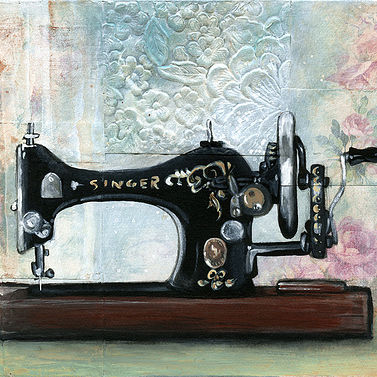}} \hspace{1pt}
    \subfloat[Detection: 0\%] 
    {\includegraphics[width=0.243\textwidth]{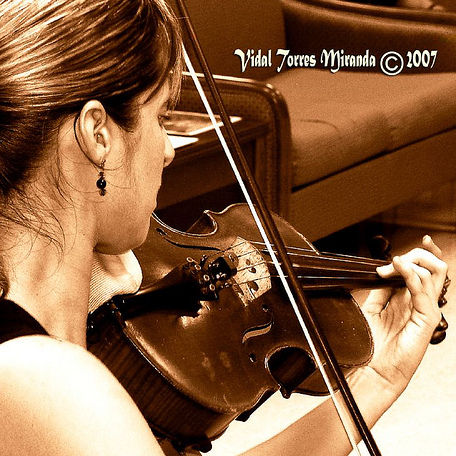}} 
    \caption{Accurately classified real images from ImageNet. For illustration purposes cropped to a square aspect ratio.}
    \label{fig:qualitative_imagenet}
\end{figure*}

\begin{figure*}
    \centering 
    \subfloat[Detection: 4\%] 
    {\includegraphics[width=0.243\textwidth]{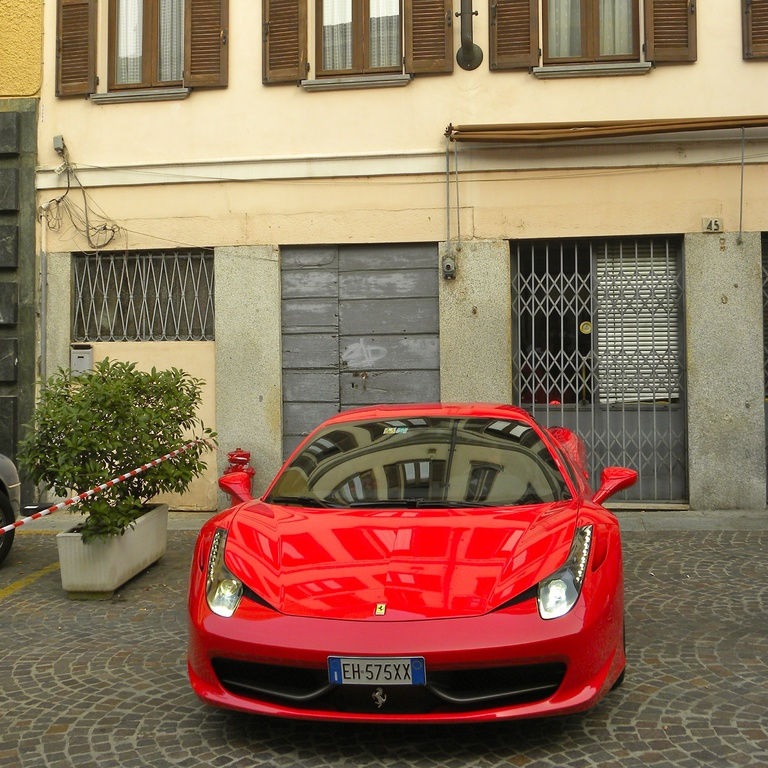}} \hspace{1pt} 
    \subfloat[Detection: 9\%] 
    {\includegraphics[width=0.243\textwidth]{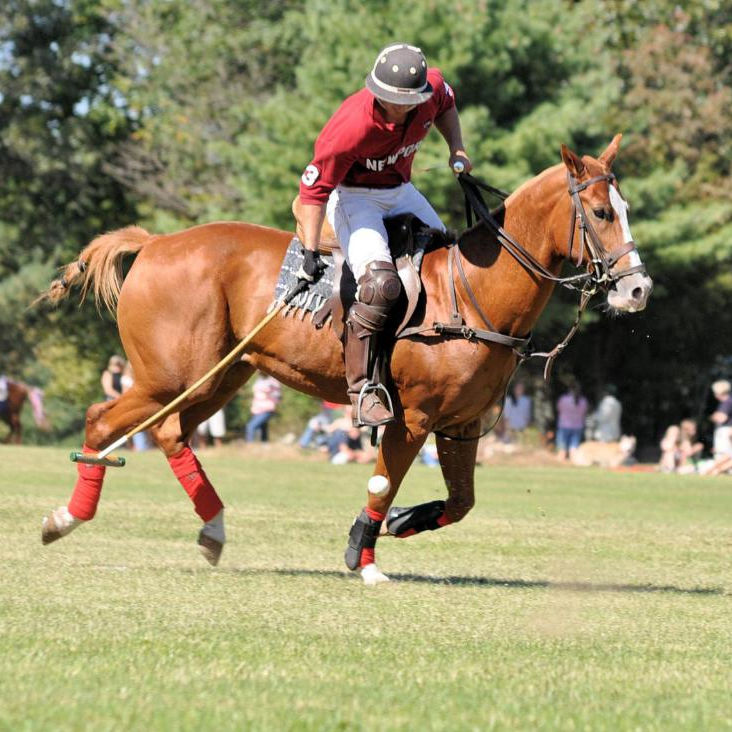}} \hspace{1pt} 
    \subfloat[Detection: 0\%] {\includegraphics[width=0.243\textwidth]{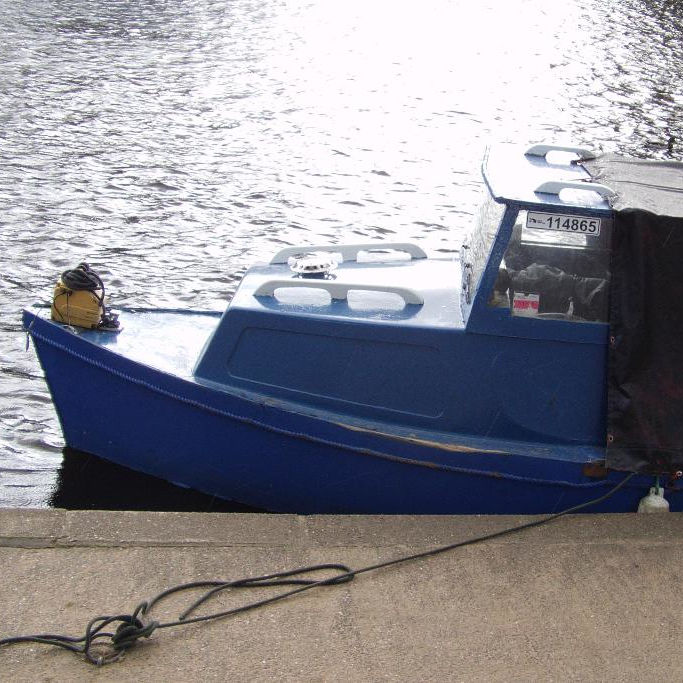}} \hspace{1pt}
    \subfloat[Detection: 0\%] 
    {\includegraphics[width=0.243\textwidth]{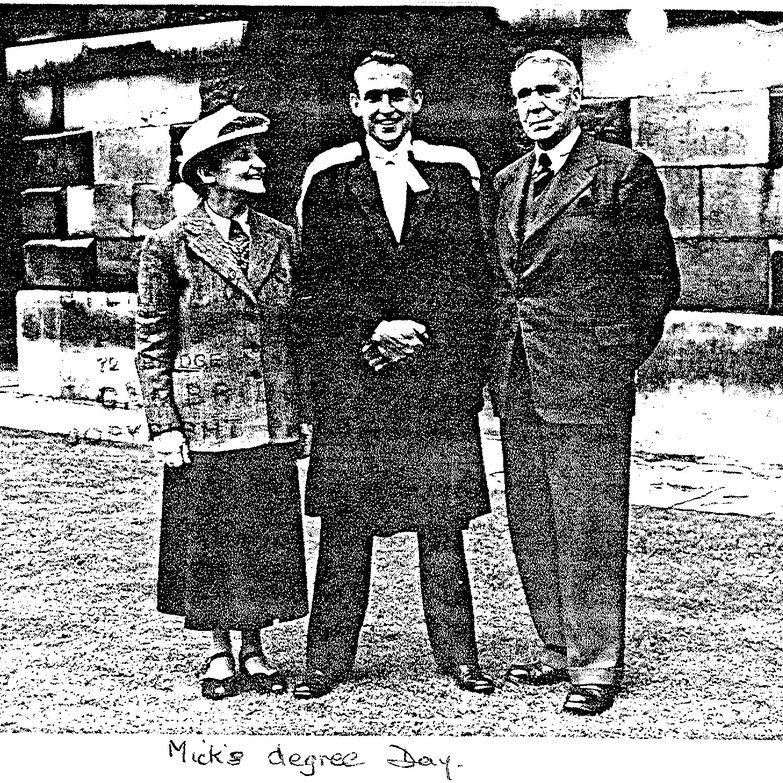}} 
    \caption{Accurately classified real images from Open Images. For illustration purposes cropped to a square aspect ratio.}
    \label{fig:qualitative_openimages}
\end{figure*}

\begin{figure*}
    \centering 
    \subfloat[Detection: 0\%] 
    {\includegraphics[width=0.243\textwidth]{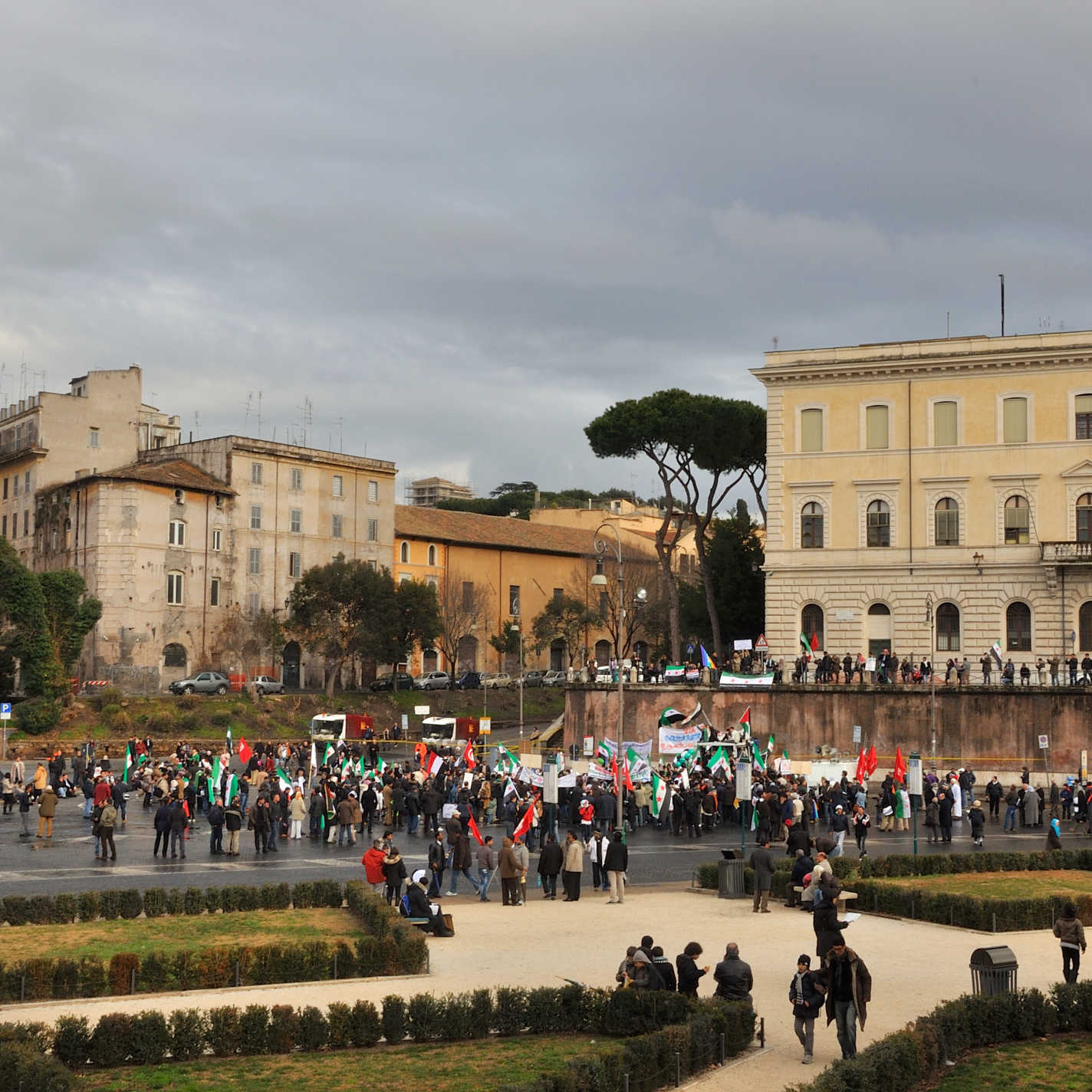}} \hspace{1pt} 
    \subfloat[Detection: 0\%] 
    {\includegraphics[width=0.243\textwidth]{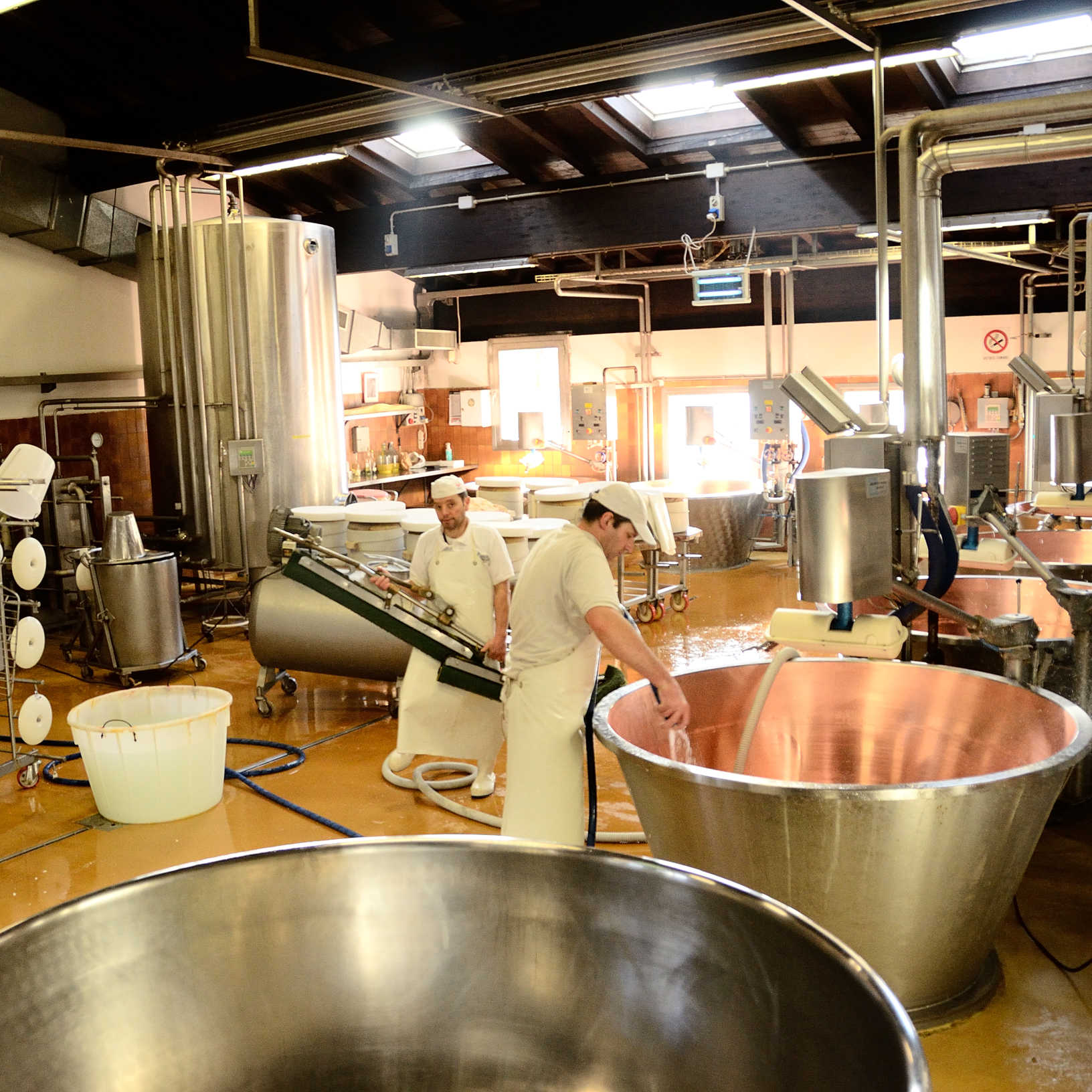}} \hspace{1pt} 
    \subfloat[Detection: 0\%] {\includegraphics[width=0.243\textwidth]{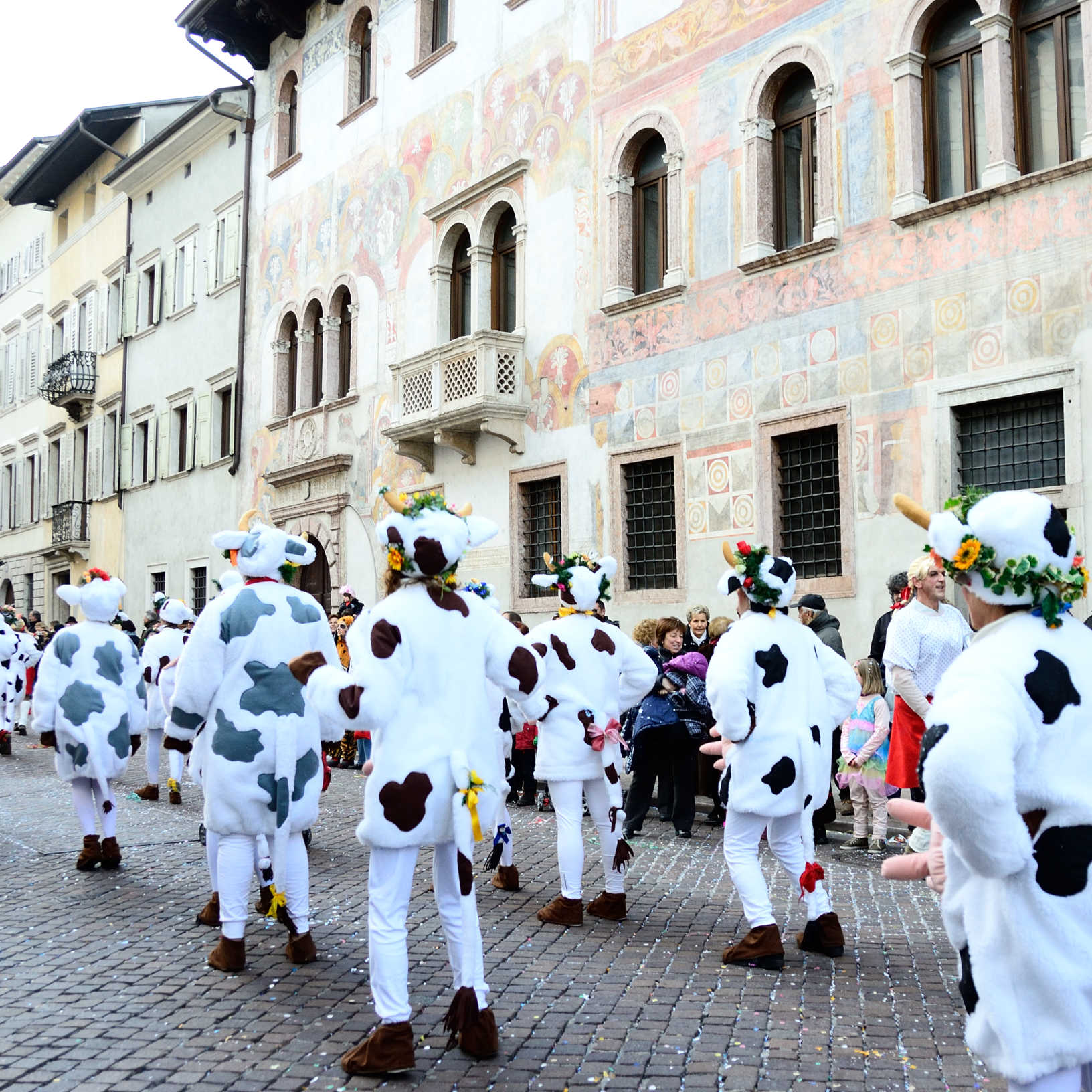}} \hspace{1pt}
    \subfloat[Detection: 0\%] 
    {\includegraphics[width=0.243\textwidth]{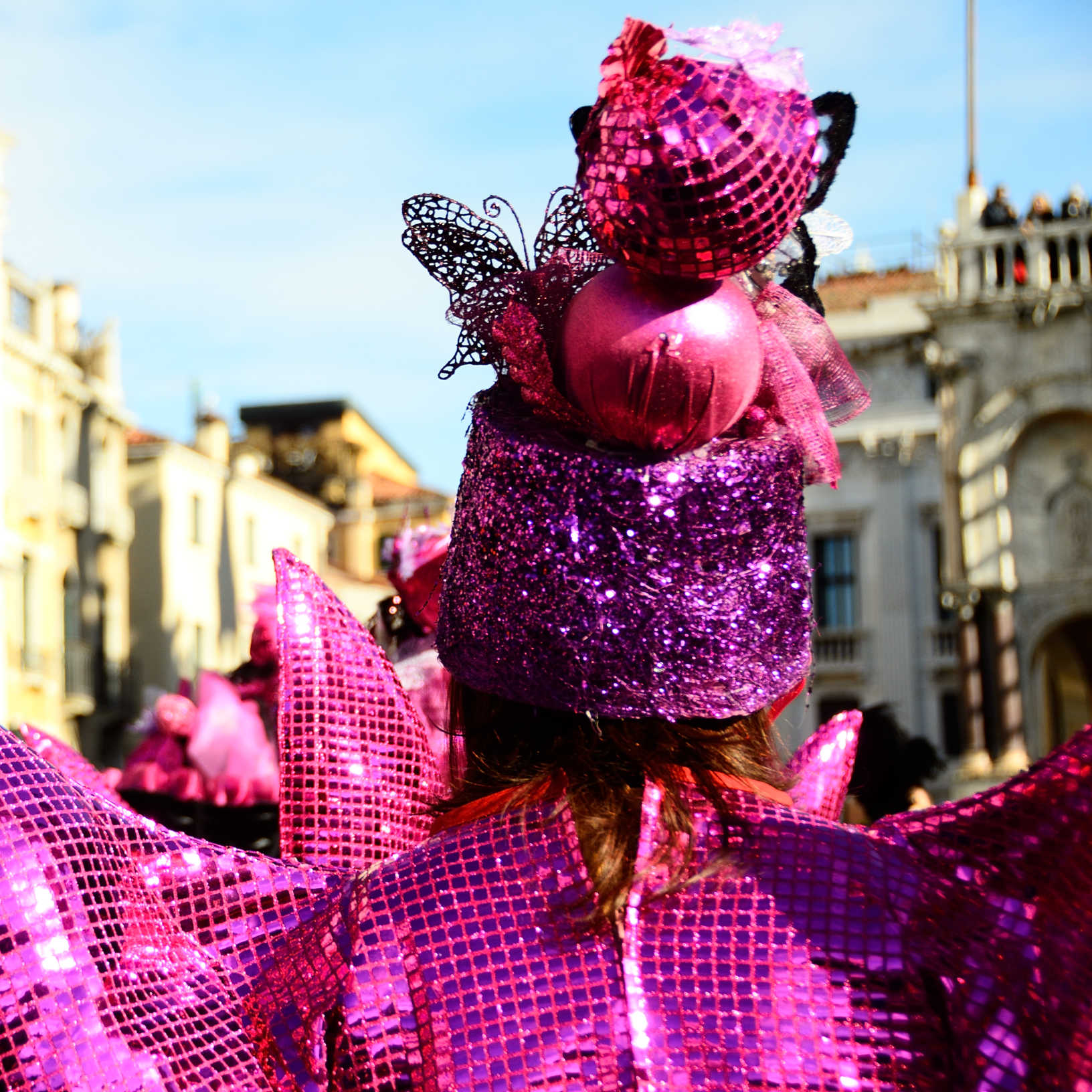}} 
    \caption{Accurately classified real images from RAISE-1k. For illustration purposes cropped to a square aspect ratio.}
    \label{fig:qualitative_raise}
\end{figure*}

We perform a qualitative evaluation of our approach across all the considered datasets and present samples for the 13 generative models in \cref{fig:qualitative_dalle2,fig:qualitative_dalle3,fig:qualitative_firefly,fig:qualitative_flux,fig:qualitative_gigagan,fig:qualitative_glide,fig:qualitative_mj5,fig:qualitative_mj61,fig:qualitative_sd13,fig:qualitative_sd14,fig:qualitative_sd2,fig:qualitative_sd3,fig:qualitative_sdxl} as well as for the 5 sources of real images in \cref{fig:qualitative_coco,fig:qualitative_fodb,fig:qualitative_imagenet,fig:qualitative_openimages,fig:qualitative_raise}. As we see, our approach accurately detects images generated by all the considered generative approaches, depicting a diverse set of topics and incorporating different levels of visual fidelity and aesthetics. At the same time, SPAI accurately classifies real images originating from all the employed sources. Therefore, employing spectral learning enables our architecture to not rely on some high-level semantics, but to effectively detect the subtle inconsistencies introduced by the generative models, to distinguish between AI-generated and real imagery.  

\section{Source Code}

To facilitate the reproduction of our results as well as further research in the field we make publicly available our source code, data and trained models on \href{https://mever-team.github.io/spai}{https://mever-team.github.io/spai}. 

\end{document}